\pdfoutput=1

\documentclass{article} %
\usepackage{arxiv,times}

\usepackage{amsmath,amsfonts,bm}

\def\eqref#1{equation~\ref{#1}}

\def\1{\bm{1}}

\DeclareMathAlphabet{\mathsfit}{\encodingdefault}{\sfdefault}{m}{sl}
\SetMathAlphabet{\mathsfit}{bold}{\encodingdefault}{\sfdefault}{bx}{n}

\usepackage{hyperref}
\usepackage{xurl} %
\usepackage{booktabs}
\usepackage{colortbl}
\usepackage{amsmath,amssymb}
\usepackage{graphicx}
\usepackage{tikz}

\newcommand{\ours}[0]{DisParQ}
\title{DisParQ~\disparticon: Self-Supervised Part Concepts for Interpretable Vision Foundation Models}

\newcommand{\RepoFigureRoot}{repo}
\newcommand{\QualHero}{kiteagamasailfour}%
\providecommand{\RepoFigureRoot}{docs/paper}
\usepackage{tikz,graphicx,amsmath,amssymb}
\usetikzlibrary{arrows.meta,calc,positioning,backgrounds,fit}
\newcommand{\CRISPSetup}{%
\definecolor{ink}{HTML}{303437}
\definecolor{line}{HTML}{747A7D}
\definecolor{blue}{HTML}{E1EDF4}
\definecolor{mint}{HTML}{E4EFE6}
\definecolor{peach}{HTML}{F8E8DA}
\definecolor{lavender}{HTML}{ECE7F2}
\definecolor{papergray}{HTML}{F6F6F3}
\definecolor{focus}{HTML}{8AAEC3}
\tikzset{
 every node/.style={font=\fontsize{8}{9.2}\selectfont,text=ink,align=center,inner sep=1mm},
 box/.style={draw=line,line width=.45pt,rounded corners=1mm,fill=papergray,minimum height=10mm},
 flow/.style={-{Stealth[length=1.7mm,width=1.35mm]},draw=line,line width=.8pt},
 wire/.style={draw=line,line width=.65pt},
 note/.style={font=\fontsize{7}{8}\selectfont,text=line},
 heading/.style={font=\bfseries\fontsize{10}{11}\selectfont,anchor=west,inner sep=0},
 sub/.style={font=\itshape\fontsize{8}{9}\selectfont},
 cell/.style={draw=white,line width=.35pt},
}
\newcommand{\FocusID}{185}
\newcommand{\FocusSlot}{2}
\newcommand{\FocusFirst}{51}
\newcommand{\FocusSecond}{58}
\newcommand{\FocusLast}{50}
\newcommand{\ExampleList}{11/185,34/675,57/283}
\newcommand{\SelectedIDPreview}{675\quad327\quad185\quad\cdots\quad20}
\newcommand{\FocusCodes}{51,58,41,14,20,39,35,2,42,12,5,11,51,58,6,50}

\newcommand{\ownership}[2]{%
 \begin{scope}[shift={##1},scale=##2]
 \foreach \row/\vals in {0/{0,1,1,0,0,2},1/{0,1,1,0,2,2},2/{0,0,3,3,0,0},3/{2,2,3,3,1,1},4/{2,0,0,0,0,1},5/{0,0,2,2,0,0}}{
  \foreach \v [count=\col from 0] in \vals {
   \ifcase\v\def\cc{papergray}\or\def\cc{peach}\or\def\cc{mint}\or\def\cc{blue}\fi
   \path[cell,fill=\cc] (\col,-\row) rectangle ++(1,-1);
  }}
 \draw[line,line width=.4pt] (0,0) rectangle (6,-6);
 \end{scope}}
\newcommand{\vectorbar}[3]{%
 \foreach \k in {0,...,5}{\path[draw=white,fill=##3] ($(##1)+({\k*##2/6},0)$) rectangle ++({##2/6},2.8);}}
\newcommand{\draftfooter}[1]{\node[note,anchor=west,inner sep=0] at (0,##1) {Draft visual example: older checkpoint. Colored grids and vectors are schematic.};}
}
\newcommand{\CRISPOverviewCaption}{%
\textbf{\ours{} describes images with concepts and attributes.} Fused and upsampled VFM features $z_u$ are matched against prototypes $w_n$ to determine which concepts occur (top-$K$ identities $i_{1:K}$) and where they occur (assignments $p_u$ at positions $u$). Discrete attributes $q_{i_k}$ encode how each concept appears in the image. The assignment map selects the learned concept embeddings $e_n$ and attributes for the decoder, which predicts $\hat c(x)$ by distillation from the VFM's global representation.}

\newcommand{\CRISPQuantizationCaption}{%
\textbf{Separating concept identity and appearance.} Selected-concept maps $\bar{A}_{i_k}$ determine exclusive spatial assignments $p_u$ and pool each concept's region into $\bar z_{i_k}$. Its normalized deviation from the concept prototype, $r_{i_k}$, describes the occurrence relative to the concept. Product codebooks $\mathcal B_{i_k}$ quantize its projection. The resulting attribute $q_{i_k}$ is added to the concept embedding $e_{i_k}$ and broadcast to its assigned positions. Adding positional embeddings $\mathrm{PE}_u$ forms the decoder tokens $t_u$.}

\newcommand{\CRISPTrainingCaption}{%
\textbf{Training objectives.} Two affine-related views share model parameters, and each prediction matches its own detached VFM target. Potts, entropy, slot presence, and orthogonality use the original view. Distillation and usage diversity average over views; bank presence takes maxima across both; equivariance compares aligned full-bank channels. $\dagger$: auxiliary distillation is used only in continuous training. $\ddagger$: assignment regularizers train the front end in stage 1 but have no gradient to it during frozen PQ training. Orthogonality can still update region modulation in stage 2.}

\usepackage{colortbl}
\usepackage{wrapfig}%
\definecolor{oursrow}{HTML}{E1EDF4}   %

\usepackage{twemojis}

\usepackage{graphicx}

\DeclareRobustCommand{\disparticon}{%
  \raisebox{-0.1em}{%
    \includegraphics[height=0.9em]{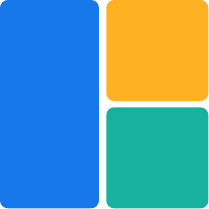}%
  }%
}

\newtoggle{comments}
\newtoggle{BSetal}

\toggletrue{comments}
\toggletrue{BSetal}

\newtoggle{qualfourrows}
\toggletrue{qualfourrows}
\newtoggle{tightspacing}
\togglefalse{tightspacing}%
\makeatletter
\newskip\sid@parskip \newskip\sid@textfloatsep \newskip\sid@floatsep \newskip\sid@intextsep
\newskip\sid@dbltextfloatsep \newskip\sid@dblfloatsep
\AtBeginDocument{%
  \sid@parskip=\parskip \sid@textfloatsep=\textfloatsep \sid@floatsep=\floatsep \sid@intextsep=\intextsep
  \sid@dbltextfloatsep=\dbltextfloatsep \sid@dblfloatsep=\dblfloatsep \let\sid@paragraph\paragraph}
\newcommand{\mainpaperspacing}{\iftoggle{tightspacing}{%
  \setlength{\parskip}{4pt plus 1pt}%
  \def\paragraph{\@startsection{paragraph}{4}{\z@}{0.5ex plus 0.2ex minus .1ex}{-1em}{\normalsize\bf}}%
  \setlength{\textfloatsep}{10pt plus 2pt minus 3pt}\setlength{\dbltextfloatsep}{10pt plus 2pt minus 3pt}%
  \setlength{\floatsep}{8pt plus 2pt minus 2pt}\setlength{\dblfloatsep}{8pt plus 2pt minus 2pt}%
  \setlength{\intextsep}{8pt plus 2pt minus 2pt}}{}}
\newcommand{\defaultspacing}{%
  \parskip=\sid@parskip \textfloatsep=\sid@textfloatsep \floatsep=\sid@floatsep \intextsep=\sid@intextsep
  \dbltextfloatsep=\sid@dbltextfloatsep \dblfloatsep=\sid@dblfloatsep \let\paragraph\sid@paragraph}
\makeatother

\usepackage[dvipsnames]{xcolor}

\newcommand\sid[1]{\iftoggle{comments}{\textcolor{blue}{Sid: #1}}{}}

\newcommand\sukrut[1]{\iftoggle{comments}{\textcolor{ForestGreen}{Sukrut: #1}}{}}
\newcommand{\bernt}[1]{\iftoggle{BSetal}{\textcolor[rgb]{0.87, 0.19, 0.39}{\textbf{Bernt:} #1}}{}}

\DeclareRobustCommand\sidreview{\iftoggle{comments}{\ \textcolor{red}{Sid: Please review}}{}}

\newcommand{\myparagraph}[2][0]{\vspace{#1em}\noindent{\bf #2}}

\usepackage{placeins}

\newcommand{\AblationFusionCaption}{%
Layer fusion versus final-block features under the bilinear stage-1 control. Gray entries are absolute reference scores, and colored entries are differences in pp, with SDs as defined in Appendix~\ref{app:ablations}.}

\newcommand{\AblationAttributesCaption}{%
Continuous appearance attributes during concept discovery under the bilinear stage-1 control. This differs from removing attributes after freezing concept assignment. Gray entries are absolute reference scores, and colored entries are differences in pp, with SDs as defined in Appendix~\ref{app:ablations}.}

\newcommand{\AblationDecoderCaption}{%
Spatial concept decoder versus the auxiliary-only region readout under the bilinear stage-1 control. The auxiliary readout consumes continuous pooled region features. The stage-1 decoder consumes concept identities, spatial assignments, and continuous appearance attributes. The comparison changes both training and readout, as discussed in Appendix~\ref{abl:decoder}. Gray entries are absolute reference scores, and colored entries are differences in pp, with SDs as defined in Appendix~\ref{app:ablations}.}

\newcommand{\AblationUpscalingCaption}{%
Feature upsampling with continuous AnyUp training as the reference. The 20\% AnyUp arm mixes AnyUp and bilinear interpolation during training and uses full AnyUp at evaluation. Native resolution changes both the assignment and decoder grids to $16\times16$. Gray entries are absolute reference scores, and colored entries are differences in pp, with SDs as defined in Appendix~\ref{app:ablations}.}

\newcommand{\AblationDepthCaption}{%
Decoder depth at fixed width 768 under the bilinear stage-1 control. Zero blocks denotes mean pooling, and labels include decoder parameter counts. The gray three-block row gives absolute reference scores. Colored entries are differences in pp, with SDs as defined in Appendix~\ref{app:ablations}.}

\newcommand{\AblationBankCaption}{%
Concept-bank size $N$ at fixed $K=16$ during continuous DINOv2-B/14 discovery with AnyUp, using the sweep setup in Appendix~\ref{abl:bank}. The gray $N=1024$ row gives absolute reference scores, and colored entries give differences in pp. Each row is one completed run.}

\newcommand{\AblationKCaption}{%
Selected concept count $K$ at fixed $N=1024$ during continuous DINOv2-B/14 discovery with AnyUp. The sweep setup and accompanying equivariance budgets are described in Appendix~\ref{abl:bank-sec}. The gray $K=16$ row gives absolute reference scores, and colored entries give differences in pp. Each row is one completed run.}

\newcommand{\AblationQuantisationCaption}{%
Attribute quantizers after equal 200-epoch continuation from a seed-matched, frozen bilinear concept model. Entries are means over three seeds. The gray private PQ16$\times$64 row gives absolute reference scores, and other score entries give paired differences, in pp for LP, mAP, and kNN. Sample SDs appear beneath scores, using paired-difference SDs outside the reference row. Cosine loss is the single-view evaluation cosine distance between the decoder output and frozen global target, in raw units. Rates are absolute attribute-index bits per selected concept in every row and exclude model parameters.}

\newcommand{\AblationGradientsCaption}{%
Gradient routes under the bilinear stage-1 control. The identity-surrogate removal preserves hard forward identities, while the residual-gradient arm enables derivatives through the appearance residual. Gray entries are absolute reference scores, and colored entries are differences in pp, with SDs as defined in Appendix~\ref{app:ablations}.}

\newcommand{\AblationSelectionCaption}{%
Selection rules and prototype maintenance under the bilinear stage-1 control. The zero-bias Sinkhorn and zero-bias mass rows form the direct selector comparison. Gray entries are absolute reference scores, and colored entries are differences in pp, with SDs as defined in Appendix~\ref{app:ablations}.}

\newcommand{\AblationObjectiveCaption}{%
Individual auxiliary or regularization loss removals under the bilinear stage-1 control, retaining the global decoder loss. Objectives are defined in Appendix~\ref{app:objectives}. Gray entries are absolute reference scores, and colored entries are differences in pp, with SDs as defined in Appendix~\ref{app:ablations}.}

\definecolor{linkblue}{HTML}{3A7DC0}
\hypersetup{colorlinks=true, citecolor=linkblue, linkcolor=linkblue, urlcolor=linkblue}

\iclrarxivcopy

\author{Adam Pardyl$^{1,2*}$, Siddhartha Gairola$^{3*}$, Sukrut Rao$^{3}$, Adam Wróbel$^{1,2}$, \\
\bfseries Bartosz Zieliński$^{1,4}$, Bernt Schiele$^{3}$, Dawid Rymarczyk$^{1,5}$%
\thanks{$^*$Equal contribution. $^1$Jagiellonian University, Faculty of Mathematics and Computer Science; $^2$Jagiellonian University, Doctoral School of Exact and Natural Sciences; $^3$Max Planck Institute for Informatics, Saarland Informatics Campus; $^4$Jagiellonian University, Jagiellonian Center for Artificial Intelligence; $^5$Ardigen SA. Correspondence to: Adam Pardyl, adam.pardyl@doctoral.uj.edu.pl}
}

\begin{document}

\begingroup
\makeatletter
\renewcommand{\@fnsymbol}[1]{}
\maketitle
\makeatother
\endgroup

\mainpaperspacing%

\providecommand{\TeaserVersion}{5}
\providecommand{\TeaserLayout}{boxed}
\providecommand{\TeaserThreeCaption}{\textbf{\ours{} describes an image by which part concepts occur and where they are.} It is trained entirely self-supervised on a frozen vision foundation model backbone without class labels, part annotations, or language. In each panel, \emph{left:} the concepts \ours{} assigns to the input; the four that contribute the most to the prediction are numbered. \emph{Right:} the contribution of each of these concepts to the predicted class.%
  }
\ifnum\TeaserVersion=5
\providecommand{\TeaserBottleneckCaption}{%
\textbf{\ours{} describes images through semantic visual concepts, their exclusive location assignment,
and per-concept appearance attributes.}
\emph{Left:} a learned interpretable bottleneck on a frozen VFM
identifies part concepts and assigns each location to one concept. Attribute examples illustrate similar appearances within the same learned concept. The concept representation is learned without class labels, part annotations, or language.
\emph{Right:} Top four concepts explain a prediction.}

\vspace{-4mm}
\begin{center}
  \begingroup
\providecommand{\AlternativeTeaserWidth}{\linewidth}
\newdimen\altUnit
\pgfmathsetlength{\altUnit}{\AlternativeTeaserWidth/140}
\def\altAssets{\RepoFigureRoot/figures/teaser/alternative/assets}
\definecolor{altInk}{HTML}{303437}
\definecolor{altLine}{HTML}{747A7D}
\definecolor{altBlue}{HTML}{BEDCEC}
\definecolor{altYellow}{HTML}{FFE496}
\definecolor{altFrozen}{HTML}{285F91}
\definecolor{altLearned}{HTML}{A87825}
\definecolor{altC0}{HTML}{0072B2}
\definecolor{altC1}{HTML}{F2C200}
\definecolor{altC2}{HTML}{009E73}
\definecolor{altC3}{HTML}{8E44AD}
\begin{tikzpicture}[x=\altUnit,y=\altUnit,
 every node/.style={font=\footnotesize,text=altInk,inner sep=0,outer sep=0,align=center},
 altNote/.style={font=\scriptsize,text=altLine},
 altTitle/.style={font=\scriptsize\bfseries,anchor=base west},
 altFlow/.style={draw=altLine,line width=.55pt,-{Stealth[length=1.25mm,width=.95mm]}},
 altTag/.style={circle,fill=#1,text=white,font=\tiny\bfseries,inner sep=.3mm,minimum size=2.3mm}]
\path[use as bounding box] (0,3) rectangle (140,43);
\node[altTitle] at (0,41) {(a) Interpretable image description};
\node[altTitle] at (96,41) {(b) Explaining a prediction};
\draw[altLine!40,line width=.4pt] (89.5,3)--(89.5,39);

\begin{scope}[shift={(-7.5,1.5)}]
\node (input) at (11.25,21) {\includegraphics[width=7.5\altUnit]{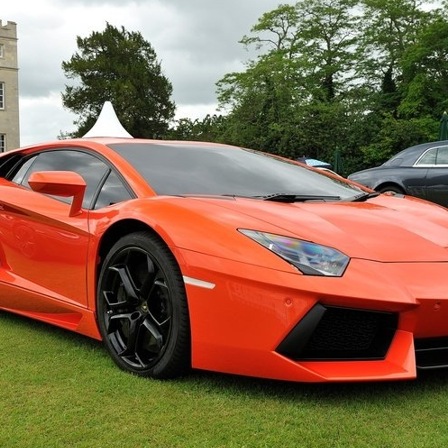}};
\node[altNote,anchor=north] at (11.25,15) {Input\\image};
\path[draw=altLine,fill=altBlue,line width=.45pt]
 (20,27)--(31,25)--(31,17)--(20,15)--cycle;
\node at (25.5,21) {VFM};
\node[altNote,anchor=north] at (25.5,13) {Frozen Vision\\Foundational\\Model};
\begin{scope}[shift={(22.3,24.9)},draw=altFrozen,line width=.55pt]
 \foreach \a in {0,60,...,300}{\begin{scope}[rotate=\a]
  \draw (0,0)--(0,1.2);\draw (-.32,.56)--(0,.88)--(.32,.56);
 \end{scope}}
\end{scope}
\node[draw=altLine,fill=altYellow,line width=.45pt,rounded corners=.5mm,
 minimum width=13\altUnit,minimum height=12\altUnit] (method) at (42.5,21) {\ours{}};
\begin{scope}[shift={(46.5,24.7)}]
 \path[fill=altLearned] (0,-1.5) .. controls (-2,-1.2) and (-1.3,.4) .. (-.6,.8)
 .. controls (-.7,0) and (-.2,.2) .. (0,1.8)
 .. controls (1.8,.4) and (2,-1.2) .. (0,-1.5);
 \path[fill=altYellow] (0,-1.1) .. controls (-.7,-.7) and (-.4,-.1) .. (.1,.5)
 .. controls (.7,-.5) and (.7,-.9) .. (0,-1.1);
\end{scope}
\node[altNote,anchor=north] at (42.5,13) {Learned\\interpretable\\bottleneck};
\draw[altFlow] (input.east)--(20,21);
\draw[altFlow] (31,21)--(method.west);
\draw[altLine,line width=.55pt] (method.east)--(52,21);
\draw[altLine,line width=.55pt] (52,10.5)--(52,31.5);
\foreach \yy in {31.5,21,10.5}{\draw[altFlow] (52,\yy)--(55,\yy);}

\foreach \xx/\file/\col in {60.5/car_wheel/altC2,71/car_headlight/altC1,81.5/car_vent/altC0}{
 \node[draw=\col,line width=1.6pt] (identity-\file) at (\xx,31.5)
  {\includegraphics[width=7.5\altUnit]{\altAssets/\file.jpg}};
}
\node[altNote] at (89,31.5) {$\cdots$};
\coordinate (identityTitleEdge) at (56,0 |- identity-car_wheel.north);
\node[altTitle,anchor=south west] at ([yshift=1.2mm]identityTitleEdge)
 {Concepts (what is it?)};

\node[altTitle,anchor=west] at (69.5,21) {Locations (where is it?)};
\node[anchor=south west] (locations) at (56.75,17.25)
 {\includegraphics[width=7.5\altUnit]{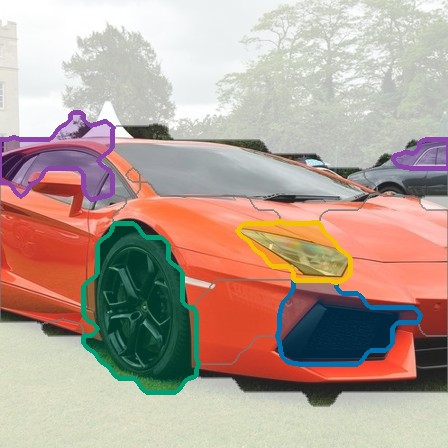}};
\draw[altLine,line width=.4pt] (56.75,17.25) rectangle (64.25,24.75);
\foreach \sx/\sy/\file/\col in {
 59.177/19.184/car_wheel/altC2,
 61.973/20.480/car_headlight/altC1,
 62.074/19.326/car_vent/altC0}{
 \draw[draw=\col!85!altInk,line width=.45pt,line cap=round,line join=round]
  (\sx,\sy)--(identity-\file.south);
}

\foreach \xx/\file/\lw in {60.5/car_headlight/1.8,71/headlight_variant_1/.45,81.5/headlight_variant_2/.45}{
 \node[draw=altC1,line width=\lw pt] (attribute-\file) at (\xx,10.5)
  {\includegraphics[width=7.5\altUnit]{\altAssets/\file.jpg}};
}
\coordinate (attributeTitleEdge) at (56,0 |- attribute-car_headlight.south);
\node[altTitle,anchor=north west] at ([yshift=-1.2mm]attributeTitleEdge)
 {Attributes (how does it look?)};
\draw[draw=altC1!85!altInk,line width=.45pt,line cap=round,line join=round]
 (61.973,20.480)--(attribute-car_headlight.north);
\draw[draw=altLine,line width=.45pt]
 (attribute-car_headlight.north east)--(66,15.6)--(69,15.6);
\draw[altFlow,line width=.45pt] (69,15.6)--(attribute-headlight_variant_1.north);
\draw[altFlow,line width=.45pt] (69,15.6)--(79.5,15.6)--(attribute-headlight_variant_2.north);
\node[altNote,anchor=south] at (73,16.1) {similar to};

\end{scope}

\node[altNote,anchor=base] at (131,39) {Contribution};
\node[anchor=south west] at (96,13) {\includegraphics[width=24\altUnit]{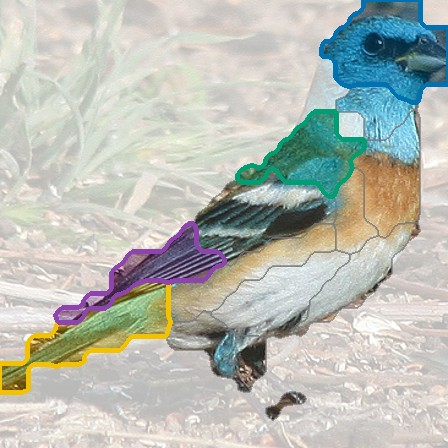}};
\draw[altLine,line width=.4pt] (96,13) rectangle (120,37);
\foreach \xx/\yy/\num/\col/\txt in {116.625/34.696/1/altC0/white,103.607/20.393/2/altC1/black,112.821/28.804/3/altC2/white,105.696/23.072/4/altC3/white}{
 \node[altTag=\col,text=\txt,draw=white,line width=.5pt] at (\xx,\yy) {\num};
}
\foreach \yy/\idx/\num/\col/\txt/\length/\pct in {
 34.8/0/1/altC0/white/6.5/36,
 29.6/1/2/altC1/black/5.279/29,
 24.4/2/3/altC2/white/2.972/16,
 19.2/3/4/altC3/white/1.744/10}{
 \node[altTag=\col,text=\txt] at (122,\yy) {\num};
 \node[draw=\col,line width=.65pt] at (126,\yy)
  {\includegraphics[width=4\altUnit]{\altAssets/bird_q\idx.jpg}};
 \fill[\col] (129,{\yy-1}) rectangle ++(\length,2);
 \node[font=\scriptsize,anchor=west] at ({129.5+\length},\yy) {\pct\%};
}
\node[altNote,font=\fontsize{6.5}{7.5}\selectfont,anchor=west] at (121,13.4) {Other concepts};
\node[altNote,anchor=west] at (136,13.4) {8\%};
\node[font=\scriptsize\bfseries] at (118,8.5) {Lazuli Bunting};
\node[altNote] at (118,5) {Prediction confidence: $96\%$};
\end{tikzpicture}
\endgroup
  \makeatletter\def\@captype{figure}\makeatother
  \vspace{-6mm}
  \caption{\TeaserBottleneckCaption}
  \label{fig:teaser}
\end{center}
\vspace{2mm}
\else\ifnum\TeaserVersion=3
\begin{figure}[b]
  \vspace{-3.5mm}%
  \centering
  \includegraphics[width=\linewidth]{\RepoFigureRoot/figures/teaser/teaser_v3.pdf}%
  \vspace{-2mm}%
  \caption{\TeaserThreeCaption}
  \label{fig:teaser}
\end{figure}
\else\ifnum\TeaserVersion=4
\vspace{-4mm}%
\begin{center}
  \includegraphics[width=\linewidth]{\RepoFigureRoot/figures/teaser/teaser_v3.pdf}%
  \makeatletter\def\@captype{figure}\makeatother
  \vspace{-4mm}%
  \caption{\TeaserThreeCaption}
  \label{fig:teaser}
\end{center}
\vspace{2mm}%
\else
\begin{center}
  \ifnum\TeaserVersion=1
    \def\TeaserFile{teaser_v1}%
    \def\TeaserExample{wheels $\rightarrow$ wheels and tail lights  $\rightarrow$ tail lights}%
  \else
    \def\TeaserFile{teaser_v2}%
    \def\TeaserExample{headlights  $\rightarrow$ headlights, air vents  $\rightarrow$ air vents, wheels
       $\rightarrow$ wheels, and windows  $\rightarrow$ windows}%
  \fi
  \long\def\TeaserBoxed{boxed}%
  \ifx\TeaserLayout\TeaserBoxed
    \includegraphics{\RepoFigureRoot/figures/teaser/\TeaserFile_boxed.pdf}%
  \else
    \includegraphics{\RepoFigureRoot/figures/teaser/\TeaserFile_plain.pdf}%
  \fi
  \makeatletter\def\@captype{figure}\makeatother
  \vspace{-1mm}
  \caption{\textbf{\ours{} describes an image by which part concepts occur, where they are, and   how they look.} It is trained entirely self-supervised on a frozen backbone without class labels, part annotations, or language. \emph{Left:} an input and the concepts \ours{} assigns   to it; the four that contribute the most to the prediction are numbered. \emph{Middle:} each of these
  concepts next to its nearest occurrences in the training set, ranked by attribute similarity within the same concept identity: \TeaserExample{} \sukrut{This is quite confusing}, across car models. \emph{Right:} the contribution of each concept to the predicted class: the drop in that class's logit when the concept is removed from the discrete description and the image is decoded again. \bernt{I like the teaser figure in general - but it is a bit complicated - and thus i would not have it a the top of the first page but at the bottom - at least after the abstract - probably literally at the bottom of page 1. Also - it might be better to have a simpler message but 2-3 examples instead? E.g. I find the concept nearest neighbor visualization in the middle not that easy to digest first view of this? So maybe having the concept map and contributions only for 2-3 examples instead?} \sid{Ok! will create another variant and move to the bottom of P1 / abstract.}}
  \label{fig:teaser}
\end{center}
\fi\fi\fi

\begin{abstract}
\vspace{-1.5mm}%
Concept-based vision models represent images through an intermediate layer of human-inspectable concepts, so what a model relies on can be traced to those concepts.
However, those models %
are often limited to fixed categories or depend on language to define their concepts. We introduce \textbf{\ours{}} (\textbf{Dis}crete \textbf{Par}ts with \textbf{Q}uantized attributes), a method that learns spatially grounded, discrete concept representations from a powerful frozen vision-only self-supervised backbone. It requires no class labels and no language supervision. Each image patch is assigned to exactly one concept from a learnable prototype dictionary and only a sparse subset of concepts may activate per image. %
To capture how each concept varies across images (\emph{e.g.,} the type of a ``wheel''), we learn continuous residuals alongside the concepts, and then quantize them into discrete attributes. A spatial decoder reconstructs the backbone's representation from the concepts and attributes alone, \emph{so successful reconstruction means that the discrete representation preserves the backbone's information.}
We evaluate \ours{} across seven datasets, from general recognition (ImageNet, PartImageNet, Places) to fine-grained benchmarks (CUB, Cars, Dogs, Flowers). We show that \ours{} closely matches its frozen DINOv2 teacher on ImageNet linear probing \textbf{(83.2\% top-1)}, achieves higher concept consistency than language-aligned models, remains competitive on fine-grained recognition, and enables cross-category part-based retrieval.
\end{abstract}

\vspace{-1mm}%
\section{Introduction}
\label{sec:introduction}
\vspace{-1.5mm}%
\providecommand{\sidblue}{\iftoggle{comments}{\color[HTML]{1F5FBF}}{}}

Self-supervised vision transformers produce rich feature representations, serving as versatile foundation backbones for classification, segmentation, and retrieval~\citep{caron2021emerging, oquab2023dinov2}. As these models enter safety-critical domains, %
strong task performance alone is not sufficient, and we also need to understand \textbf{\emph{what}} semantic concepts a model relies on and \textbf{\emph{where}} it finds them during inference~\citep{rudin2019stop, rudin2022interpretable}.
This need has motivated research on inherently interpretable architectures. However, current paradigms make varying trade-offs in spatial grounding, concept representation, and supervision. Prototype-based networks~\citep{chen2019looks, janusz2026protoquant, nauta2023pip, rymarczyk2022interpretable} connect visual concepts to fixed class labels and usually require fine-tuning the entire backbone. Concept Bottleneck Models (CBMs)~\citep{koh2020concept, oikarinen2023label, yang2023language, yuksekgonul2022post, rao2024discover} often rely on human-specified or language-derived concepts, but lack localized spatial grounding~\citep{prasse2024dcbm}. Recent variants such as CFM~\citep{wittenmayer2026cfm} introduce spatial grounding, yet remain reliant on language-aligned backbones. Sparse autoencoders (SAEs) have been applied to the patch tokens of vision models~\citep{lim2025sparse, zaigrajew2025interpreting}. They extract features without concept annotations and provide patch-wise spatial attributions, but their continuous, potentially overlapping activations do not assign exactly one concept to each location. Moreover, these methods capture which concepts appear and where, but none encodes \textbf{\emph{how}} each concept varies across images (\emph{e.g.,} the type of a ``wheel'') as a separate attribute. Together, these methods leave open the combination of (1) spatial grounding, (2) one concept assignment per location, (3) attributes that capture how each concept varies across images, and (4) no class labels or language supervision.

\textit{We ask an essential question: Can a representation with all four properties be learned entirely from frozen, vision-only self-supervised features, without sacrificing the expressive power of the underlying backbone?}

\begin{figure}[t]
  \centering
  \includegraphics[width=\linewidth]{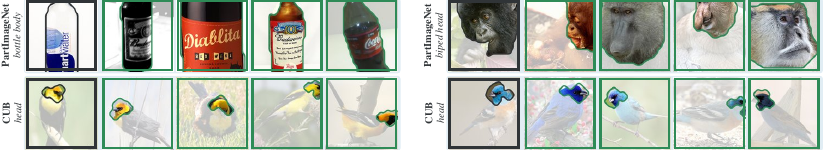}%
  \vspace{-2.3mm}%
  \caption{\textbf{Concepts retrieve the same part across categories.} A query region (black frame) and the four regions of other images with the closest \ours{} concept description (green frame).%
  }
  \label{fig:intro-retrieval}
  \vspace{-3.5mm}%
\end{figure}

To answer this, we introduce \textbf{\ours{}} (\textbf{Dis}crete \textbf{Par}ts with \textbf{Q}uantized attributes)\footnote{Concept visualizations: %
\url{https://disparq.gmum.net}}, a framework that discovers spatially grounded visual concepts from frozen self-supervised representations, assigns exactly one concept to each patch, and captures how each concept varies across images with an attribute, without class labels or language supervision. \ours{} maintains a learnable dictionary of prototype vectors, each representing a global concept. Every patch is assigned to one concept, and only a sparse subset of concepts can activate per image. To preserve fine appearance differences across occurrences of the same concept, we learn a separate attribute codebook per concept. Training proceeds in two stages: we first learn the high-level concepts with continuous residuals, then quantize these residuals into discrete attributes, resulting in a fully discrete bottleneck at inference.

Figure~\ref{fig:teaser} illustrates how \ours{} describes images with spatially grounded part concepts and explains a class prediction. Although only a small set of concepts %
(at most 16) are active per image, this description costs almost no accuracy: \ours{} matches its frozen backbone on ImageNet~\citep{deng2009imagenet} linear probing ($83.2\%$ vs.\ $83.4\%$, Figure~\ref{fig:classification-user-study}) and stays within $0$-$2.2$ p.p. of it on four fine-grained datasets (Table~\ref{tab:fine-grained}). Among self-supervised methods, it ranks first on most part-quality metrics on PartImageNet (PIN)~\citep{he2022partimagenet}, and its concepts are more consistent across images than language-grounded ones (Table~\ref{tab:pin-concepts}). We further confirm this in a user study (Section~\ref{sec:qualitative}).

We also introduce a part-to-part retrieval benchmark, which tests whether a concept names the same part across categories: given a part region, retrieve the same part in other images %
(Figure~\ref{fig:intro-retrieval}). \ours{} consistently leads self-supervised baselines (Tables~\ref{tab:part-retrieval} and~\ref{tab:retrieval-full}) on PIN and CUB~\citep{wah2011caltech}. %

To summarize, our \textbf{contributions} are as follows. \textbf{(C1)}~We propose \textbf{\ours{}}, a framework that learns part concepts with all four properties on a frozen self-supervised vision foundation backbone: spatial grounding, one concept per patch, attributes that capture how each concept varies across images, and no class labels or language supervision required. \textbf{(C2)}~We introduce a part-to-part retrieval benchmark on PartImageNet and CUB that tests whether a concept names the same part across images and categories. %
\textbf{(C3)}~We study the design of \ours{} on six frozen backbones and across feature extraction, concept budgets and selection, appearance quantization, decoder depth, gradient routes, and training objectives.

We make the following \textbf{empirical findings}, detailed in Section~\ref{sec:results}. \textbf{(F1)~A discrete bottleneck costs almost no accuracy.} With only a small set of active concepts ($k\leq16$) per image, \ours{} stays within $0.2$~p.p.\ of its teacher on ImageNet and within $2.2$~p.p.\ on four fine-grained datasets, using one configuration for all of them (Figure~\ref{fig:classification-user-study} and Table~\ref{tab:fine-grained}). \textbf{(F2)~Vision-only self-supervision grounds a consistent part vocabulary.} Without any language supervision, the concepts of \ours{} are more consistent across images than language-grounded ones, and \ours{} leads self-supervised baselines on most part-quality metrics (Tables~\ref{tab:pin-concepts} and~\ref{tab:fine-grained}). \textbf{(F3)~Concepts preserve their identity across categories.} In part-to-part retrieval, \ours{} leads the self-supervised baselines on most metrics, by up to \textbf{3.5} mAP on PartImageNet and \textbf{5.9}~p.p.\ R@1 on CUB (Tables~\ref{tab:part-retrieval} and~\ref{tab:retrieval-full}).

With \textbf{\ours{}}, we show that frozen vision foundation models can be interpreted through discrete parts and their attributes, without any human supervision and at almost no cost in accuracy.

\vspace{-1mm}%
\section{Related work}
\vspace{-1.5mm}%
\providecommand{\sidblue}{\iftoggle{comments}{\color[HTML]{1F5FBF}}{}}

\textbf{Prototype models} classify images by matching patches to learned exemplars~\citep{chen2019looks, nauta2023pip}, with newer ones using a VFM and quantized prototypes but still classifying through continuous similarity scores with class supervision~\citep{janusz2026protoquant}. \textbf{Part discovery models} instead learn a small, fixed set of parts, either without labels~\citep{hung2019scops, amir2021deep} or from class supervision~\citep{van2023pdisconet, aniraj2024pdiscoformer}. \ours{} learns a large shared dictionary without class labels or a task-specific classifier, letting each image select a few concepts to describe its parts.
\textbf{Concept bottleneck} models predict through human-defined or language-generated concepts~\citep{koh2020concept, oikarinen2023label}, including spatial concept maps~\citep{benou2025show}. Sparse autoencoders (SAEs) learn visual dictionaries from model activations~\citep{lim2025sparse, zaigrajew2025interpreting}, and CFM grounds concepts spatially and names them through a text vocabulary~\citep{wittenmayer2026cfm}. These SAE dictionaries use continuous, overlapping codes that entangle concept identity with appearance. \ours{} learns concepts from visual features alone, assigns each patch exactly one concept, and encodes concept identity and within-concept attributes separately using integer indices alone (Table~\ref{tab:related-properties}).
\textbf{Existing evaluations} measure prototype purity, part alignment, or concept locality and consistency~\citep{nauta2023pip, aniraj2024pdiscoformer, wittenmayer2026cfm}. Our part-to-part retrieval benchmark tests whether concept descriptions preserve part identity across images and categories by retrieving the same part from all other images (Section~\ref{sec:setup}). Appendix~\ref{app:related-work} gives the full related-work discussion.
\begin{table}[t]
  \centering
  \definecolor{propyes}{HTML}{2E7D32}\definecolor{proppart}{HTML}{B7791F}\definecolor{propno}{HTML}{B4B4B4}%
  \providecommand{\cmark}{\textcolor{propyes}{\checkmark}}%
  \providecommand{\pmark}{\textcolor{proppart}{(\checkmark)}}%
  \providecommand{\xmark}{\textcolor{propno}{$\times$}}%
  \providecommand{\hd}[1]{\shortstack{#1}}%
  \caption{\textbf{Properties of interpretable concept representations.} \ours{} is the only method with all nine: it states \emph{which} part concepts occur, \emph{where} they are and \emph{how} each one varies across images, without any labels. (\cmark~yes, \pmark~partially, \xmark~no)
  }
  \label{tab:related-properties}
  \scriptsize
  \setlength{\tabcolsep}{2.5pt}
  \renewcommand{\arraystretch}{1.0}
  \begin{tabular}{@{}l ccc cc cc cc@{}}
    \toprule
    & \multicolumn{3}{c}{\textit{Training}} & \multicolumn{2}{c}{\textit{Where}}
    & \multicolumn{2}{c}{\textit{Which \& how}} & \multicolumn{2}{c}{\textit{Scope}} \\
    \cmidrule(lr){2-4}\cmidrule(lr){5-6}\cmidrule(lr){7-8}\cmidrule(l){9-10}
    Method & \hd{No class\\labels} & \hd{No\\text} & \hd{Frozen\\backbone}
           & \hd{Concept\\maps} & \hd{One per\\patch}
           & \hd{Fully\\discrete} & \hd{Within-concept\\attributes}
           & \hd{Task-\\agnostic} & \hd{Any\\backbone} \\
    \midrule
    ProtoPNet~\citep{chen2019looks}          & \xmark & \cmark & \xmark & \cmark & \xmark & \xmark & \xmark & \xmark & \xmark \\
    PIP-Net~\citep{nauta2023pip}             & \pmark & \cmark & \xmark & \cmark & \pmark & \xmark & \xmark & \xmark & \xmark \\
    ProtoQuant~\citep{janusz2026protoquant}  & \pmark & \cmark & \cmark & \cmark & \pmark & \pmark & \xmark & \xmark & \cmark \\
    \addlinespace[2pt]
    PDiscoFormer~\citep{aniraj2024pdiscoformer} & \xmark & \cmark & \pmark & \cmark & \pmark & \xmark & \pmark & \xmark & \xmark \\
    \addlinespace[2pt]
    LF-CBM~\citep{oikarinen2023label}        & \xmark & \xmark & \cmark & \xmark & \xmark & \xmark & \xmark & \xmark & \cmark \\
    SALF-CBM~\citep{benou2025show}           & \xmark & \xmark & \cmark & \cmark & \xmark & \xmark & \xmark & \xmark & \cmark \\
    \addlinespace[2pt]
    PatchSAE~\citep{lim2025sparse}           & \cmark & \cmark & \cmark & \cmark & \xmark & \xmark & \xmark & \cmark & \cmark \\
    MSAE~\citep{zaigrajew2025interpreting}   & \cmark & \cmark & \cmark & \cmark & \xmark & \xmark & \xmark & \cmark & \cmark \\
    CFM~\citep{wittenmayer2026cfm}           & \cmark & \cmark & \cmark & \cmark & \xmark & \xmark & \xmark & \cmark & \pmark \\
    \midrule
    \rowcolor{oursrow}
    \textbf{\ours{}} (ours)                  & \cmark & \cmark & \cmark & \cmark & \cmark & \cmark & \cmark & \cmark & \cmark \\
    \bottomrule
  \end{tabular}
  \vspace{-3.5mm}%
\end{table}

\vspace{-1mm}%
\section{\ours{}}
\label{sec:architecture}
\vspace{-1.5mm}%

\ours{} provides an interpretable, discrete, spatially grounded concept bottleneck on top of a Vision Foundation Model (VFM).
An overview of \ours{} is presented in Figure~\ref{fig:crisp-overview}. We first extract latent features (Section~\ref{subsec:feature-extractor}), and organize them into a small set of exclusive concept regions to ensure spatial grounding (Section~\ref{subsec:concept-assignment}). Then we encode each region's appearance to represent details of the concepts (Section~\ref{subsec:attribute-quant}). Lastly, a spatial decoder predicts the global representation from these concepts (Section~\ref{subsec:decoder}). Training transforms continuous appearance attributes to discrete codes (Section~\ref{subsec:training}). Additional details are provided in Appendix~\ref{app:method}.%

\begin{figure*}[t]
  \centering
  \begingroup
  \CRISPSetup
  \input{\RepoFigureRoot/figures/01_overview_body.tex}%
  \endgroup
  \vspace{-2mm}%
  \caption{\CRISPOverviewCaption}
  \label{fig:crisp-overview}
  \vspace{-3.5mm}%
\end{figure*}

\vspace{-1mm}%
\subsection{Feature extraction}
\label{subsec:feature-extractor}
\vspace{-1.5mm}%

We build on a pretrained, frozen VFM encoder $f_{\mathrm{VFM}}$. An image $x$ is passed through it, producing (1) encoded features (patch tokens) that serve as input to \ours{}, and (2) a global image representation $c(x)$ (typically the \texttt{cls} token), which is later used as a distillation target. These two outputs serve different purposes. Local features ground concepts in image regions, while the global target encourages their collective description to preserve the image's semantics.

As different details are encoded at different depths of the $f_{\mathrm{VFM}}$, we introduce a lightweight shared-query
attention module $f_{\mathrm{fuse}}$ that at each location combines features from different encoder blocks. Then, as the feature map is of low resolution, which results in low precision of concept activations, we upsample it with an AnyUp module $f_{\mathrm{up}}$~\citep{wimmer2026anyup}. AnyUp uses the original normalized image to guide feature upsampling, providing a finer grid for concept assignment. As a result we obtain spatial feature vectors $z_u$, where $u$ is the patch position in the upsampled feature grid.

\vspace{-1mm}%
\subsection{Concept identification and assignment}
\label{subsec:concept-assignment}
\vspace{-1.5mm}%

Concept identification proceeds in two steps: we select a small set of concepts from the shared bank for each image (Figure~\ref{fig:crisp-overview}), and then assign each spatial location to one of those concepts (Figure~\ref{fig:crisp-quantization}).

\myparagraph{Selecting concepts.}
The shared bank must accommodate diverse visual structures across images. At the same time, any single image should have a compact description. That is why we train a large bank of $N$ concept prototypes, from which we select a small subset of $K$ concepts per image ($N=1024$, $K=16$ by default). Each prototype $w_n$, with $n\in\{1,\ldots,N\}$, is a learned reference vector. We compare it with the upsampled spatial features $z_u$ to obtain a soft activation map $A_n$, and average that map to get concept scores $\chi_n$:
\begin{equation}
  \ell_{n,u}=-\|z_u-w_n\|_2^2,
  \qquad A_{n,u}=\operatorname{softmax}_{n}(\ell_{n,u}),
  \qquad \chi_n=\frac{1}{HW}\sum_u A_{n,u}.
  \label{eq:bank-competition}
\end{equation}
Here $H\times W$ is the upsampled feature grid size, and the softmax makes all $N$ prototypes compete at each location. During training, Sinkhorn--Knopp balancing~\citep{sinkhorn1967concerning} adjusts the scores $\chi_n$ across images in a batch to encourage balanced prototype usage. We select the $K$ concepts with the highest balanced scores for each image. The bank index $i_k\in\{1,\ldots,N\}$ identifies its $k$th selected concept, with $k\in\{1,\ldots,K\}$.

\myparagraph{Assigning locations.}
Having chosen \emph{which} $K$ concepts participate, we then determine \emph{where} they occur. We recompute the softmax over only the selected concepts, so spatial competition is restricted to the image's chosen subset. This produces the post-selection soft maps $\bar{A}_{i_k}$ and hard identity map $p$ shown in Figure~\ref{fig:crisp-quantization}:
\begin{equation}
  \bar{A}_{i_k,u}=\operatorname{softmax}_{k}(\ell_{i_k,u}),
  \qquad p_u=i_{\arg\max_k \bar{A}_{i_k,u}}.
  \label{eq:selected-activation}
\end{equation}
The strongest response at each location determines $p_u$, making the concept assignment mutually exclusive. Training and inference details are specified in Appendix~\ref{app:selection}.

\vspace{-1mm}%
\subsection{Attribute encoding and quantization}
\label{subsec:attribute-quant}
\vspace{-1.5mm}%

A concept such as a dog's nose can vary in appearance across images. Describing every appearance difference with a new prototype would fragment the concept bank, whereas ignoring those differences would discard information needed to match the VFM representation. We therefore learn a concept-specific \emph{attribute} that captures \emph{how} each occurrence differs from its concept reference.

The attribute pathway in Figure~\ref{fig:crisp-quantization} follows three steps: pooling region features, computing their deviation from the prototype, and quantizing that residual. For each selected concept $i_k$, the soft map $\bar{A}_{i_k}$ weights the features $z_u$ to form a region mean $\bar z_{i_k}$. We normalize both vectors with $\nu$ and subtract the prototype from the region mean to compute the residual $r_{i_k}$:
\begin{equation}
  \bar z_{i_k}=\frac{\sum_u \bar{A}_{i_k,u}z_u}{\sum_u \bar{A}_{i_k,u}},
  \qquad r_{i_k}=\nu(\bar z_{i_k})-\nu(w_{i_k}).
  \label{eq:appearance-residual}
\end{equation}
Here $\nu$ denotes $\ell_2$ normalization. The residual $r_{i_k}$ describes the whole region relative to its concept reference, tying appearance information to that concept occurrence. We encode the residual with product quantization (PQ)~\citep{jegou2011product} using concept-specific codebooks $\mathcal B_{i_k}$. The result is a discrete appearance code and its quantized vector $q_{i_k}$, known as an \emph{attribute}. Pooling and quantization details are given in Appendix~\ref{app:attributes}.

\begin{figure*}[t]
  \centering
  \begingroup
  \CRISPSetup
  \input{\RepoFigureRoot/figures/02_product_quantization_body.tex}%
  \endgroup
  \vspace{-2mm}%
  \caption{\CRISPQuantizationCaption}
  \label{fig:crisp-quantization}
  \vspace{-3.5mm}%
\end{figure*}

\vspace{-1mm}%
\subsection{Concept decoder}
\label{subsec:decoder}
\vspace{-1.5mm}%

The decoder maps the interpretable concept description to a global image representation for downstream tasks. It combines concept identities, spatial assignments, and appearance attributes to predict the frozen VFM's representation through feature distillation.

The right portion of Figure~\ref{fig:crisp-quantization} shows how each spatial token combines the assigned concept's embedding, its appearance attribute, and a positional embedding $\mathrm{PE}_u$. A concept embedding $e_n$ represents each concept in the decoder and is learned independently from the prototype used in previous steps. At each position $u\in\{1,\ldots,HW\}$, the identity $p_u$ selects the concept embedding and attribute:
\begin{equation}
  t_u=e_{p_u}+\mathrm{PE}_u+q_{p_u},\qquad
  \hat c(x)=f_{\mathrm{dec}}\bigl([t_{\mathrm{CLS}},t_1,\ldots,t_{HW}]\bigr).
  \label{eq:spatial-decoder}
\end{equation}
A small transformer model integrates these tokens and predicts the VFM's global representation from the output of its learned decoder CLS token $t_{\mathrm{CLS}}$ (separate from the VFM's CLS token).

Every image-dependent input to this decoder passes through the concept description. In the quantized model, the selected identities $i_{1:K}$, assignment map $p$, and discrete appearance codes completely specify that description. Decoding requires only these discrete quantities, learned lookup tables, and the decoder, making explicit what information survives the bottleneck. The output $\hat c(x)$ has the same dimension as the VFM representation and can be used for downstream tasks such as classification and retrieval.

\vspace{-1mm}%
\subsection{Training objectives and regime}
\label{subsec:training}
\vspace{-1.5mm}%

We train through feature distillation without class labels, text descriptions, or region annotations. For an augmented image $x$ and a sampled affine transformation $T$, each view $v\in\{x,T(x)\}$ predicts its corresponding frozen VFM target $c(v)$:
\begin{equation}
  \mathcal L_{\mathrm{global}}=
  \frac{1}{2}\sum_{v\in\{x,T(x)\}}
  \left[1-\operatorname{cos}\bigl(\hat c(v),\operatorname{stopgrad}(c(v))\bigr)\right],
  \label{eq:global-distillation}
\end{equation}
The loss is averaged over images, $\operatorname{cos}$ denotes cosine similarity and $\operatorname{stopgrad}$ stops gradients. Following PDiscoFormer~\citep{aniraj2024pdiscoformer}, we regularize assignments to encourage coherent, confident regions and distinct, active concepts. Affine equivariance encourages spatial responses to follow image content under geometric transformations. Appendix~\ref{app:objectives} specifies the losses (Figure~\ref{fig:crisp-training}) and weights (Table~\ref{tab:method-losses}). Appendix~\ref{app:train-ours} gives the augmentation and optimization recipe.

\myparagraph{Stage 1: discovering concepts.}
We jointly train feature fusion, concept prototypes, the continuous appearance projection, and the decoder. Continuous attributes provide flexibility during concept discovery, while an auxiliary distillation head encourages pooled region features to preserve global image information. Assignments are hard in the forward pass. \textbf{Stage 2: quantizing attributes.} We freeze feature fusion and the prototypes, replace continuous attributes with product-quantized vectors, and train the quantization module, concept embeddings, and decoder through $\mathcal L_{\mathrm{global}}$. The auxiliary head is disabled. The VFM and AnyUp remain frozen throughout both stages. Appendices~\ref{app:quantization},~\ref{app:gradients}, and~\ref{app:curricula} detail quantization, gradient handling, and the stage transition.

\providecommand{\sidblue}{\iftoggle{comments}{\color[HTML]{1F5FBF}}{}}

\vspace{-1mm}%
\section{Experimental evaluation}
\label{sec:experiments}
\vspace{-1.5mm}%

In this section, we compare \ours{} with concept dictionaries and prototype models. Section~\ref{sec:setup} describes the evaluation setup, Section~\ref{sec:results} the quantitative results, and Section~\ref{sec:qualitative} the qualitative results and how \ours{} explains a prediction. The implementation details for training and evaluating \ours{} as well as the baselines are in Appendices~\ref{app:curricula} and~\ref{app:protocol}.

\vspace{-1mm}%
\subsection{Experimental setup}
\label{sec:setup}
\vspace{-1.5mm}%

\myparagraph{Datasets.} We train one \ours{} model per dataset, without labels. ImageNet-1k~\citep{deng2009imagenet} and Places365~\citep{zhou2017places} test classification at scale. PartImageNet~\citep{he2022partimagenet}, with masks for $40$ part classes, and CUB-200-2011~\citep{wah2011caltech}, with $15$ keypoints per bird, test the part concepts. Stanford Cars~\citep{krause2013collecting}, Stanford Dogs~\citep{khosla2011novel} and Oxford Flowers~\citep{nilsback2008automated} test fine-grained classification (Appendix~\ref{app:datasets}).

\myparagraph{Backbone and baselines.} The main self-supervised comparison uses frozen DINOv2-B/14 with registers~\citep{oquab2023dinov2,darcet2024registers}, except CFM~\citep{wittenmayer2026cfm}, whose released dictionary is built on CLIP-DINOiser~\citep{wysoczanska2024clipdinoiser}. We compare \ours{} to (i)~sparse autoencoders trained on the same patch tokens: TopK SAE~\citep{gao2025scaling}, PatchSAE~\citep{lim2025sparse} and MSAE~\citep{zaigrajew2025interpreting}, (ii)~the concept dictionary of CFM, and (iii)~the label-free first stage of the prototype models PIP-Net~\citep{nauta2023pip} and ProtoQuant~\citep{janusz2026protoquant}. As unranked references, gray rows give PDiscoFormer~\citep{aniraj2024pdiscoformer} and the classifier stage of PIP-Net and ProtoQuant, which use class labels, and the frozen backbone without a bottleneck. For spatial evaluation and part retrieval, the main self-supervised comparison uses $K=16$ concepts per image, with \ours{} selecting them from $N=1024$. TopK SAE is trained with this budget. PatchSAE, MSAE and CFM keep their $16$ strongest concepts for these evaluations, but use unbudgeted representations for classification and image retrieval. Appendix~\ref{app:protocol} states selection rules and protocol exceptions. Appendix~\ref{app:pin-full} discusses more settings. %

\myparagraph{Classification.} We evaluate the representation that each method passes downstream, which for \ours{} is the CLS token its decoder predicts from the discrete description. For the baselines, the probes read the pooled reconstruction of the sparse autoencoders, the pooled concept activations of CFM, and the prototype scores of PIP-Net and ProtoQuant (see Appendix~\ref{app:pin-protocol}). We report the top-1 accuracy of a linear probe (LP) and of a weighted $20$-nearest-neighbor classifier (kNN).

\myparagraph{Part concepts.} On PartImageNet, we compare concept maps with part masks: foreground purity (Pur$_\mathrm{fg}$), NMI and ARI after mapping each concept to its dominant part (NMI$_m$, ARI$_m$)~\citep{van2023pdisconet,aniraj2024pdiscoformer}, and the mIoU~\citep{he2022partimagenet} and FG-ARI~\citep{locatello2020objectcentric} of the resulting part segmentation. We add the locality, consistency and impurity of CFM~\citep{wittenmayer2026cfm}, which ask whether a concept covers one part and whether it covers the same part in every image. On CUB, we report PIP purity~\citep{nauta2023pip} and the normalized mean error of keypoint regression (Kp NME)~\citep{aniraj2024pdiscoformer}. %
Appendix~\ref{app:metrics} defines all metrics in detail.

\myparagraph{Part-to-part retrieval.} We introduce a benchmark that asks whether a concept names the same part across images and categories. Each annotated part region, a mask on PartImageNet or a visible keypoint on CUB, is described by the histogram of the concepts it contains, and the regions of all other images are ranked by cosine similarity. Appendix~\ref{app:metrics} gives the details on the benchmark.

\vspace{-1mm}%
\subsection{Quantitative results}
\label{sec:results}
\vspace{-1.5mm}%

We first test whether the discrete bottleneck keeps the backbone's accuracy (F1), then the quality of its part concepts (F2), and finally whether a concept keeps its identity across categories (F3).

\begin{figure}[t]
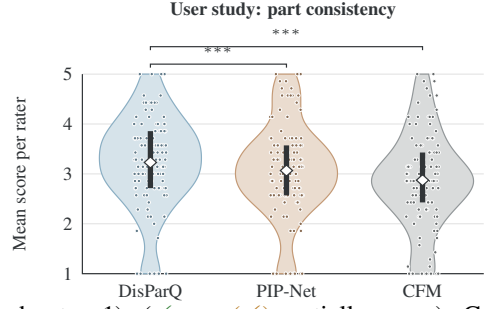

  \centering
  \definecolor{propyes}{HTML}{2E7D32}\definecolor{proppart}{HTML}{B7791F}\definecolor{propno}{HTML}{B4B4B4}%
  \providecommand{\cmark}{\textcolor{propyes}{\checkmark}}%
  \providecommand{\pmark}{\textcolor{proppart}{(\checkmark)}}%
  \providecommand{\xmark}{\textcolor{propno}{$\times$}}%
  \providecommand{\hd}[1]{\shortstack{#1}}%
  \begin{minipage}[c]{0.49\textwidth}
    \centering
      \definecolor{propyes}{HTML}{2E7D32}\definecolor{proppart}{HTML}{B7791F}\definecolor{propno}{HTML}{B4B4B4}%
  \providecommand{\cmark}{\textcolor{propyes}{\checkmark}}%
  \providecommand{\pmark}{\textcolor{proppart}{(\checkmark)}}%
  \providecommand{\xmark}{\textcolor{propno}{$\times$}}%
  \providecommand{\suprow}[1]{\textcolor{black!65}{#1}}%
  \scriptsize
  \setlength{\tabcolsep}{2pt}
  \renewcommand{\arraystretch}{1.0}
  \begin{tabular}{@{}lcccc@{}}
    \toprule
    Method & \shortstack{Spatial\\grounding} & \shortstack{Discrete\\bottleneck} & IN-1k\,$\uparrow$ & Places\,$\uparrow$ \\
    \midrule
    \suprow{ViT-B/16, supervised} &  \suprow{N/A} &  \suprow{N/A} & \suprow{81.1} & \suprow{\textemdash} \\
    \suprow{ProtoQuant} & \suprow{(\checkmark)} & \suprow{(\checkmark)} & \suprow{80.1} & \suprow{\textemdash} \\
    \midrule
    \suprow{CLIP ViT-B/16} &  \suprow{N/A} &  \suprow{N/A} & \suprow{80.2} & \suprow{55.1} \\
    DN-CBM & \xmark & \xmark & \underline{79.5} & 55.1 \\
    CDM & \xmark & \pmark & 79.3 & 52.6 \\
    CFM & \pmark & \xmark & 78.9 & \textbf{55.4} \\
    \midrule
    \suprow{DINOv2-B/14} &  \suprow{N/A} &  \suprow{N/A} & \suprow{83.4} & \suprow{55.4} \\
    \rowcolor{oursrow} \textbf{\ours{}} (ours) & \cmark & \cmark & \textbf{83.2} & \underline{55.3} \\
    \bottomrule
  \end{tabular}

  \end{minipage}\hfill
  \begin{minipage}[c]{0.49\textwidth}
    \centering
    \begingroup
\definecolor{ink}{HTML}{303437}
\definecolor{studyblue}{HTML}{8AAEC3}%
\definecolor{studypeach}{HTML}{C49A76}%
\definecolor{studygray}{HTML}{929799}
\begin{tikzpicture}[x=18mm,y=6.6mm,
  every node/.style={font=\scriptsize,text=ink,inner sep=1pt}]
  \node[font=\scriptsize\bfseries,anchor=south] at (.975,6.05) {User study: part consistency};
  \foreach \y in {1,...,5} {
    \draw[black!12,line width=.3pt] (-.5,\y) -- (2.45,\y);
    \node[anchor=east] at (-.55,\y) {\y};
  }
  \draw[ink,line width=.45pt] (-.5,5) -- (-.5,1) -- (2.45,1);
  \node[rotate=90] at (-.95,3) {Mean score per rater};
  \input{\RepoFigureRoot/figures/user_study/violin_data.tex}
  \foreach \x/\method in {0/\ours{},1/PIP-Net,2/CFM}
    \node[anchor=north] at (\x,.85) {\method};
  \input{\RepoFigureRoot/figures/user_study/violin_stats.tex}
\end{tikzpicture}
\endgroup
  \end{minipage}
    \vspace{-3.5mm}%

  \caption{\textit{Left:} Large-scale classification (linear probe, top-1). (\cmark~yes, \pmark~partially, \xmark~no). Gray rows are unranked references. All methods and their sources are in Table~\ref{tab:classification-full} in Appendix~\ref{app:classification-full}. \textit{Right:} \ours{} achieves the highest perceived part consistency in our user study. Diamonds mark means and bars show interquartile ranges. $^{***}p<0.001$, paired Wilcoxon, Holm-corrected.}
  \label{fig:classification-user-study}
  \label{fig:user_study}
\end{figure}

\myparagraph{Classification.} The discrete description costs almost no accuracy, although only concept identities, regions and quantized attributes reach the decoder. On ImageNet-1k, its linear probe reaches \textbf{83.2\%}, the highest of all concept models in Figure~\ref{fig:classification-user-study} (left) (\textbf{+3.7}~p.p.\ over DN-CBM) and only $0.2$~p.p.\ below its DINOv2 backbone, while the vision-language concept models lose $0.7$ to $1.3$~p.p.\ to their CLIP backbone. On Places365, it reaches $55.3\%$, $0.1$~p.p.\ below both its backbone and CFM. Against the self-supervised baselines of Tables~\ref{tab:pin-concepts} and~\ref{tab:fine-grained}, whose best rows use the same DINOv2 backbone, its LP is \textbf{3.6}~p.p.\ higher on PartImageNet and \textbf{4.0} to \textbf{9.3}~p.p.\ higher on CUB, Cars and Dogs. None of the other methods has concepts that are both exclusive and discrete (Appendix~\ref{app:classification-full}). Quantized attributes retain appearance lost by concepts alone; removing them costs 12.2 p.p. in LP (Appendix~\ref{abl:quantization}). %

\myparagraph{Part concepts on PartImageNet.} Among self-supervised methods, \ours{} is best on \textbf{7 of the 10} metrics of Table~\ref{tab:pin-concepts} and second on two more, Pur$_\mathrm{fg}$ and FG-ARI, where the label-free stage of ProtoQuant leads. Its largest margins are on ARI$_m$ (\textbf{62.3} vs.\ $56.3$), locality (\textbf{36.4} vs.\ $30.5$) and consistency, where its concepts are more consistent than those of the language-grounded CFM (\textbf{16.3} vs.\ $12.4$). Its kNN accuracy is \textbf{4.8}~p.p.\ above the best baseline, within $0.2$~p.p.\ of the frozen backbone. It also exceeds the supervised PDiscoFormer on every part and grounding metric except FG-ARI.

\begin{table*}[t]
  \centering
  \gdef\PinTopKSize{2{,}048}\gdef\PinPatchSAESize{49{,}152}\gdef\PinMSAESize{8{,}192}\gdef\PinMSAEWeighting{reverse}%
  \definecolor{deltapos}{HTML}{1F5FBF}\definecolor{deltaneg}{HTML}{C62828}%
  \providecommand{\dpos}[1]{\textsubscript{\textcolor{deltapos}{\textbf{#1}}}}%
  \providecommand{\dneg}[1]{\textsubscript{\textcolor{deltaneg}{\textbf{#1}}}}%
  \caption{\textbf{Part concepts on PartImageNet} (val, as $\%$ except Imp.). \textbf{Best} and \underline{second best} among self-supervised methods. Subscripts: gap to the best. Gray: unranked references. $K/N$: concepts per image/dictionary size. Protocol: Appendix~\ref{app:pin-protocol}. \ours{} shows superior performance on 7/10.}
  \label{tab:pin-concepts}
  \footnotesize
  \providecommand{\suprow}[1]{\textcolor{black!65}{#1}}%
  \setlength{\tabcolsep}{3pt}
  \renewcommand{\arraystretch}{1.08}
  \resizebox{\textwidth}{!}{%
  \begin{tabular}{@{}lllcccccccccc@{}}
    \toprule
    & & & \multicolumn{2}{c}{Downstream} & \multicolumn{5}{c}{Part quality} & \multicolumn{3}{c}{Concept grounding} \\
    \cmidrule(lr){4-5}\cmidrule(lr){6-10}\cmidrule(l){11-13}
    Method & Backbone & $K/N$ & LP\,$\uparrow$ & kNN\,$\uparrow$ & Pur$_{\mathrm{fg}}$\,$\uparrow$ & NMI$_m$\,$\uparrow$ & ARI$_m$\,$\uparrow$ & mIoU\,$\uparrow$ & FG-ARI\,$\uparrow$ & Loc.\,$\uparrow$ & Cons.\,$\uparrow$ & Imp.\,$\downarrow$ \\
    \midrule
    \multicolumn{13}{@{}l}{\suprow{\textit{Supervised (reference)}}} \\
    \suprow{PDiscoFormer} & \suprow{DINOv2-B/14} & \suprow{50/50} & \suprow{89.8} & \suprow{88.1} & \suprow{54.9} & \suprow{48.4} & \suprow{37.9} & \suprow{29.1} & \suprow{28.8} & \suprow{34.9} & \suprow{2.0} & \suprow{0.505} \\
    \suprow{PIP-Net\textsubscript{stage 2}} & \suprow{DINOv2-B/14} & \suprow{16/768} & \suprow{87.5} & \suprow{86.8} & \suprow{59.3} & \suprow{37.3} & \suprow{11.5} & \suprow{24.9} & \suprow{18.9} & \suprow{25.2} & \suprow{6.4} & \suprow{0.437} \\
    \suprow{ProtoQuant\textsubscript{stage 2}} & \suprow{DINOv2-B/14} & \suprow{16/2048} & \suprow{87.8} & \suprow{89.5} & \suprow{71.7} & \suprow{55.6} & \suprow{30.0} & \suprow{26.5} & \suprow{19.5} & \suprow{13.1} & \suprow{4.8} & \suprow{0.513} \\
    \addlinespace
    \midrule
    \multicolumn{13}{@{}l}{\suprow{\textit{Frozen backbone, no concept bottleneck (reference)}}} \\
    \suprow{CLS token} & \suprow{DINOv2-B/14} & \suprow{N/A} & \suprow{89.3} & \suprow{88.9} & \suprow{N/A} & \suprow{N/A} & \suprow{N/A} & \suprow{N/A} & \suprow{N/A} & \suprow{N/A} & \suprow{N/A} & \suprow{N/A} \\
    \suprow{Patch tokens, mean-pooled} & \suprow{DINOv2-B/14} & \suprow{N/A} & \suprow{85.7} & \suprow{79.3} & \suprow{N/A} & \suprow{N/A} & \suprow{N/A} & \suprow{N/A} & \suprow{N/A} & \suprow{N/A} & \suprow{N/A} & \suprow{N/A} \\
    \addlinespace
    \midrule
    \multicolumn{13}{@{}l}{\textit{Self-supervised}} \\
    CFM & CLIP-DI-B/16 & 16/8192 & 72.6 & 69.2 & 54.7 & 51.4 & 24.5 & 22.7 & 12.5 & 22.9 & \underline{12.4} & \underline{0.378} \\
    TopK SAE & DINOv2-B/14 & 16/2048 & 80.5 & 73.5 & 68.8 & 58.0 & 36.0 & 24.8 & 16.1 & 20.6 & 9.8 & 0.462 \\
    PatchSAE & DINOv2-B/14 & 16/49152 & 84.1 & 77.2 & 76.2 & 60.1 & 34.9 & 25.0 & 17.0 & 24.7 & 4.4 & 0.455 \\
    MSAE & DINOv2-B/14 & 16/8192 & \underline{85.6} & 79.5 & 59.0 & 50.3 & 23.5 & 20.5 & 12.1 & 18.0 & 8.1 & 0.460 \\
    PIP-Net\textsubscript{stage 1} & DINOv2-B/14 & 16/768 & 76.9 & 71.5 & 68.5 & 54.2 & 34.3 & 29.0 & 19.1 & \underline{30.5} & 10.2 & \textbf{0.376} \\
    ProtoQuant\textsubscript{stage 1} & DINOv2-B/14 & 16/2048 & 85.2 & \underline{83.9} & \textbf{82.1} & \underline{68.7} & \underline{56.3} & \underline{33.7} & \textbf{28.4} & 18.1 & 7.3 & 0.514 \\
    \addlinespace
    \midrule
    \rowcolor{oursrow} \textbf{\ours{}} (ours) & DINOv2-B/14 & 16/1024 & \textbf{89.2}\dpos{+3.6} & \textbf{88.7}\dpos{+4.8} & \underline{81.5}\dneg{-0.6} & \textbf{69.8}\dpos{+1.1} & \textbf{62.3}\dpos{+6.0} & \textbf{35.9}\dpos{+2.2} & \underline{26.3}\dneg{-2.1} & \textbf{36.4}\dpos{+5.9} & \textbf{16.3}\dpos{+3.9} & 0.385\dneg{+0.009} \\
    \bottomrule
  \end{tabular}}
  \vspace{-3.5mm}%
\end{table*}

\myparagraph{Fine-grained datasets.} On the four fine-grained datasets, \ours{} is best on \textbf{all 10} metrics of Table~\ref{tab:fine-grained} among self-supervised methods (tied on Flowers LP). Its kNN accuracy is \textbf{16.7} to \textbf{19.8}~p.p.\ above the best baseline on CUB, Cars and Dogs. On CUB, its concepts also localize the keypoints best, with PIP purity \textbf{54.5} vs.\ $37.3$ and keypoint error \textbf{9.36} vs.\ $9.87$ (Appendix~\ref{app:metrics} explains how the purity window is placed on exclusive maps). The supervised PDiscoFormer\footnote{Checkpoints released only for CUB and Flowers; see Appendix~\ref{app:pdisco-repro} for our reproduction on Cars and Dogs.} scores higher on the CUB clustering metrics and on keypoint error, and lower on PIP purity (Table~\ref{tab:cub-concepts} in Appendix~\ref{app:cub-full}).

\begin{table*}[t]
  \centering
  \definecolor{deltapos}{HTML}{1F5FBF}\definecolor{deltaneg}{HTML}{C62828}%
  \providecommand{\dpos}[1]{\textsubscript{\textcolor{deltapos}{\textbf{#1}}}}%
  \providecommand{\dneg}[1]{\textsubscript{\textcolor{deltaneg}{\textbf{#1}}}}%
  \providecommand{\suprow}[1]{\textcolor{black!65}{#1}}%
  \caption{\textbf{Fine-grained datasets} (test, as $\%$ except Kp NME; CFM on CLIP-DINOiser). Marks as in Table~\ref{tab:pin-concepts}.
  For PDiscoFormer some scores are reproduced as in Appendix~\ref{app:pdisco-repro}. $^{\ast}$4000 on CUB. Protocol: Appendix~\ref{app:finegrained-protocol}. \ours{} consistently outperforms baseline methods on all but one metric.}
  \label{tab:fine-grained}
  \footnotesize
  \setlength{\tabcolsep}{3.5pt}
  \renewcommand{\arraystretch}{1.0}%
  \resizebox{0.88\textwidth}{!}{%
  \begin{tabular}{@{}llcccccccccc@{}}
    \toprule
    & & \multicolumn{4}{c}{CUB-200-2011} & \multicolumn{2}{c}{Stanford Cars} & \multicolumn{2}{c}{Stanford Dogs} & \multicolumn{2}{c}{Oxford Flowers} \\
    \cmidrule(lr){3-6}\cmidrule(lr){7-8}\cmidrule(lr){9-10}\cmidrule(lr){11-12}
    Method & $K/N$ & LP\,$\uparrow$ & kNN\,$\uparrow$ & PIP pur.\,$\uparrow$ & Kp NME\,$\downarrow$ & LP\,$\uparrow$ & kNN\,$\uparrow$ & LP\,$\uparrow$ & kNN\,$\uparrow$ & LP\,$\uparrow$ & kNN\,$\uparrow$ \\
    \midrule
    \multicolumn{12}{@{}l}{\suprow{\textit{Supervised (reference)}}} \\
    \suprow{PDiscoFormer} & \suprow{$K/K$} & \suprow{86.8} & \suprow{74.8} & \suprow{40.4} & \suprow{2.87} & \suprow{N/A} & \suprow{N/A} & \suprow{N/A} & \suprow{N/A} & \suprow{99.6} & \suprow{99.7} \\
    \suprow{PIP-Net\textsubscript{stage 2}} & \suprow{16/768} & \suprow{88.9} & \suprow{88.9} & \suprow{47.0} & \suprow{10.16} & \suprow{91.4} & \suprow{87.3} & \suprow{87.1} & \suprow{85.2} & \suprow{99.6} & \suprow{99.7} \\
    \suprow{ProtoQuant\textsubscript{stage 2}} & \suprow{16/2048$^{\ast}$} & \suprow{84.9} & \suprow{83.0} & \suprow{25.2} & \suprow{12.16} & \suprow{91.3} & \suprow{86.3} & \suprow{86.0} & \suprow{79.4} & \suprow{99.6} & \suprow{99.6} \\
    \midrule
    \multicolumn{12}{@{}l}{\suprow{\textit{Frozen backbone, no concept bottleneck (reference)}}} \\
    \suprow{CLS token} & \suprow{N/A} & \suprow{90.4} & \suprow{86.7} & \suprow{N/A} & \suprow{N/A} & \suprow{93.2} & \suprow{85.6} & \suprow{88.6} & \suprow{86.8} & \suprow{99.7} & \suprow{99.7} \\
    \midrule
    \multicolumn{12}{@{}l}{\textit{Self-supervised}} \\
    CFM & 16/8192 & 54.8 & 42.8 & 16.1 & 12.84 & 71.0 & 61.3 & 56.0 & 52.5 & 90.2 & 88.8 \\
    TopK SAE & 16/2048 & 36.9 & 24.3 & 8.7 & 10.92 & 41.9 & 22.7 & 61.3 & 41.1 & 99.4 & 99.0 \\
    PatchSAE & 16/49152 & 62.0 & 33.4 & 6.9 & 19.12 & 74.6 & 35.8 & 77.6 & 53.5 & 99.3 & 99.0 \\
    MSAE & 16/6144 & 73.5 & 42.2 & 15.1 & 12.21 & 80.3 & 41.7 & 79.4 & 55.2 & \underline{99.6} & 99.4 \\
    PIP-Net\textsubscript{stage 1} & 16/768 & 77.8 & \underline{68.4} & \underline{37.3} & \underline{9.87} & 82.3 & \underline{63.6} & 77.8 & \underline{69.1} & 99.1 & 98.4 \\
    ProtoQuant\textsubscript{stage 1} & 16/2048$^{\ast}$ & \underline{80.1} & 67.7 & 24.6 & 11.59 & \underline{87.0} & 60.2 & \underline{82.2} & 66.9 & \textbf{99.7} & \underline{99.5} \\
    \midrule
    \rowcolor{oursrow} \textbf{\ours{}} (ours) & 16/1024 & \textbf{89.4}\dpos{+9.3} & \textbf{85.5}\dpos{+17.1} & \textbf{54.5}\dpos{+17.2} & \textbf{9.36}\dpos{-0.51} & \textbf{91.0}\dpos{+4.0} & \textbf{83.4}\dpos{+19.8} & \textbf{87.8}\dpos{+5.6} & \textbf{85.8}\dpos{+16.7} & \textbf{99.7} & \textbf{99.7}\dpos{+0.2} \\
    \bottomrule
  \end{tabular}}
  \vspace{-3.5mm}%
\end{table*}

\myparagraph{Cross-category part retrieval.} The concepts of \ours{} keep their identity across categories. Among self-supervised methods, \ours{} is best on \textbf{7 of the 8} metrics of Table~\ref{tab:part-retrieval} and second on part-to-part R@5. On PartImageNet, it leads part-to-part retrieval with an R@1 of \textbf{79.7\%} and an mAP of \textbf{30.5}, vs.\ $79.2\%$ and $27.0$ for the best baselines, and image-to-image retrieval with an mAP of \textbf{76.1} vs.\ $61.8$. An oracle based only on part-label co-occurrence groups reaches an R@1 of $43.9\%$ (Appendix~\ref{app:retrieval-full}). Against the supervised PDiscoFormer, \ours{} leads \textbf{4 of the 6} retrieval metrics at each of its three part counts, including part-to-part R@1 (\textbf{79.7\%} vs.\ at most $71.8\%$) and image-to-image mAP (\textbf{76.1} vs.\ at most $70.2$), while PDiscoFormer reaches a higher part-to-part mAP. On CUB, \ours{} also leads R@1 (\textbf{36.8\%} vs.\ $30.9\%$; see Table~\ref{tab:retrieval-full} in Appendix~\ref{app:retrieval-full}).

\begin{table*}[t]
  \centering
  \definecolor{deltapos}{HTML}{1F5FBF}\definecolor{deltaneg}{HTML}{C62828}%
  \providecommand{\dpos}[1]{\textsubscript{\textcolor{deltapos}{\textbf{#1}}}}%
  \providecommand{\dneg}[1]{\textsubscript{\textcolor{deltaneg}{\textbf{#1}}}}%
  \providecommand{\suprow}[1]{\textcolor{black!65}{#1}}%
  \caption{\textbf{Retrieval on PartImageNet} (val, as $\%$). Marks as in Table~\ref{tab:pin-concepts}. Dictionary methods appear at our $N{=}1024$ and their best $N$. $^{\ddagger}$At the method's native resolution, on another query set (Appendix~\ref{app:retrieval-full}). The controls and PCA are in Table~\ref{tab:retrieval-full}. \ours{} is the best on all but one metric.}
  \label{tab:part-retrieval}
  \footnotesize
  \setlength{\tabcolsep}{3pt}
  \renewcommand{\arraystretch}{1.0}%
  \resizebox{0.76\textwidth}{!}{%
  \begin{tabular}{@{}llcccccccc@{}}
    \toprule
    & & \multicolumn{2}{c}{Accuracy} & \multicolumn{3}{c}{Image $\to$ image} & \multicolumn{3}{c}{Part $\to$ part} \\
    \cmidrule(lr){3-4}\cmidrule(lr){5-7}\cmidrule(lr){8-10}
    Method & $K/N$ & LP\,$\uparrow$ & kNN\,$\uparrow$ & R@1\,$\uparrow$ & R@5\,$\uparrow$ & mAP\,$\uparrow$ & R@1\,$\uparrow$ & R@5\,$\uparrow$ & mAP\,$\uparrow$ \\
    \midrule
    \multicolumn{10}{@{}l}{\suprow{\textit{Supervised (reference)}}} \\
    \suprow{PDiscoFormer$^{\ddagger}$} & \suprow{50/50} & \suprow{89.8} & \suprow{88.1} & \suprow{78.4} & \suprow{93.4} & \suprow{69.3} & \suprow{71.8} & \suprow{89.0} & \suprow{45.9} \\
    \suprow{PIP-Net\textsubscript{stage 2}} & \suprow{16/768} & \suprow{87.5} & \suprow{86.8} & \suprow{77.6} & \suprow{91.9} & \suprow{64.2} & \suprow{65.4} & \suprow{85.2} & \suprow{21.1} \\
    \suprow{ProtoQuant\textsubscript{stage 2}} & \suprow{16/2048} & \suprow{87.8} & \suprow{89.5} & \suprow{82.3} & \suprow{93.8} & \suprow{76.6} & \suprow{62.6} & \suprow{89.6} & \suprow{15.5} \\
    \midrule
    \multicolumn{10}{@{}l}{\suprow{\textit{Frozen backbone, no concept bottleneck (reference)}}} \\
    \suprow{CLS token} & \suprow{N/A} & \suprow{89.3} & \suprow{88.9} & \suprow{81.0} & \suprow{92.9} & \suprow{75.1} & \suprow{N/A} & \suprow{N/A} & \suprow{N/A} \\
    \suprow{Patch tokens, mean-pooled} & \suprow{N/A} & \suprow{85.7} & \suprow{79.3} & \suprow{57.5} & \suprow{82.2} & \suprow{44.7} & \suprow{N/A} & \suprow{N/A} & \suprow{N/A} \\
    \midrule
    \multicolumn{10}{@{}l}{\textit{Self-supervised}} \\
    CFM$^{\ddagger}$ & 16/8192 & 72.6 & 69.2 & 47.0 & 75.4 & 34.6 & 31.6 & 68.8 & 16.0 \\
    TopK SAE & 16/1024 & 78.1 & 73.0 & 52.6 & 79.2 & 42.7 & 56.2 & 86.5 & 17.2 \\
    TopK SAE & 16/2048 & 80.5 & 73.5 & 53.8 & 77.5 & 43.3 & 58.3 & 86.9 & 18.1 \\
    PatchSAE & 16/1024 & 82.9 & 76.5 & 56.8 & 80.8 & 44.6 & 41.9 & 83.6 & 11.8 \\
    PatchSAE & 16/49152 & 84.1 & 77.2 & 56.8 & 81.1 & 44.4 & 48.6 & 82.9 & 8.0 \\
    MSAE & 16/1024 & 85.2 & 79.9 & 58.9 & 81.7 & 46.1 & 41.4 & 82.2 & 12.5 \\
    MSAE & 16/8192 & \underline{85.6} & 79.5 & 58.3 & 82.3 & 45.4 & 41.2 & 79.9 & 12.6 \\
    PIP-Net\textsubscript{stage 1} & 16/768 & 76.9 & 71.5 & 49.1 & 78.6 & 36.3 & 66.6 & 87.7 & \underline{27.0} \\
    ProtoQuant\textsubscript{stage 1} & 16/2048 & 85.2 & \underline{83.9} & \underline{72.5} & \underline{90.4} & \underline{61.8} & \underline{79.2} & \textbf{92.8} & 22.7 \\
    \midrule
    \rowcolor{oursrow} \textbf{\ours{}} (ours) & 16/1024 & \textbf{89.2}\dpos{+3.6} & \textbf{88.7}\dpos{+4.8} & \textbf{80.6}\dpos{+8.1} & \textbf{92.9}\dpos{+2.5} & \textbf{76.1}\dpos{+14.3} & \textbf{79.7}\dpos{+0.5} & \underline{92.4}\dneg{-0.4} & \textbf{30.5}\dpos{+3.5} \\
    \bottomrule
  \end{tabular}}
  \vspace{-1mm}%
\end{table*}

\myparagraph{Backbones and design choices.} \ours{} is not tied to one pretraining objective. Across seven configurations of six frozen backbones, self-supervised (DINOv2 and DINOv3~\citep{simeoni2025dinov3}) and vision-language (CLIP~\citep{radford2021learning}, SigLIP2~\citep{tschannen2025siglip2}, TIPSv2~\citep{cao2026tipsv2} and CLIP-DINOiser), it stays within $2$~p.p.\ of each backbone's own linear probe on \textbf{6 of the 7} and within $3.5$~p.p.\ on CLIP-DINOiser (see Figure~\ref{fig:pin-full-heatmap-backbones} in Appendix~\ref{abl:backbones}). At the same budget of $16/1024$, DINOv2 gives the best concepts: it leads the other six configurations on \textbf{7 of the 8} concept metrics of Table~\ref{tab:pin-concepts}, all but impurity, and on part-to-part retrieval. Appendix~\ref{app:ablations} presents our design choices for the bank size $N$, the per-image budget $K$ and the losses in detail.

\vspace{-1mm}%
\subsection{Qualitative results and interpretability}
\label{sec:qualitative}
\vspace{-1.5mm}%

\myparagraph{Part maps.} Figure~\ref{fig:qualitative} compares the concept maps of \ours{} with those of ProtoQuant, CFM and the supervised PDiscoFormer. \ours{} follows the annotated parts and stops at the object boundary: its non-overlapping concept regions match distinct parts of the object, such as the head, body, wings, feet and tail of the kite (Figure~\ref{fig:qualitative}a, first row), where the baselines cover larger regions with one concept or spread into the background. For a part such as a sail, one concept of \ours{} marks it in all four images, while a baseline's best-matching concept misses some of them or covers the whole object. Appendix~\ref{app:qualitative} gives further examples on additional datasets, and failure cases.

\providecommand{\QualLayout}{hero}
\providecommand{\QualHero}{kiteagama}
\providecommand{\QualBody}[1]{\includegraphics[width=\linewidth]{\RepoFigureRoot/figures/qualitative/#1.pdf}}
\edef\QualSelectedLayout{\QualLayout}
\edef\QualSelectedHero{\QualHero}
\def\QualIsComposite{composite}\def\QualIsMaps{maps}\def\QualIsHero{hero}\def\QualHeroPinCub{pincub}\def\QualHeroKiteAgama{kiteagama}\def\QualHeroKitePatas{kitepatas}%
\def\QualHeroKiteAgamaSail{kiteagamasail}\def\QualHeroKiteAgamaSailFour{kiteagamasailfour}%
\def\QualHeroKiteMarmoset{kitemarmoset}\def\QualHeroKiteAirliner{kiteairliner}\def\QualHeroKiteGoldfish{kitegoldfish}%
\def\QualHeroKiteBlackbird{kiteblackbird}%
\newcommand{\QualMapsLegendText}{Each method's hard concept map is cut into segments, the segments are
matched one-to-one to the ground-truth parts of that image, and a matched segment takes its part's color,
so agreement with the \emph{GT parts} column reads as matching colors. Hatched segments lie on the object
but match no part (over-segmentation); pale outlined segments lie on the background.
$^{\dagger}$\,trained with class labels}
\newcommand{\QualMapsLegendTextSub}{Each method's hard concept map is cut into segments, the segments are
matched one-to-one to the ground-truth parts of that image, and a matched segment takes its part's color,
so agreement with the \emph{GT parts} column reads as matching colors. Hatched segments lie on the object
but match no part (over-segmentation); they take a part's color when at least 60\% of the segment lies
within that part, and are gray otherwise. Pale outlined segments lie on the background.
$^{\dagger}$\,trained with class labels}
\ifx\QualSelectedLayout\QualIsHero
\begin{figure*}[t]
  \centering
  \ifx\QualSelectedHero\QualHeroPinCub\QualBody{hero_pin_cub}\else\ifx\QualSelectedHero\QualHeroKiteAgama\QualBody{hero_kite_agama}\else
  \ifx\QualSelectedHero\QualHeroKitePatas\QualBody{hero_kite_patas}\else
  \ifx\QualSelectedHero\QualHeroKiteAgamaSail\QualBody{hero_kite_agama_sail}\else
  \ifx\QualSelectedHero\QualHeroKiteAgamaSailFour\QualBody{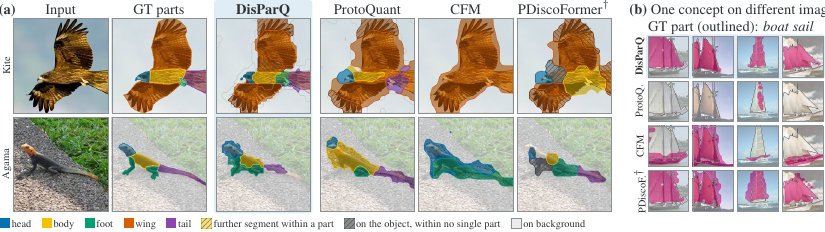}\else
  \ifx\QualSelectedHero\QualHeroKiteMarmoset\QualBody{hero_kite_marmoset}\else
  \ifx\QualSelectedHero\QualHeroKiteAirliner\QualBody{hero_kite_airliner}\else
  \ifx\QualSelectedHero\QualHeroKiteGoldfish\QualBody{hero_kite_goldfish}\else
  \ifx\QualSelectedHero\QualHeroKiteBlackbird\QualBody{hero_kite_blackbird}\else
  \QualBody{hero_pin_pin}\fi\fi\fi\fi\fi\fi\fi\fi\fi%
  \vspace{-2.3mm}%
  \ifx\QualSelectedHero\QualHeroKiteAgamaSail
  \caption{\textbf{Part concepts on PartImageNet.} \emph{(a)}~Part concept maps for \ours{} vs.\ baselines. \ours{} assigns each location exactly one concept, with no overlap, following object parts, while the baselines' overlapping concepts merge parts or spread into the background. \emph{(b)}~Each method's concept most consistent with the outlined sail, on five images. $^{\dagger}$\,trained with class labels. %
  }
  \else\ifx\QualSelectedHero\QualHeroKiteAgamaSailFour
  \caption{\textbf{Part concepts on PartImageNet.} \emph{(a)}~Part concept maps for \ours{} vs.\ baselines. \ours{} assigns each location exactly one concept, with no overlap, following object parts, while the baselines' overlapping concepts merge parts or spread into the background. \emph{(b)}~Each method's concept most consistent with the outlined sail, on four images. $^{\dagger}$\,trained with class labels.}
  \else\ifx\QualSelectedHero\QualHeroKiteBlackbird
  \caption{\textbf{Part concepts on PartImageNet and CUB.} Hard concept maps of \ours{}, ProtoQuant, CFM and the supervised PDiscoFormer. A segment matched to an annotated part (a part mask on PartImageNet, a keypoint group on CUB) takes its color; a hatched segment is a further piece on the object, in its part's color when it lies within one part; pale segments lie on the background. $^{\dagger}$\,trained with class labels. Further examples, and one concept followed across images, in Appendix~\ref{app:qualitative-pin}.\protect}
  \else\ifx\QualSelectedHero\QualHeroPinCub
  \caption{\textbf{Part concepts on PartImageNet and CUB.} Hard concept maps of \ours{}, ProtoQuant, CFM and the supervised PDiscoFormer. A segment matched to an annotated part (a part mask on PartImageNet, a keypoint group on CUB) takes its color; a hatched segment is a further piece on the object, in its part's color when it lies within one part; pale segments lie on the background. $^{\dagger}$\,trained with class labels. Further examples, and one concept followed across images, in Appendix~\ref{app:qualitative-pin}.\protect}
  \else
  \caption{\textbf{Part concepts on PartImageNet.} Hard concept maps of \ours{}, ProtoQuant, CFM and the supervised PDiscoFormer. A segment matched to a ground-truth part takes its color; a hatched segment is a further piece on the object, in its part's color when it lies within one part; pale segments lie on the background. $^{\dagger}$\,trained with class labels. Further examples, and one concept followed across images, in Appendix~\ref{app:qualitative-pin}.\protect}
  \fi\fi\fi\fi
  \label{fig:qualitative}
  \vspace{-3.5mm}%
\end{figure*}
\else\ifx\QualSelectedLayout\QualIsComposite
\begin{figure*}[t]
  \centering
  \providetoggle{qualfourrows}%
  \iftoggle{qualfourrows}{\QualBody{pin_composite_cfm_subpart_4rows}}{\QualBody{pin_composite_cfm_subpart}}%
  \vspace{-2.3mm}%
  \caption{\textbf{Part concepts on PartImageNet.} \emph{(a)}~Hard concept maps: a segment matched to a ground-truth part takes its color; hatched segments lie on the object but match no part, and pale ones lie on the background. \emph{(b)}~For an outlined part, each method's concept most consistent with it over the validation split, on five images. $^{\dagger}$\,trained with class labels. More examples in Appendix~\ref{app:qualitative-pin}.\protect}
  \label{fig:qualitative}
  \vspace{-3.5mm}%
\end{figure*}
\else\ifx\QualSelectedLayout\QualIsMaps
\begin{figure*}[t]
  \centering
  \QualBody{pin_maps_msae}
  \caption{\textbf{Part concepts of \ours{} and of the baselines on PartImageNet (selected examples).}
  \QualMapsLegendText{}. Images were chosen where \ours{} does best; uncurated samples and failure cases are in
  Appendix~\ref{app:qualitative}.}
  \label{fig:qualitative}
\end{figure*}
\else
\begin{figure*}[t]
  \centering
  \QualBody{pin_figA_msae}
  \caption{\textbf{Part concepts of \ours{} and of the baselines on PartImageNet (selected examples).}
  \QualMapsLegendText{}. Images were chosen where \ours{} does best; uncurated samples and failure cases are in
  Appendix~\ref{app:qualitative}.}
  \label{fig:qualitative}
\end{figure*}
\begin{figure*}[t]
  \centering
  \QualBody{pin_figB_cfm}
  \caption{\textbf{One concept, different images (selected examples).} For a ground-truth part (outlined), the
  one concept of each method that is most consistent with that part over the validation split, shown on the
  same five images. The concept of \ours{} marks the part in every image; a baseline's best concept is absent
  from most of them, or covers the whole object. $^{\dagger}$\,trained with class labels.}
  \label{fig:qualitative-concepts}
\end{figure*}
\fi\fi\fi

\myparagraph{Interpretability.} \ours{} explains a prediction through its parts (Figure~\ref{fig:teaser}). Removing a concept from the discrete description and decoding the rest again measures how much the predicted class depends on that part, an occlusion test~\citep{zeiler2014visualizing} on parts instead of pixels. The four largest contributions account for $56$-$100\%$ of the positive logit drops, and each retrieves the same part in other images, so a prediction reads as a short list of recognizable parts with their weights (details in Appendix~\ref{app:qualitative-ours}).

\myparagraph{User study.}
We asked 120 participants to evaluate concepts of DisParQ, PIP-Net, and CFM on CUB, Cars, and PartImageNet. For each prototype, they rated how consistently it captures the same object part (1--5, where 5 is most consistent) and identified which of four regions in a new image it corresponds to. DisParQ prototypes were rated significantly more consistent than both baselines (Figure~\ref{fig:user_study}, right) and were matched to the correct part most often (69\% vs.\ 60\% for PIP-Net and 53\% for CFM, random chance is 25\%). Details are in Appendix~\ref{app:user-study}.

\vspace{-1mm}%
\section{Conclusion}
\label{sec:conclusion}
\vspace{-1.5mm}%

We presented \ours{}, which describes an image from a frozen self-supervised Vision Foundation Model (VFM) by a few discrete part concepts, one per location, each with a quantized attribute, learned without class labels or language. This description preserves the backbone's accuracy, gives part concepts that are more consistent across images than those of language-grounded and sparse-autoencoder dictionaries, and keeps a concept's identity across categories, so a prediction reads as a short list of parts and their contributions. In future work, the discrete concepts could support searching by part and appearance, and correcting a prediction by removing or replacing concepts.

\section*{Ethics statement}
We use existing public datasets and pretrained models for model training and benchmark evaluation, without using private datasets. To assess interpretation quality, we also conduct a user study on Clickworker that was compliant with our organization procedures and regulations. Participants were informed about the study and compensated for their participation, and we collected no identifying information.

\ours{} makes the representations of vision foundation models easier to inspect by expressing them through discrete part concepts. Although these concepts can reveal patterns used by a model, they do not by themselves establish that its predictions are fair, reliable, or causally grounded. The method inherits the biases and limitations of its frozen backbone and training data, so even visually coherent concepts may encode spurious correlations. Use in consequential applications would therefore require validation in the intended setting, including an assessment of errors and bias.

\section*{Reproducibility statement}
To support reproduction of our experiments, we describe the model and training objectives in Section~\ref{sec:architecture} and provide a highly detailed, technical description in Appendix~\ref{app:method}. Appendix~\ref{app:protocol} details the datasets and splits, pretrained backbones, evaluation metrics, and baseline protocols. Throughout these descriptions, we distinguish label-free concept learning from the use of labels for downstream probes and evaluation.

\section*{Acknowledgments}

Adam Pardyl's research was supported by a grant from the Faculty of Mathematics and Computer Science under the Strategic Programme Excellence Initiative at Jagiellonian University. Siddhartha Gairola's research was supported in part by ELSA – European Lighthouse on Secure and Safe AI, funded by the European Union under grant agreement No. 101070617. Sukrut Rao and Bernt Schiele were funded in part by the Deutsche Forschungsgemeinschaft (DFG, German Research Foundation) – GRK 2853/1 “Neuroexplicit Models of Language, Vision, and Action” – project number 471607914. The work of Bartosz Zieliński and Dawid Rymarczyk was funded by the “Interpretable and Interactive Multimodal Retrieval in Drug Discovery” project. This project (FENG.02.02-IP.05-0040/23) is carried out within the First Team programme of the Foundation for Polish Science co-financed by the European Union under the European Funds for Smart Economy 2021–2027 (FENG).

Views and opinions expressed are, however, those of the authors only and do not necessarily reflect those of the European Union or European Commission. Neither the European Union nor the European Commission can be held responsible for them.

We would like to thank and acknowledge the Max Planck Computing and Data Facility (MPCDF), whose HPC systems supported a substantial portion of our experiments, including the development of the proposed method, reproduction of baselines, and computation of the final results. Some experiments were performed on servers purchased with funds from the flagship project entitled “Artificial Intelligence Computing Center Core Facility” from the DigiWorld Priority Research Area within the Excellence Initiative -- Research University program at Jagiellonian University in Kraków. We gratefully acknowledge Polish high-performance computing infrastructure PLGrid (HPC Center: ACK Cyfronet AGH) for providing computer facilities and support within computational grant no. PLG/2026/019504 and PLG/2025/018688.

\defaultspacing%
\bibliography{iclr2026_conference}
\bibliographystyle{iclr2026_conference}

\clearpage
\appendix
\makeatletter
\let\app@orig@addcontentsline\addcontentsline
\renewcommand{\addcontentsline}[3]{%
  \app@orig@addcontentsline{#1}{#2}{#3}%
  \ifstrequal{#1}{toc}{\app@orig@addcontentsline{atoc}{#2}{#3}}{}}
\makeatother

\providecommand{\sidolder}[1]{\iftoggle{comments}{{\color{black!55}\textcolor{red}{Older text:}\ #1\sidreview\par}}{}}
\providecommand{\sidmoved}[1]{\iftoggle{comments}{{\noindent\textcolor{red}{Moved here unchanged from #1.}\sidreview\par}}{}}

\section*{Appendix}

\providecommand{\sidblue}{\iftoggle{comments}{\color[HTML]{1F5FBF}}{}}
This appendix provides additional details and results for \ours{}. We first expand the discussion of related work (Appendix~\ref{app:related-work}) and discuss limitations (Appendix~\ref{app:limiations}). The architecture and training procedure are described in detail next (Appendix~\ref{app:method}). Then we present the experimental protocol, including the datasets, backbones, training procedures, and evaluation metrics used for \ours{} and the baselines (Appendix~\ref{app:protocol}). We examine each component through ablations (Appendix~\ref{app:ablations}) and provide extended results, including the full main-paper tables, a qualitative analysis of quantized attributes, and user-study details (Appendix~\ref{app:additional-results}). Lastly, we show qualitative results across all five datasets and illustrate how \ours{} explains predictions (Appendix~\ref{app:qualitative}), before describing how learned concepts can be named using a fixed vocabulary (Appendix~\ref{app:concept-naming}).

\noindent\textbf{Table of contents}

\begingroup
\makeatletter
\renewcommand*{\numberline}[1]{\makebox[\app@numw][l]{\color{black}#1}}%
\renewcommand*{\l@section}[2]{\par\vspace{7pt}\noindent{\def\app@numw{2em}\bfseries #1}\dotfill #2\par}
\renewcommand*{\l@subsection}[2]{\par\vspace{2pt}\noindent\hspace*{2em}{\def\app@numw{2.6em}#1}\dotfill #2\par}
\renewcommand*{\l@subsubsection}[2]{\par\vspace{1pt}\noindent\hspace*{4.6em}{\def\app@numw{3.3em}#1}\dotfill #2\par}
\@starttoc{atoc}
\makeatother
\endgroup

\clearpage

\section{Related work}
\label{app:related-work}

\myparagraph{Prototypical parts and part discovery.} ProtoPNet~\citep{chen2019looks} introduced ``this looks like that'' reasoning: an image is classified by how closely its patches match prototypes learned for each class. Later work shares prototypes across classes to build compact vocabularies~\citep{nauta2021neural, rymarczyk2021protopshare, rymarczyk2022interpretable, nauta2023pip}.
Others refine the geometry of the prototype space~\citep{wang2021interpretable, donnelly2022deformable} or separate color from other visual features~\citep{pach2025lucidppn}. Recent variants build on Vision Transformers~\citep{xue2022protopformer, ma2024interpretable}, including frozen foundation backbones~\citep{struski2024infodisent, turbe2024protos, dubovik2025side, janusz2026protoquant}. Among these, ProtoQuant~\citep{janusz2026protoquant} also quantizes its prototypes into a codebook, but still classifies from continuous similarity scores. Part discovery methods segment objects into parts, either without any labels~\citep{hung2019scops, choudhury2021unsupervised, amir2021deep} or with
class labels as the only signal~\citep{huang2020interpretable, van2023pdisconet, aniraj2024pdiscoformer}. Prototypes are learned to serve a classifier, and discovered parts form a small, fixed set per category or dataset. \ours{} instead learns, without class labels, a large shared dictionary from which each image selects a few concepts, assigns each patch exactly one of them, and captures how each of them varies across images with an attribute, making a frozen vision foundation model interpretable at the level of parts (Table~\ref{tab:related-properties}).

\myparagraph{Concept bottlenecks and concept dictionaries.} Concept bottleneck models (CBMs) make predictions through a layer of human-defined concepts~\citep{koh2020concept}. Recent CBMs let a language model write the concept list and score it with a vision-language model~\citep{yuksekgonul2022post, oikarinen2023label, yang2023language}, and spatial variants score concepts on image regions~\citep{panousis2024coarse} or as concept maps~\citep{benou2025show}. Sparse autoencoders (SAEs) instead learn an unnamed, overcomplete dictionary from a model's activations. They have recently been used to interpret language models~\citep{cunningham2024sparse, gao2025scaling} and have since been applied to vision: PatchSAE~\citep{lim2025sparse} trains them on the patch tokens of a vision transformer, and later variants improve the dictionary itself~\citep{zaigrajew2025interpreting, fel2025archetypal}.
DN-CBM~\citep{rao2024discover} names SAE concepts with CLIP, and the concept foundation model CFM~\citep{wittenmayer2026cfm} builds a spatially grounded concept dictionary on a language-aligned backbone and names its concepts from a predefined text vocabulary. Most CBMs name their concepts in advance, and SAE codes are continuous weights over concepts that overlap at each patch, which entangles a part's identity with its appearance. \ours{} learns its concepts from visual features alone, assigns each patch exactly one concept, separates a part's identity from how it varies across images (its attribute), and describes an image by at most 16 active concepts in integer indices alone, the budget at which we compare all dictionaries (Section~\ref{sec:setup}).

\myparagraph{Evaluating part concepts.} Part concepts are usually scored by comparing each concept's map with part annotations: through the purity of prototypes on annotated keypoints~\citep{nauta2023pip}, keypoint regression and clustering agreement with annotated parts~\citep{van2023pdisconet,
aniraj2024pdiscoformer}, or the locality and consistency of concept maps~\citep{wittenmayer2026cfm}. A separate benchmark tests whether a prototype's activation map marks the image region that actually drives it~\citep{sacha2024interpretability}, and semantic correspondence matches keypoints between two given images, usually of the same category~\citep{min2019spair, amir2021deep}. Our part-to-part retrieval benchmark instead starts from one annotated part region and asks whether its description finds the same part among the regions of all other images, across categories (Section~\ref{sec:setup} and Appendix~\ref{app:metrics}).\par

\section{Limitations}
\label{app:limiations}

In this section we discuss the limitations in our proposed method and study. Part annotations may be coarser than the learned concepts, but over-segmentation and incorrect semantic grouping can also cause failures (Appendix~\ref{app:qualitative-pin}). The frozen backbone passes on its features and biases, including part of the class structure of the attributes (Appendix~\ref{app:qualitative-attrspace}). Contextual backbone features and soft pooling can carry information from outside a displayed hard region, so spatial assignments do not establish pixel-level causal locality. Concepts remain unnamed without examples or alignment (Appendix~\ref{app:concept-naming}). They cover the whole image, including background, with the same budget of $K{=}16$ concepts regardless of its number of parts (Appendix~\ref{abl:budget}).

\clearpage
\providecommand{\sidblue}{\iftoggle{comments}{\color[HTML]{1F5FBF}}{}}
\section{Architecture and training details}
\label{app:method}

This section specifies details of \ours{} as introduced in Section~\ref{sec:architecture}, following the same narrative structure. It describes the canonical form used in the main experiments. Differences for studies with alternative configurations are described in the relevant sections.

The default \ours{} configuration uses frozen DINOv2-B/14~\citep{oquab2023dinov2} with registers~\citep{darcet2024registers}, a bank of $N=1024$ concepts, $K=16$ selected concepts per image, and private $16\times64$ product quantization. Every patch is assigned to one concept. Table~\ref{tab:method-notation} recaps the principal symbols and dimensions used throughout this section. Spatial positions are flattened in row-major order. Batch statistics refer to the local device batch unless stated otherwise.

\begin{table}[ht]
\centering
\caption{Notation and dimensions for the canonical model. Feature width $D$ and decoder width $d$ are distinct quantities, both equal to 768 here. $K$ counts concepts and $G$ counts attribute groups.}
\label{tab:method-notation}
\small
\begin{tabular}{@{}lll@{}}
\toprule
Symbol & Meaning & Value / shape \\
\midrule
$D,d$ & Feature width, decoder width & $768$, $768$ \\
$B,b,u$ & Local batch size, image index, spatial index & $16$; $1\le b\le B$; $1\le u\le HW$ \\
$z_u,c(x)$ & Spatial features, global target & $\mathbb R^D$ \\
$w_n,e_n$ & Concept prototype, decoder value & $\mathbb R^D$, $\mathbb R^d$ \\
$N,K$ & Bank size, selected concepts & $1024$, $16$ \\
$i_k,s_u,p_u$ & Global ID, local owner, spatial global ID & $p_u=i_{s_u}$ \\
$\chi_{b,n},\psi_n$ & Selection score, full-bank usage distribution & Scalars \\
$A,\bar{A}$ & Pre- and post-top-$K$ soft maps & $N\times HW$, $K\times HW$ \\
$H\times W$ & Shared assignment and decoder grid & $32\times32$ \\
$\bar z_{i_k},r_{i_k},q_{i_k}$ & Pooled feature, residual, attribute & $\mathbb R^D$, $\mathbb R^D$, $\mathbb R^d$ \\
$\mathcal B,j_{i_k,g}$ & Private codebooks, codeword index & $N\times G\times M\times(d/G)$; $0\leq j<M$ \\
$G,M$ & PQ groups, entries per group & $16$, $64$ \\
$\rho,\eta$ & Usage EMA, selection dual & $\mathbb R^N$ each \\
$\widetilde Z,V_m$ & Area-scaled and modulated region features & $D\times K$ \\
$W_{\mathrm{cont}},W_{\mathrm{PQ}}$ & Continuous and quantization projections & $d\times D$ each \\
$\mathrm{PE},t_u$ & Position table, decoder token & $HW\times d$, $\mathbb R^d$ \\
\bottomrule
\end{tabular}
\end{table}

\subsection{Feature extraction}
\label{app:features}

Input images have size $224\times224$ and $3$ RGB channels. The frozen DINOv2-B/14 encoder~\citep{oquab2023dinov2} produces width-$D=768$ patch features on a native $16\times16$ grid ($14\times14$ patch size). Its final layer-normalized CLS token is the global distillation target $c(x)\in\mathbb R^D$. CLS and register tokens are removed from the spatial sequence. The backbone and the pretrained upsampler remain in evaluation mode throughout training. Backbone substitutions and their target definitions are described in Appendix~\ref{abl:backbones}.

To combine information across depths while retaining spatial correspondence, the lightweight attention pooling operator $f_{\mathrm{fuse}}$ consumes raw patch outputs $h_{\lambda,\xi}\in\mathbb R^D$ from the last four blocks of the encoder. We restrict fusion to these later blocks to favor semantic features and reduce the opportunity for concepts to organize around positional cues from earlier layers. A shared bias-free key projection $W_{\mathrm{key}}\in\mathbb R^{64\times D}$ and a learned query $\omega^{\mathrm{fuse}}\in\mathbb R^{64}$ form four 16-dimensional heads using scaled dot-product attention~\citep{vaswani2017attention}:
\begin{equation}
 \alpha_{a,\lambda,\xi}=\operatorname{softmax}_{\lambda}
 \left(\frac{(\omega^{\mathrm{fuse}}_a)^\top(W_{\mathrm{key}}h_{\lambda,\xi})_a}{\sqrt{16}}\right),
 \qquad z^{\mathrm{fuse}}_\xi=\frac14\sum_{a=1}^4\sum_{\lambda=1}^4
 \alpha_{a,\lambda,\xi}h_{\lambda,\xi}.
\end{equation}
Here $\xi$ indexes the native patch grid, $\lambda\in\{1,\ldots,4\}$ indexes the selected encoder blocks, and $a\in\{1,\ldots,4\}$ indexes attention heads. Each head mixes full $D$-dimensional values. There is no value/output projection, extra normalization, residual connection, or positional encoding in fusion. It has $64D+64$ parameters.

The query is initialized from a truncated normal with standard deviation $0.02$. Throughout this appendix, truncated-normal initialization uses mean zero and absolute truncation bounds $[-2,2]$, as in the default PyTorch initializer. The key projection entries are uniform on $[-D^{-1/2},D^{-1/2}]$. Removing fusion leaves only the final-block patch features, as studied in Appendix~\ref{abl:fusion}.

\myparagraph{Image-guided upsampling.}
To represent finer concept boundaries than the native $16\times16$ grid permits, we upsample the fused features with image guidance.
The multi-backbone AnyUp model~\citep{wimmer2026anyup}, using its second released pretrained weight set, receives the ImageNet-normalized image and the fused feature map. It mixes unprojected feature values and preserves width $D$, producing $z_u\in\mathbb R^D$ on a $32\times32$ grid. Its computation uses float32 before restoring the feature dtype. AnyUp's parameters are frozen, but stage 1 propagates gradients through its feature input to fusion. Stage 2 evaluates the frozen feature path without gradients, because fusion is also frozen in that stage.

A less computationally expensive alternative is bilinear interpolation of the fused features to the same grid, with pixel-center coordinates (zero-based output coordinate $o\in\{0,\ldots,31\}$ maps to source coordinate $(o+\tfrac12)/2-\tfrac12$ along each axis, using boundary replication). Appendix~\ref{abl:upsampling} compares this alternative, native-resolution features, and mixed AnyUp/bilinear training. Bilinear and AnyUp scores are broadly comparable, with metric-dependent differences, while AnyUp yields slightly more detailed qualitative concept maps.

\subsection{Concept identification and assignment}
\label{app:selection}
Given the upsampled features, we first score the full concept bank and then restrict spatial competition to the selected concepts. The squared-distance competition in Equation~\ref{eq:bank-competition} follows the prototype assignment used in PDiscoNet and PDiscoFormer~\citep{van2023pdisconet,aniraj2024pdiscoformer}, here extended to a large bank with per-image selection. It uses unnormalized prototypes and features. Each prototype $w_n\in\mathbb R^D$, $n\in\{1,\ldots,N\}$, is initialized from a truncated normal with standard deviation $0.02$. Training adds independent standard Gumbel noise to each logit and uses temperature 1, with soft outputs~\citep{jang2017categorical}. The mean activation score is $\chi_{b,n}=(HW)^{-1}\sum_u A_{b,n,u}$. A standard Gumbel sample is $-\log(-\log U)$ for $U$ uniform on $(0,1)$, sampled independently for each image, concept, and location. Let $B$ be the local batch size and $b\in\{1,\ldots,B\}$ index its images. From the $B\times N$ score matrix, form
\begin{equation}
 \widetilde\chi_{b,n}=\chi_{b,n}-0.5\log(\rho_n+10^{-8}),\qquad
 Q^{(0)}=\frac{\exp(\widetilde\chi/0.1)}{\sum_{b,n}\exp(\widetilde\chi_{b,n}/0.1)},
\end{equation}
where $\rho_n$ is the exponential moving average (EMA) of selection frequency, updated as specified in Appendix~\ref{app:curricula}. We use Sinkhorn--Knopp normalization~\citep{sinkhorn1967concerning}, following the use of batch-balanced prototype assignments in SwAV~\citep{caron2020swav}. Here balancing selects a subset of concepts rather than providing a swapped-prediction target. Ten iterations alternately divide each column by $N$ times its sum and each row by its sum, clamping denominators to $10^{-8}$. Normalization is performed in float32, under stop-gradient. The largest $K$ entries in each final row select distinct IDs, retained in ranking order. Stable sorting favors lower bank IDs at exact ties. A per-concept score bias that is constant across images cancels under ideal Sinkhorn normalization, so the coverage term is not an independent balancing force in exact arithmetic. Top-$K$ truncation also does not guarantee exactly balanced hard selections.

The bank-size and per-image-budget studies in Appendix~\ref{abl:bank} vary $N$ and $K$ separately. Appendix~\ref{abl:selection} compares peak scoring, direct mass-based selection, and removal of the coverage bias.

Balancing is per local batch, without gathering score matrices across devices. A training batch of one degenerates to ties. The post-selection maps $\bar{A}$ are recomputed from clean logits with a second, independent Gumbel sample. Slicing the first sampled maps would change the stochastic computation. Each location has local owner $s_u=\arg\max_k \bar{A}_{i_k,u}$ and global concept identity $p_u=i_{s_u}$, with $k\in\{1,\ldots,K\}$. The $K$ selected concepts need not all own a location: a selected concept can have positive soft mass but an empty hard region. Its attribute then contributes no spatial decoder token. These discrete choices have no derivative. An identity surrogate in Appendix~\ref{app:gradients} supplies a gradient to the soft maps.

\myparagraph{Batch-independent evaluation.}
During training, a second balancing calculation on the original view's raw scores estimates a centered dual. If $Q$ is its balanced matrix, set
\begin{equation}
 \delta_{b,n}=\log\max(Q_{b,n},10^{-30})-\chi_{b,n}/0.1,\qquad
 \eta^{\mathrm{new}}_n=\operatorname{center}_n\!\left[
   \frac1B\sum_b\operatorname{center}_n(\delta_{b,n})\right].
\end{equation}
Here $\operatorname{center}_n$ subtracts the mean over the $N$ concept indices. In distributed training, we average this estimate across devices. We then update $\eta$ by exponential moving average with decay $0.99$ and recenter it, initializing it with the first estimate. Evaluation uses ordinary softmax and selects the largest $K$ values of $\chi_n(x)/0.1+\eta_n$. It omits the coverage term. The frozen PQ stage retains the dual rather than updating it.

\subsection{Attribute encoding and quantization}
\label{app:attributes}

Once the concept regions are defined, appearance encoding summarizes how each occurrence differs from its prototype. We first compute area-scaled pooled features, as in PDiscoNet~\citep{van2023pdisconet}, together with their soft region masses:
\begin{equation}
 \widetilde z_{i_k}=(HW)^{-1}\sum_u \bar{A}_{i_k,u}z_u,\qquad
 \mu_{i_k}=(HW)^{-1}\sum_u \bar{A}_{i_k,u}.
\end{equation}
The weighted region mean in Equation~\ref{eq:appearance-residual} is stabilized as $\bar z_{i_k}=\widetilde z_{i_k}/(\mu_{i_k}+10^{-6})$. The normalization $\nu(y)=y/\max(\|y\|_2,10^{-12})$ is evaluated in float32 for each vector $y$. The residual is then cast back to the feature dtype without further normalization. Both stages pass $\operatorname{stopgrad}(r_{i_k})$ to the appearance encoder, preventing the appearance route from changing the concept reference or region features to make its own prediction easier. Assignment learning instead receives the identity-surrogate, auxiliary, and regularization signals described below. In stage 1, $q_{i_k}=W_{\mathrm{cont}}\operatorname{stopgrad}(r_{i_k})$, where the shared bias-free projection $W_{\mathrm{cont}}\in\mathbb R^{d\times D}$ is initialized from a truncated normal with standard deviation $0.02$. Appendix~\ref{abl:gradients} tests restoring residual gradients, while Appendix~\ref{abl:attributes} removes the appearance channel entirely.

\myparagraph{Product quantization.}
\label{app:quantization}
To make this appearance channel discrete, we use product quantization (PQ)~\citep{jegou2011product}. We project each residual to the decoder width with $y_{i_k}=W_{\mathrm{PQ}}\operatorname{stopgrad}(r_{i_k})\in\mathbb R^d$ and split it into $G$ channel groups indexed by $g\in\{1,\ldots,G\}$. Each selected concept uses the codebooks of its bank identity $i_k$.

Each group is represented by one of $M$ codewords:
\begin{equation}
  j_{i_k,g}=\arg\min_{j\in\{0,\ldots,M-1\}}
       \|y_{i_k}^{(g)}-\mathcal B_{i_k,g,j}\|_2^2,
  \qquad q_{i_k}=\bigoplus_{g=1}^{G}\mathcal B_{i_k,g,j_{i_k,g}},
  \label{eq:private-pq}
\end{equation}
Here $\mathcal B_{i_k,g,j}$ is the $j$th codeword in group $g$ of concept $i_k$, and $\bigoplus$ concatenates the selected codewords into the quantized attribute vector $q_{i_k}$.

The codebooks $\mathcal B_{i_k,g}$ are concept-specific, so appearance codes can specialize to each concept. Product quantization represents variation through combinations of codewords rather than a separate entry for every complete attribute vector. The indices $j_{i_k,1:G}$ form the discrete appearance code. Figure~\ref{fig:crisp-quantization} illustrates the attribute path and its connection to the decoder.

\myparagraph{PQ surrogate.}
Hard codeword selection requires a surrogate gradient during training. Following the straight-through principle~\citep{bengio2013estimating}, we retain hard forward values while differentiating a soft codeword mixture. We use $G=16$ channel groups and $M=64$ entries in each concept-specific codebook, with group width $d/G$. For each group in Equation~\ref{eq:private-pq}, training uses
\begin{equation}
 \begin{aligned}
 \pi_{i_k,g,j}&=\operatorname{softmax}_j\!\left(-\|y_{i_k}^{(g)}-\mathcal B_{i_k,g,j}\|_2^2\right),\\
 \bar q_{i_k}^{(g)}&=\sum_j\pi_{i_k,g,j}\mathcal B_{i_k,g,j},\qquad
 q_{i_k}^{(g)}=\mathcal B_{i_k,g,j_{i_k,g}}+\bar q_{i_k}^{(g)}-\operatorname{stopgrad}(\bar q_{i_k}^{(g)}).
 \end{aligned}
 \label{eq:pq-surrogate}
\end{equation}
The softmax temperature is fixed at 1. Here $\pi_{i_k,g,j}$ is the soft probability of codeword $j$ for concept $i_k$ and group $g$, and $\bar q_{i_k}^{(g)}$ is their weighted mixture. The backward pass includes the direct gradient to the chosen codeword and the gradient through the soft mixture, including its dependence on the codebooks. Thus, unlike the identity-gradient estimator in VQ-VAE~\citep{oord2017neural}, our surrogate differentiates the soft assignment probabilities as well as the codeword values. Evaluation performs hard lookup only. The shared bias-free projection and private tables are learned by global cosine distillation. There is no commitment loss, residual-reconstruction loss, codebook EMA, k-means initialization, or attribute-entropy penalty. Both projection and codebooks start from a truncated normal with standard deviation $0.02$. Appendix~\ref{abl:quantization} compares continuous attributes, private and shared codebooks, and vector, residual, and product quantization at different bit rates.

\subsection{Concept decoder}
\label{app:decoder}

The decoder predicts the global VFM representation (its CLS token) from concept identities, spatial assignments, and attributes, as in Equation~\ref{eq:spatial-decoder}. Spatial self-attention~\citep{vaswani2017attention} lets the readout combine concepts according to their arrangement, while learned positions and a decoder CLS token follow the ViT readout pattern~\citep{dosovitskiy2021image}. Its $N$ learnable concept embeddings $e_n\in\mathbb R^d$ are independent of the matching prototypes $w_n$, allowing the representation used for decoding to differ from the reference used for feature matching. Assignment and decoding share the $H\times W$ grid, so identities and attributes are broadcast directly using $p_u$, without any spatial pooling. Concept values, the $HW\times d$ positional table, and the learned decoder CLS token are initialized from a truncated normal with standard deviation $0.02$.

The decoder has three pre-normalization blocks, four attention heads, and feed-forward width $4d$ with GELU~\citep{hendrycks2016gelu}. Attention projections are bias-free, feed-forward layers have biases. LayerNorm~\citep{ba2016layer} has learned affine parameters and $\epsilon=10^{-5}$. There is no attention/transformer dropout, mask, or stochastic depth. A final $\operatorname{LayerNorm}(d)\to\operatorname{Linear}(d,D)$ maps the output CLS token to the predicted global representation $\hat c(x)$. The CLS input has no separate positional embedding. The output is not explicitly unit-normalized. Each block applies pre-normalized self-attention with a residual connection, then a pre-normalized two-layer feed-forward network with a second residual connection. LayerNorm scales and offsets start at one and zero. Unless initialized explicitly above, linear weights and biases are uniform on $[-d_{\mathrm{in}}^{-1/2},d_{\mathrm{in}}^{-1/2}]$, where $d_{\mathrm{in}}$ is that layer's input width. The concatenated attention query/key/value projection instead uses Xavier-uniform initialization~\citep{glorot2010understanding}.

Given fixed model parameters, decoding requires only the selected bank IDs $i_{1:K}$, the $H\times W$ local-owner map $s$, and the $K\times G$ PQ indices. Its nominal fixed-width payload is
\begin{equation}
 K\log_2N+HW\log_2K+KG\log_2M,
\end{equation}
or 2,720 bits at $H=W=16$ and 5,792 bits at $H=W=32$. These counts exclude model parameters and serialization overhead. Private PQ tables contain $NMd$ parameters (25,165,824 at $d=384$ and 50,331,648 at $d=768$), plus $Dd$ projection parameters. Compact per-image codes therefore do not imply a small decoder-side dictionary. Appendix~\ref{abl:quantization} separates attribute rate from sharing of the dictionary, and Appendix~\ref{abl:decoder} compares this decoder with pooling and an auxiliary-only readout, then varies transformer depth.

\subsection{Training objectives and regime}
\label{app:objectives}

\begin{figure*}[htbp]
  \centering
  \begingroup
  \CRISPSetup
  \resizebox{\linewidth}{!}{%
    \input{\RepoFigureRoot/figures/03_training_body.tex}%
  }
  \endgroup
  \caption{\CRISPTrainingCaption}
  \label{fig:crisp-training}
\end{figure*}

\myparagraph{Auxiliary region readout.}
The auxiliary head supervises pooled region content directly, complementing the decoder that sees concept identities and attributes. We adopt PDiscoFormer's normalization-based modulation~\citep{aniraj2024pdiscoformer}, which develops the part modulation of PDiscoNet~\citep{van2023pdisconet}. A learned $\operatorname{LayerNorm}([D,K])$ acts on the area-scaled $\widetilde Z=[\widetilde z_{i_1},\ldots,\widetilde z_{i_K}]\in\mathbb R^{D\times K}$ to form modulated features $V_m$. It normalizes jointly over feature and slot axes. These features support orthogonality and the stage-1 auxiliary head $f_{\mathrm{aux}}$. Following PDiscoNet's part dropout~\citep{van2023pdisconet}, we discourage reliance on a single informative region: the auxiliary branch zeros whole slots with slot dropout at probability $0.3$ (scaling retained slots by $1/0.7$), averages over all $K$ slot positions, and applies $\operatorname{LayerNorm}(D)\to\operatorname{Linear}(D,D)\to\operatorname{GELU}\to\operatorname{Linear}(D,D)$. Modulation and dropout do not enter the primary residual branch.

\myparagraph{Identity surrogate in continuous training.}
\label{app:gradients}
To let global distillation shape assignments despite hard ownership, we use a straight-through construction~\citep{bengio2013estimating,jang2017categorical} based on the selected soft maps $\bar{A}$. During stage 1, we replace the identity contribution in Equation~\ref{eq:spatial-decoder} by
\begin{equation}
 t^{\mathrm{id}}_u=e_{p_u}+\bar e_u-\operatorname{stopgrad}(\bar e_u),
 \qquad \bar e_u=\sum_k \bar{A}_{i_k,u}e_{i_k}.
 \label{eq:identity-ste}
\end{equation}
This has hard forward values but supplies gradients to soft spatial assignments and an additional gradient to concept values. It differentiates neither top-$K$ selection nor hard attribute broadcasting. It connects global distillation to prototypes through $\bar{A}$ and to fusion through AnyUp's feature inputs, while the appearance residual remains detached. It is disabled in stage 2 and evaluation.

Removing the surrogate or enabling residual gradients is studied in Appendix~\ref{abl:gradients}. The auxiliary-only alternative and the removal of auxiliary distillation are distinguished in Appendices~\ref{abl:decoder} and~\ref{abl:objectives}.

\myparagraph{Combined objective.}
Global and auxiliary distillation retain target information, while spatial and usage regularizers shape its organization into concepts. Table~\ref{tab:method-losses} specifies the weights of the nine terms shown in Figure~\ref{fig:crisp-training}. For each view $v\in\{x,T(x)\}$, the auxiliary prediction is $\hat c_{\mathrm{aux}}(v)=f_{\mathrm{aux}}(V_m(v))$ with the pooling and dropout above, and
\begin{equation}
 \mathcal L_{\mathrm{aux}}=\frac12\sum_{v\in\{x,T(x)\}}
 [1-\operatorname{cos}(\hat c_{\mathrm{aux}}(v),\operatorname{stopgrad}(c(v)))].
\end{equation}
The total objective is the sum of the nine losses multiplied by their table weights, averaged over batch images wherever an image index remains. Both cosine losses and usage diversity average their two per-view values. Potts, entropy, orthogonality, and slot presence use only the original view. Bank presence takes maxima over both views, and equivariance compares them. Cosine similarity uses $\epsilon=10^{-8}$. The auxiliary head matches the same view-specific backbone target as the main decoder.

\begin{table}[t]
\centering
\caption{Loss weights and view accounting. Frozen PQ still computes assignment regularizers, but they cannot train the frozen prototype-assignment path. Orthogonality can update the unfrozen modulation parameters.}
\label{tab:method-losses}
\small
\begin{tabular}{@{}lccc@{}}
\toprule
Term & Stage 1 & Stage 2 & Views \\
\midrule
$\mathcal L_{\mathrm{global}}$ & 1 & 1 & Mean of both \\
$\mathcal L_{\mathrm{aux}}$ & 1 & 0 & Mean of both \\
$\mathcal L_{\mathrm{Potts}}$ & 1 & 1 & Original \\
$\mathcal L_{\mathrm{ent}}$ & 1 & 0.5 & Original \\
$\mathcal L_{\mathrm{orth}}$ & 1 & 1 & Original \\
$\mathcal L_{\mathrm{slot}}$ & 1 & 1 & Original \\
$\mathcal L_{\mathrm{eq}}$ & 1 & 1 & Paired \\
$\mathcal L_{\mathrm{div}}$ & 0.1 & 0.1 & Mean of both \\
$\mathcal L_{\mathrm{bank}}$ & 0.1 & 0.1 & Maximum over both \\
\bottomrule
\end{tabular}
\end{table}

\myparagraph{Feature-weighted Potts.}
We favor agreement between neighboring assignments when their features are similar, while weakening smoothing across feature discontinuities. This is a local soft Potts penalty with a Gaussian feature kernel, related to the appearance-dependent pairwise potentials of~\citet{krahenbuhl2011efficient}. We use it in place of PDiscoFormer's unweighted total-variation prior~\citep{aniraj2024pdiscoformer}. Our adaptive temperature keeps confident logits from saturating the smoothing gradient, and the bandwidth adapts to feature-distance scale. Specifically, soften the clean selected logits without Gumbel noise using
$C_{i_k,u}=\operatorname{softmax}_k(\ell_{i_k,u}/\tau_{\mathrm P})$, where
$\tau_{\mathrm P}=\max(1,\operatorname{median}_{b,u}(\ell^{(1)}_{b,u}-\ell^{(2)}_{b,u})/2)$
and superscripts denote the two largest selected logits. Here $C$ denotes the spatial probabilities used by the Potts loss. Image indices are suppressed in the spatial expressions below. For each horizontal/vertical neighbor pair $(u,u')$, counted once, let $\zeta_{uu'}=\|z_u-z_{u'}\|_2^2$ denote its squared feature distance and set
$\sigma^2=\max(10^{-8},\operatorname{quantile}_{0.9}\{\zeta_{uu'}\})$. Then
\begin{equation}
 \mathcal L_{\mathrm{Potts}}=\operatorname{mean}_{b,(u,u')}
 \exp[-\zeta_{uu'}/(2\sigma^2)]\left(1-\sum_k C_{i_k,u}C_{i_k,u'}\right).
\end{equation}
Features used for the kernel, temperature, and bandwidth are detached. Quantiles and medians are over the local batch. There is no normalization by total edge weight.

\myparagraph{Entropy, orthogonality, and presence.}
The entropy term follows PDiscoFormer~\citep{aniraj2024pdiscoformer} and encourages confident spatial ownership. Orthogonality discourages redundant region features, following the part-decorrelation prior in PDiscoNet~\citep{van2023pdisconet} and earlier SCOPS~\citep{hung2019scops}. We use squared Gram-matrix deviations, penalizing both positive and negative correlations, rather than the signed cosine sum written in PDiscoFormer. Selected-slot presence adapts PDiscoNet's batch-presence term: spatial averaging prevents an isolated activation from satisfying it. We additionally apply this criterion to full-bank identities to supply gradients to concepts outside the selected subset.

For batch-indexed maps, $i_{b,k}$ is the bank ID at selected-list position $k$ in image $b$, and $k,l\in\{1,\ldots,K\}$ index selected-list positions. With $\widehat V_m$ denoting channel-normalized columns of $V_m$,
\begin{align}
 \mathcal L_{\mathrm{ent}}&=-\operatorname{mean}_{b,u}\sum_k \bar{A}_{b,i_{b,k},u}\log(\bar{A}_{b,i_{b,k},u}+10^{-8}),\\
 \mathcal L_{\mathrm{orth}}&=\operatorname{mean}_{b,k,l}
 \left[(\widehat V_{m,b}^{\top}\widehat V_{m,b})_{kl}-\mathbf1_{k=l}\right]^2,\\
 \mathcal L_{\mathrm{slot}}&=1-\frac1K\sum_k\max_{b,u}
        \operatorname{AvgPool}_{3\times3}(\bar{A}_{b,i_{b,k}})_u,\\
 \mathcal L_{\mathrm{bank}}&=1-\frac1N\sum_n\max_{b,u,v\in\{x,T(x)\}}
        \operatorname{AvgPool}_{3\times3}(A^{v}_{b,n})_u.
\end{align}
Pooling uses stride 1 without padding. Orthogonality averages over all $BK^2$ entries. Slot presence is indexed by local list position. Bank presence is indexed by globally shared identity and can train unselected prototypes. Presence maxima are not gathered across devices.

\myparagraph{Prototype-usage diversity.}
Subset balancing does not directly penalize concentrated full-bank spatial responses. We therefore add a differentiable usage penalty toward a detached, Sinkhorn-balanced target, using the same balancing principle as above~\citep{sinkhorn1967concerning,caron2020swav}. For each view, let $\psi_n=B^{-1}\sum_b\chi_{b,n}$ be the batch-mean full-bank activation, normalized to sum to one. To form a detached target $\psi^{\mathrm{bal}}$, apply the same ten-iteration balancing routine to $\exp(\operatorname{stopgrad}(\chi))^\top\in\mathbb R^{N\times B}$ (with the column factor now $B$), transpose back, average over images, and normalize. The loss is
\begin{equation}
 \mathcal L_{\mathrm{div}}=\sum_n \psi_n\bigl[\log(\psi_n+10^{-8})-\log(\psi^{\mathrm{bal}}_n+10^{-8})\bigr].
\end{equation}
The exponent here has temperature 1, distinct from the selection temperature $0.1$. The final row normalization of the $N\times B$ matrix makes each prototype row sum to one. Thus, when denominator clamps are inactive, averaging over images and normalizing gives $\psi_n^{\mathrm{bal}}=1/N$ in exact arithmetic. The loss is therefore an epsilon-stabilized KL divergence toward uniform usage. Without the numerical stabilizers, it equals $\log N-\mathcal H(\psi)$, where $\mathcal H(\psi)=-\sum_n\psi_n\log\psi_n$. This term regularizes prototype usage rather than quantizer commitment.

\begin{samepage}
\myparagraph{Affine equivariance.}
To encourage concept identities to follow image content under changes of pose, we adapt the inverse-warp cosine objective of PDiscoNet and PDiscoFormer~\citep{van2023pdisconet,aniraj2024pdiscoformer}. Equivariant part maps were also encouraged in SCOPS~\citep{hung2019scops}. Our comparison uses active full-bank identities so that a change in the selected subset does not change which concepts are compared. For each image $b$, let $\mathcal I_b^{32}$ contain the 32 bank indices with the largest $\max(\chi_{b,n}(x),\chi_{b,n}(T(x)))$. We compare their full-bank maps after inverse warping:\par
\end{samepage}
\begin{equation}
 \mathcal L_{\mathrm{eq}}=1-\operatorname{mean}_{b,n\in\mathcal I_b^{32}}
 \operatorname{cos}\!\left(A_{b,n}(x),
             \mathcal W_{T}^{-1}A_{b,n}(T(x))\right).
\end{equation}
Cosine similarity acts on flattened maps. The resampling operator $\mathcal W_T^{-1}$ first undoes translation, then rotation and scale, using two bilinear resampling operations. Map-space translations are rounded after scaling by the ratio of map to image resolution. No overlap mask removes filled borders. Using globally indexed maps avoids matching unrelated slot positions when the two views select different concepts.

The affine transforms $T$ are sampled as described in Appendix~\ref{app:train-ours}.

\subsection{Training stages and optimization}
\label{app:curricula}

\myparagraph{Two-stage curriculum.}
The continuous first stage lets concepts form before imposing a discrete appearance bottleneck. Freezing assignment in the second stage then gives quantization a stable concept reference. Stage 1 jointly learns fusion, matching prototypes, continuous appearance projection, concept embeddings, decoder, region modulation, and the auxiliary head. It uses detached residuals and the identity surrogate. Stage 2 starts from the stage-1 checkpoint with the lowest validation global cosine distance among checkpoints saved every ten epochs. It retains fusion, prototypes, decoder embeddings, positional and CLS embeddings, transformer and output head, and region modulation. Fusion and prototypes are frozen. Usage and dual estimates are retained without updates, prototype resampling is disabled, and the auxiliary head is removed. The continuous projection is discarded. A new PQ projection and private codebooks are initialized as specified in Appendix~\ref{app:quantization}. These, the concept embeddings, and decoder learn through global distillation. Region modulation remains trainable through orthogonality but does not enter the decoder's appearance path. A fresh optimizer and learning-rate schedule are used for stage 2.

Freezing assignment parameters does not make the training forward deterministic: both stages use Gumbel noise and batch-balanced selection. Evaluation removes the noise, uses the stored dual for independent selection, and performs hard PQ lookup. Table~\ref{tab:method-stages} summarizes the gradient and state changes.

\begin{table}[ht]
\centering
\caption{Canonical stage transition. The backbone and AnyUp weights are frozen in both stages, residuals are detached in both.}
\label{tab:method-stages}
\small
\begin{tabular}{@{}lll@{}}
\toprule
Component or operation & Stage 1 & Stage 2 \\
\midrule
Fusion and matching prototypes & Learned & Retained, frozen \\
Appearance encoder & Continuous projection & New private PQ \\
Concept embeddings and decoder & Learned & Retained, learned \\
Auxiliary distillation & Enabled & Disabled \\
Identity gradient surrogate & Enabled & Disabled \\
AnyUp feature-input gradients & Enabled & Disabled \\
Usage and selection dual & Updated & Retained, fixed \\
Dead-prototype resampling & Enabled & Disabled \\
\bottomrule
\end{tabular}
\end{table}

\myparagraph{Usage and prototype resampling.}
Balancing alone cannot restore a prototype that receives little useful learning signal. We therefore replace persistently underused prototypes with current region features. Stage 1 initializes $\rho_n=1/N$. After computing both views, selected-ID counts are summed across views and devices and divided by the total number of selected IDs, forming a frequency vector. Usage is updated as $0.99$ times its old value plus $0.01$ times this frequency vector. After each optimizer step, prototypes with $\rho_n<10^{-4}$ are replaced by normalized area-scaled region features drawn uniformly from the current batch's pooled features across both views (without replacement when the pool is large enough), scaled to the mean active-prototype norm, plus Gaussian noise of standard deviation $10^{-3}$. If all prototypes are inactive, use the whole-bank mean norm. Reset the replaced prototype's usage to $1/N$, its dual to zero, and its value embedding from the replacement prototype. Resetting individual dual entries can temporarily change the vector's mean. It is recentered at the next dual update. A common additive shift does not affect selection rankings. Optimizer moments are not reset. Under distributed training, one device samples from its own feature pool and broadcasts the changes. Appendix~\ref{abl:maintenance} measures the effect of disabling this replacement.

The augmentation, optimizer, schedules and checkpoint selection of \ours{} are given in Appendix~\ref{app:train-ours}, together with the schedules of every dataset.

\myparagraph{Order of operations.}
For each training batch, form the two views and compute their frozen targets and fused, upsampled features. Independently for each view, compute full-bank maps, balance mean scores, select concepts, recompute selected maps, pool residuals, and decode using continuous or quantized attributes. Sum the weighted losses with the view accounting in Table~\ref{tab:method-losses}, backpropagate, clip gradients, and take one optimizer step. In stage 1, update usage from both views and estimate the selection dual from the original view during the forward computation. After the optimizer step, resample inactive prototypes. In stage 2 these assignment states remain fixed. At evaluation, a single view preprocessed according to Appendix~\ref{app:evaluation} supplies ordinary soft maps, dual-based selection, hard ownership, and hard attribute codes. Because balancing and several regularizers use local batches, distributed execution is not equivalent to a single solve on the concatenated batch.

The objective-removal studies in Appendix~\ref{abl:objectives} distinguish preservation of downstream information from spatial and semantic organization.

\providecommand{\sidblue}{\iftoggle{comments}{\color[HTML]{1F5FBF}}{}}
\section{Experimental protocol}
\label{app:protocol}
This section specifies the datasets, backbones, training procedures and evaluation metrics used in Section~\ref{sec:experiments} and Appendices~\ref{app:ablations}--\ref{app:qualitative}, including method-specific exceptions.

\subsection{Datasets and splits}
\label{app:datasets}

Table~\ref{tab:app-datasets} lists ImageNet-1k~\citep{deng2009imagenet}, Places365-Standard~\citep{zhou2017places}, PartImageNet~\citep{he2022partimagenet}, CUB-200-2011~\citep{wah2011caltech}, Stanford Cars~\citep{krause2013collecting}, Stanford Dogs~\citep{khosla2011novel} and Oxford Flowers~\citep{nilsback2008automated}. We train one \ours{} model per dataset without class labels or part annotations. We use class labels to fit classification probes and label the kNN reference set. Keypoint annotations train the keypoint regressors, and part annotations support concept evaluation. Supervised baseline training and selection among evaluated configurations are described in Section~\ref{app:baselines}.

We report results on the validation splits of ImageNet-1k, Places365 and PartImageNet and on the test splits of the other datasets. Appendix~\ref{app:qualitative-attrspace} additionally uses the PartImageNet test split. On Oxford Flowers, we merge the training and validation splits, following PDiscoFormer~\citep{aniraj2024pdiscoformer}.

\begin{table}[!htbp]
  \centering
  \caption{\textbf{Datasets.} Split sizes, available part annotations and corresponding results tables. Oxford Flowers uses the combined training and validation splits for training.}
  \label{tab:app-datasets}
  \small
  \setlength{\tabcolsep}{3.5pt}
  \begin{tabular}{@{}lrrrll@{}}
    \toprule
    Dataset & Classes & Training & Evaluation & Part annotation & Tables \\
    \midrule
    ImageNet-1k & 1000 & 1,281,167 & val 50,000 & none & \ref{tab:classification-full} \\
    Places365 & 365 & 1,803,460 & val 36,500 & none & \ref{tab:classification-full} \\
    PartImageNet & 158 & 20,466 & val 1,206 & 40 part masks & \ref{tab:pin-concepts}, \ref{tab:part-retrieval}, \ref{tab:retrieval-full} \\
    CUB-200-2011 & 200 & 5,994 & test 5,794 & 15 keypoints & \ref{tab:fine-grained}, \ref{tab:cub-concepts}, \ref{tab:retrieval-full} \\
    Stanford Cars & 196 & 8,144 & test 8,041 & none & \ref{tab:fine-grained} \\
    Stanford Dogs & 120 & 12,000 & test 8,580 & none & \ref{tab:fine-grained} \\
    Oxford Flowers & 102 & 2,040 & test 6,149 & none & \ref{tab:fine-grained} \\
    \bottomrule
  \end{tabular}
\end{table}

\subsection{Backbones}
\label{app:backbones}

Table~\ref{tab:app-backbones} lists DINOv2~\citep{oquab2023dinov2} with registers~\citep{darcet2024registers}, DINOv3~\citep{simeoni2025dinov3}, TIPSv2~\citep{cao2026tipsv2}, CLIP~\citep{radford2021learning}, SigLIP2~\citep{tschannen2025siglip2} and CLIP-DINOiser~\citep{wysoczanska2024clipdinoiser}. In \ours{}, these feature extractors are frozen and receive $224\times224$ images. The input wrapper converts ImageNet-normalized images to each encoder's pretrained normalization, while AnyUp retains ImageNet-normalized guidance. We fuse patch tokens from several layers as described in Appendix~\ref{app:features}, except for CLIP-DINOiser, which directly supplies its refined projected features. The global distillation targets are listed in the table.

AnyUp~\citep{wimmer2026anyup} increases the native $16\times16$ grid of patch-14 models to $32\times32$ and the $14\times14$ grid of patch-16 models to $28\times28$. Baseline input resolutions, trainable tokens and scoring grids have the exceptions described in Sections~\ref{app:baselines} and~\ref{app:evaluation}.

The main controlled comparison uses DINOv2-B/14, with CFM evaluated on its released CLIP-DINOiser dictionary. Figure~\ref{fig:pin-full-heatmap-backbones} covers seven configurations of six feature extractors because SigLIP2 has two targets. We use CLIP ViT-L/14 for the MSAE reference on its original backbone (Figure~\ref{fig:pin-full-heatmap-msae} and Table~\ref{tab:cub-concepts}).

CLIP-DINOiser combines a frozen CLIP encoder with the authors' released $1.8$M-parameter adapter\footnote{\url{https://github.com/wysoczanska/clip_dinoiser}}. Two convolutions predict patch affinities that pool dense CLIP features into the denoised $512$-dimensional map used by \ours{}. Its encoder is the listed OpenCLIP checkpoint~\citep{cherti2023reproducible}, trained on LAION-2B~\citep{schuhmann2022laion}. The separate CLIP ViT-B/16 row uses OpenAI weights.

\begin{table}[!htbp]
  \centering
  \caption{\textbf{Feature extractors and targets.} Checkpoints are Hugging Face Hub identifiers. The grid is the assignment grid of \ours{}, or the scoring grid of the CLIP-L/14 MSAE reference. Targets are the frozen global features distilled by \ours{} at $224\times224$. A dash indicates a baseline-only encoder.}
  \label{tab:app-backbones}
  \scriptsize
  \setlength{\tabcolsep}{3pt}
  \begin{tabular}{@{}lllp{0.19\linewidth}@{}}
    \toprule
    Backbone & Checkpoint & Grid & Target \\
    \midrule
    DINOv2-B/14, registers & \texttt{facebook/dinov2-with-registers-base} & $32\times32$ & CLS token \\
    DINOv3-B/16 & \texttt{timm/vit\_base\_patch16\_dinov3.lvd1689m} & $28\times28$ & CLS token \\
    TIPSv2-B/14 & \texttt{google/tipsv2-b14} & $32\times32$ & CLS token \\
    CLIP ViT-B/16 & \texttt{openai/clip-vit-base-patch16} & $28\times28$ & CLS token after the last layer norm \\
    SigLIP2-B/16 & \texttt{google/siglip2-base-patch16-224} & $28\times28$ & attention-pooler output, or mean patch token \\
    CLIP-DINOiser & \texttt{laion/CLIP-ViT-B-16-laion2B-s34B-b88K} & $28\times28$ & projected CLIP image embedding \\
    CLIP ViT-L/14 & \texttt{openai/clip-vit-large-patch14} & $32\times32$ & \textemdash \\
    \bottomrule
  \end{tabular}
\end{table}

\subsection{Training \ours{}}
\label{app:train-ours}

The main DINOv2 experiments use the architecture, losses and two-stage curriculum of Appendix~\ref{app:method}, with $N=1024$ concepts, $K=16$ selected concepts per image and concept-specific PQ with $G=16$ groups of $M=64$ codewords. Schedules and batch sizes vary across datasets, and Places365 uses mixed upsampling in stage~2.

\myparagraph{Augmentation.} Each training image produces two views. The first combines crop and photometric augmentations commonly used in self-supervised vision~\citep{chen2020simclr,caron2021emerging} to diversify appearance. The second applies a random affine transform to the normalized first view, following the affine view pairing used by PDiscoNet and PDiscoFormer~\citep{van2023pdisconet,aniraj2024pdiscoformer}. We draw one transform per local batch, providing known spatial correspondences for $\mathcal L_{\mathrm{eq}}$. Table~\ref{tab:app-augmentation} specifies our recipe.

\begin{table}[!htbp]
  \centering
  \caption{\textbf{Augmentations of \ours{}.} The first-view transformations are applied in the listed order. The second view is an affine transform of the normalized first view.}
  \label{tab:app-augmentation}
  \scriptsize
  \setlength{\tabcolsep}{3pt}
  \def\BL{\raggedright\arraybackslash\leavevmode}%
  \begin{tabular}{@{}>{\BL}p{0.19\linewidth}>{\BL}p{0.10\linewidth}>{\BL}p{0.66\linewidth}@{}}
    \toprule
    Transformation & Probability & Parameters \\
    \midrule
    \multicolumn{3}{@{}l}{\textit{First view}} \\
    Random resized crop & $1$ & scale $[0.64,1]$, ratio $[3/4,4/3]$, $224\times224$, bicubic \\
    Horizontal flip & $0.5$ & \\
    Color jitter & $0.8$ & brightness and contrast $0.4$, saturation $0.2$, hue $0.1$ \\
    Grayscale & $0.2$ & \\
    Gaussian blur & $0.5$ & kernel $9$, $\sigma\in[0.1,2]$ \\
    Normalization & $1$ & mean $(0.485,0.456,0.406)$, std $(0.229,0.224,0.225)$ \\
    \midrule
    \multicolumn{3}{@{}l}{\textit{Second view}} \\
    Affine transform & $1$ & rotation $[-90^\circ,90^\circ]$, translation $\pm10\%$, scale $[0.7,1.3]$, no shear, bilinear interpolation, zero fill in normalized units \\
    \bottomrule
  \end{tabular}
\end{table}

\myparagraph{Optimization.} Tables~\ref{tab:app-optimization} and~\ref{tab:method-schedules} specify optimization and dataset-specific schedules. Both views are balanced separately within each local batch and jointly contribute to one optimizer update. Training loaders shuffle and drop incomplete batches. Learning rates are not automatically scaled with batch size.

\begin{table}[!htbp]
  \centering
  \caption{\textbf{Optimization and checkpoint selection of \ours{}.} Schedules and per-device batch sizes are in Table~\ref{tab:method-schedules}.}
  \label{tab:app-optimization}
  \scriptsize
  \setlength{\tabcolsep}{3pt}
  \def\BL{\raggedright\arraybackslash\leavevmode}%
  \begin{tabular}{@{}>{\BL}p{0.24\linewidth}>{\BL}p{0.70\linewidth}@{}}
    \toprule
    Setting & Value \\
    \midrule
    Optimizer & AdamW~\citep{loshchilov2019decoupled}, $(\beta_1,\beta_2)=(0.9,0.999)$, $\epsilon=10^{-8}$ \\
    Weight decay & $10^{-4}$, on all trainable parameters \\
    Learning-rate schedule & linear warm-up per epoch from $10^{-4}\times$ the peak, then cosine decay~\citep{loshchilov2017sgdr} to $10^{-6}$ \\
    Devices and accumulation & one GPU, or four H100s for ImageNet-1k and Places365. No gradient accumulation \\
    Precision and clipping & bfloat16 mixed precision and gradient-norm clipping at $1$ \\
    Seed & $42$ for main experiments. Ablation replication is specified in Appendix~\ref{app:ablations} \\
    Checkpoint selection & lowest validation $\mathcal L_{\mathrm{global}}$ among checkpoints saved every $10$ epochs (stage~1) or every epoch (stage~2). Stage~2 starts from the selected stage-1 checkpoint \\
    \bottomrule
  \end{tabular}
\end{table}

\begin{table}[t]
\centering
\caption{\textbf{Training schedules of \ours{} with DINOv2-B/14 and width 768.} The PartImageNet rows give the canonical schedule. Batch size is per device.}
\label{tab:method-schedules}
\small
\begin{tabular}{@{}lrrrr@{}}
\toprule
Stage & Epochs & Warm-up & Peak LR & Batch \\
\midrule
\multicolumn{5}{@{}l}{\textit{PartImageNet}} \\
Continuous concept discovery & 150 & 10 & $3\cdot10^{-4}$ & 16 \\
Concept-specific PQ, $16\times64$ & 200 & 10 & $10^{-3}$ & 16 \\
\midrule
\multicolumn{5}{@{}l}{\textit{CUB-200-2011, Stanford Cars, Stanford Dogs, Oxford Flowers}} \\
Continuous concept discovery & 200 & 40 & $3\cdot10^{-4}$ & 16 \\
Concept-specific PQ, $16\times64$ & 500 & 50 & $10^{-3}$ & 16 \\
\midrule
\multicolumn{5}{@{}l}{\textit{ImageNet-1k, Places365}} \\
Continuous concept discovery & 50 & 10 & $3\cdot10^{-4}$ & 64 \\
Concept-specific PQ, $16\times64$ & 100 & 10 & $10^{-3}$ & 64 \\
\bottomrule
\end{tabular}
\end{table}

\myparagraph{Mixed upsampling on Places365.} To reduce computation, Places365 uses AnyUp in a random $20\%$ of stage-2 batches and bilinear interpolation otherwise. The other main training runs and downstream evaluation use AnyUp throughout. Appendix~\ref{abl:upsampling} compares these alternatives.

\subsection{Training the baselines}
\label{app:baselines}

We train the dataset-specific baselines on the training data in Section~\ref{app:datasets}, using the authors' implementations where available. The main comparison uses DINOv2-B/14. Exceptions include the released CFM dictionary, released PDiscoFormer checkpoints and the MSAE reference on CLIP-L/14. Table~\ref{tab:app-baselines} summarizes these models and their probe inputs. The vision-language and supervised reference entries in Figure~\ref{fig:classification-user-study} (left) and Table~\ref{tab:classification-full} reproduce results reported by CFM~\citep{wittenmayer2026cfm} and ProtoQuant~\citep{janusz2026protoquant}.

\begin{table}[!htbp]
  \centering
  \caption{\textbf{Baselines in the main comparison.} Size denotes the global dictionary or prototype count, with PCA specified per image or class. Grid denotes the scoring resolution. The notation $16\!\to\!32$ indicates bilinear resampling from the native $16\times16$ maps to $32\times32$. Probe inputs are used for LP and kNN.}
  \label{tab:app-baselines}
  \scriptsize
  \setlength{\tabcolsep}{3pt}
  \def\BL{\raggedright\arraybackslash\leavevmode}%
  \begin{tabular}{@{}>{\BL}p{0.17\linewidth}>{\BL}p{0.14\linewidth}>{\BL}p{0.10\linewidth}>{\BL}p{0.15\linewidth}>{\BL}p{0.14\linewidth}>{\BL}p{0.20\linewidth}@{}}
    \toprule
    Method & Code & Class labels & Size & Grid & Probe input \\
    \midrule
    TopK SAE & ours & no & 2048 latents & $16\!\to\!32$ & mean-pooled reconstruction \\
    PatchSAE & authors' & no & 49,152 latents & $16\!\to\!32$ & mean-pooled reconstruction \\
    MSAE & authors' & no & 8192 latents (fine-grained: 6144) & $16\!\to\!32$ & mean-pooled reconstruction \\
    CFM & released & no & 8192 latents & $28\times28$ & max- and mean-pooled activations \\
    PIP-Net, stage 1 / 2 & authors' & stage 2 & 768 prototypes & $32\times32$ (AnyUp) & pooled prototype scores \\
    ProtoQuant, stage 1 / 2 & authors' & stage 2 & 2048 codes (CUB: 4000) & $16\!\to\!32$ & pooled prototype scores \\
    PDiscoFormer & released or retrained & yes & $K$ parts + background & $16\times16$ (CUB: $37\!\to\!32$) & mean part embedding \\
    PCA, per image / class-wise & ours & class-wise only & 16 per image or class & $16\!\to\!32$ & none \\
    \bottomrule
  \end{tabular}
\end{table}

\myparagraph{Sparse autoencoders.} TopK SAE~\citep{gao2025scaling} is our re-implementation. PatchSAE~\citep{lim2025sparse}\footnote{\url{https://github.com/dynamical-inference/patchsae}} and MSAE~\citep{zaigrajew2025interpreting}\footnote{\url{https://github.com/WolodjaZ/MSAE}} use the authors' code and hyperparameters (Table~\ref{tab:app-baseline-training}). These models are fitted to frozen DINOv2-B/14 patch tokens for the main comparison. Following the MSAE authors, we remove its TopK constraint at inference.

On PartImageNet, we sweep dictionary sizes and, for MSAE, both loss weightings and the three published expansion factors. The main table presents the best evaluated configuration of each SAE by mean rank over the reported columns of Table~\ref{tab:pin-concepts}. For the fine-grained table, we select one MSAE configuration jointly across all four datasets, requiring an available LP result on each dataset and averaging ranks over the ten reported metrics. In both selections, metrics are ranked in their preferred direction on unrounded values, missing metrics rank last, and exact ties follow the fixed candidate order. Candidates without a foreground-purity result are excluded on PartImageNet. A tie in the aggregate rank is also resolved by candidate order. These are best-of-sweep references selected using the reported validation or test scores, including annotated-part metrics where available. Representation fitting remains label-free, but this configuration selection uses evaluation labels. The selected MSAE sizes are 8192 on PartImageNet and 6144 on the fine-grained datasets with reverse and uniform weighting respectively. Figures~\ref{fig:pin-full-heatmap-ssl} and~\ref{fig:pin-full-heatmap-msae} show the evaluated alternatives.

\myparagraph{CFM.} CFM~\citep{wittenmayer2026cfm} is a Matryoshka sparse autoencoder trained by its authors on CC12M features of CLIP-DINOiser. We use its released dictionary\footnote{\url{https://github.com/kawi19/CFM}} without refitting it to our datasets.

\begin{table}[!htbp]
  \centering
  \caption{\textbf{Baseline training settings.} The table lists the recipes used for our retraining. MSAE runs for at least $21{,}000$ steps, extending beyond $30$ epochs on smaller datasets. CFM uses released weights and PCA is fitted directly to patch features.}
  \label{tab:app-baseline-training}
  \scriptsize
  \setlength{\tabcolsep}{3pt}
  \def\BL{\raggedright\arraybackslash\leavevmode}%
  \begin{tabular}{@{}>{\BL}p{0.14\linewidth}>{\BL}p{0.08\linewidth}>{\BL}p{0.16\linewidth}>{\BL}p{0.14\linewidth}>{\BL}p{0.14\linewidth}>{\BL}p{0.26\linewidth}@{}}
    \toprule
    Method & Optimizer & Learning rate & Batch & Length & Other settings \\
    \midrule
    TopK SAE & Adam & $3\cdot10^{-4}$ & 64 images & 20 epochs & 8 latents per patch, 16 per image, auxiliary loss, weight $1/32$ \\
    PatchSAE & Adam & $4\cdot10^{-4}$, 500 warm-up steps & 128 images & 20,480 steps & L1 weight $8\cdot10^{-5}$, ghost gradients \\
    MSAE & AdamW & $10^{-4}$ & 4096 tokens & 30 epochs, ${\geq}\,21{,}000$ steps & nested levels $64, 128, \dots$ up to the dictionary size, uniform or reverse weighting \\
    PIP-Net & AdamW & $5\cdot10^{-4}$, classifier $0.05$ & 128 / 64 & 10 + 60 epochs & stage 1 / stage 2, features upsampled by AnyUp \\
    ProtoQuant & AdamW & $0.05$ & 512 & 30 + 30 epochs & stage 1 / stage 2, temperature $0.1$ \\
    PDiscoFormer, retrained & Adam & $1.414\cdot10^{-6}$ base, halved every 4 epochs & 32 (PartImageNet: $4\times32$, accumulated) & 28 epochs & position embedding, \textsc{cls} and register tokens trained \\
    \bottomrule
  \end{tabular}
\end{table}

\myparagraph{PIP-Net and ProtoQuant.} Both use the authors' code\footnote{\url{https://github.com/M-Nauta/PIPNet}, \url{https://github.com/gmum/protoquant}.} with the training settings in Table~\ref{tab:app-baseline-training}, adapted to frozen DINOv2-B/14 features. Stage~1 is label-free and appears with the self-supervised methods. Stage~2 adds supervised classification and is an unranked reference. ProtoQuant's original recipe first fine-tunes the backbone on the target dataset. Our adaptation keeps that backbone frozen to match the feature extractor used by the main self-supervised comparison.

\myparagraph{PDiscoFormer.} Tables~\ref{tab:pin-concepts} and~\ref{tab:fine-grained} evaluate the authors' released checkpoints\footnote{\url{https://github.com/ananthu-aniraj/pdiscoformer}} for PartImageNet, CUB and Oxford Flowers. We additionally retrain the evaluated part counts on PartImageNet, including a frozen-token variant at $K=25$ (marked $^\dagger$ in Figure~\ref{fig:pin-full-heatmap-supervised}), and train $K=16$ models on Stanford Cars and Stanford Dogs, which have no released checkpoints. The retraining recipe updates the positional embeddings, CLS and register tokens while keeping the encoder blocks frozen, except in the explicitly frozen-token variant. Relative to the base rate in Table~\ref{tab:app-baseline-training}, learning rates are multiplied by $10^3$ for these tokens and by $10^4$ for prototypes, modulation and classifier parameters. Following the retraining procedure, we select the snapshot with the highest test accuracy. These supervised references therefore use the evaluation labels for checkpoint selection.

PDiscoFormer's PartImageNet recipe combines the training and validation splits, so the validation images in Table~\ref{tab:pin-concepts} were also seen with their labels during representation learning. Its released CUB checkpoints retain their $518\times518$ input resolution. Appendix~\ref{app:pdisco-repro} gives the retraining results, and Sections~\ref{app:evaluation} and~\ref{app:metrics} describe how these models are scored.

\myparagraph{PCA.} We include two PCA references using $16$ components per image, fitted to native $16\times16$ DINOv2-B/14 patch features. Each patch is assigned to the component with the largest absolute projection. Per-image PCA fits a new basis for each image, so its component identities have no shared meaning across images. Class-wise PCA fits one basis per training class and selects the basis using the evaluation image's ground-truth class. Each class has distinct global component IDs, giving $16$ times the number of classes in the full bank. This ground-truth-conditioned representation makes a classification probe circular, so we report no downstream probe for it.

On PartImageNet, PCA also estimates foreground from the sign of the first component. We orient its sign so that the mean projection of the four corner patches is nonpositive and retain patches with positive projection. On CUB, all patches are retained. This heuristic changes PCA's spatial support and is part of the baseline definition.

\subsection{Evaluation protocol}
\label{app:evaluation}

For locally evaluated models, we use the metric implementations in Section~\ref{app:metrics} on the stated evaluation splits. The default budget is $K=16$ concepts per image, with exceptions indicated by the $K/N$ column. Published reference scores retain their original protocols. We disable the test-time Gumbel noise in PDiscoFormer's released implementation to make its concept maps deterministic.

\myparagraph{Inputs.} Our classification and image-to-image retrieval evaluations use a short-side resize to $256$ followed by a $224\times224$ center crop. PIP-Net and ProtoQuant use the same crop-based probing pipeline. The dictionary baselines and PDiscoFormer are probed on cached features from square-resized images, preserving the protocol used to obtain their reported results. Thus using the same probe architecture does not imply identical input geometry or augmentation.

All reported mask and keypoint metrics use paired square resizing of images and annotations. The common input size is $224\times224$, with nearest-neighbor interpolation for masks and corresponding coordinate scaling for keypoints. This retains the full annotated object and aligns spatial evaluation across methods. It follows PIP-Net's full-image $224\times224$ protocol and $32\times32$ purity windows~\citep{nauta2023pip}. For the PDiscoFormer part-discovery and CFM grounding metrics~\citep{aniraj2024pdiscoformer,wittenmayer2026cfm}, we adopt their metric definitions in this common geometry. In particular, our square-resized PartImageNet evaluation differs from PDiscoFormer's released center-crop evaluation.

Released PDiscoFormer CUB checkpoints retain the authors' $518\times518$ input size, preserving their learned spatial configuration. We scale the purity window to $74\times74$ to keep its relative size close to $32/224$. Keypoint regression uses image-normalized coordinates for every method, as defined in Section~\ref{app:metrics}.

\myparagraph{Scoring grids.} Table~\ref{tab:app-baselines} lists the baseline grids. \ours{} and PIP-Net assign concepts on AnyUp's $32\times32$ DINOv2-B/14 feature grid. The sparse autoencoders, ProtoQuant and PCA produce native $16\times16$ maps, which we resample bilinearly to $32\times32$ for scoring. CFM retains a $28\times28$ grid, and PDiscoFormer retains $16\times16$ on PartImageNet. Its $37\times37$ CUB maps are resampled to $32\times32$.

Clustering metrics and part retrieval reduce annotations to these grids. Segmentation metrics instead upsample the concept maps to the image grid, and keypoint regression retains continuous target coordinates. The main tables therefore include resolution differences for CFM and PDiscoFormer, which can affect boundary-sensitive scores. Retrieval rows with a different region set are marked $^{\ddagger}$ and accompanied by their own oracle.

\myparagraph{Concept budget.} TopK SAE is trained with at most $16$ concepts per image, and its reconstruction retains this budget for downstream probing. PatchSAE, MSAE and CFM retain each image's $16$ strongest concepts for spatial metrics and part retrieval. Their LP, kNN and image-to-image retrieval instead use the unbudgeted representations listed in Table~\ref{tab:app-baselines}. Figure~\ref{fig:pin-full-heatmap-ssl} includes other budgets and the unrestricted CFM dictionary used in its published evaluation.

\myparagraph{Probing representations.}\label{app:pin-protocol} Our probes use the predicted global representation $\hat c(x)$, which preserves the target definition for each backbone in Table~\ref{tab:app-backbones}. Baseline representations are listed in Table~\ref{tab:app-baselines}. We evaluate the representation passed downstream by each method.

\begin{enumerate}
  \item \textbf{Linear probe (LP).} We apply batch normalization without affine parameters, followed by a linear classifier trained on the training split. Optimization uses SGD with momentum $0.9$, no weight decay and learning rate $0.1\cdot B_{\mathrm{global}}/256$, where $B_{\mathrm{global}}$ is the global batch size. Training lasts $50$ epochs, with $10$ epochs of warm-up followed by cosine decay.
  For \ours{}, PIP-Net and ProtoQuant, training images are re-encoded under random resized crops and horizontal flips. Dictionary baselines and PDiscoFormer use precomputed vectors without probe-time augmentation. Cached dictionary probes use batches of $256$. These choices preserve the recorded evaluation pipelines, so LP comparisons include differences in augmentation, image geometry and optimizer step counts.

  \item \textbf{kNN.} We predict the class with the largest weighted vote among the $20$ nearest training images under cosine similarity. A neighbor with similarity $s$ contributes weight $\exp(s/0.07)$.
\end{enumerate}

\myparagraph{Fine-grained datasets.}\label{app:finegrained-protocol} Table~\ref{tab:fine-grained} reports LP and kNN accuracy on all four datasets, together with PIP purity and keypoint regression error on CUB. CUB provides annotations for $15$ keypoints. Stanford Cars, Stanford Dogs and Oxford Flowers have no part annotations.

\subsection{Metrics}
\label{app:metrics}

\myparagraph{LP and kNN.} We report top-1 classification accuracy on the evaluation split using the probes in Section~\ref{app:pin-protocol}.

\myparagraph{Image retrieval.} Images are represented by the same frozen vectors used for probing. Each evaluation image queries the other evaluation images by cosine similarity, and a retrieved image is relevant when it shares the query's class. Recall at rank $r$, denoted R@$r$, is the fraction of queries with at least one relevant result among the first $r$ candidates. For each query, average precision (AP) averages the precision at ranks containing relevant results over the full ranking. Mean average precision (mAP) averages AP over queries. For image retrieval, queries with no other image of the same class are excluded from both recall and mAP.

\myparagraph{Foreground purity.} We reduce each mask to the scoring grid by a majority vote per cell. Let $C_{nj}$ count cells assigned to global concept $n$ with ground-truth label $j$, including the background label where it is annotated. Let $\mathcal J_{\mathrm{fg}}$ be the foreground part labels and $N_{\mathrm{fg}}$ the number of annotated foreground cells in the evaluated split. Then
\begin{equation*}
  \mathrm{Pur}_{\mathrm{fg}}=\frac{1}{N_{\mathrm{fg}}}\sum_n\max_{j\in\mathcal J_{\mathrm{fg}}}C_{nj}.
\end{equation*}
A foreground cell left unassigned contributes to $N_{\mathrm{fg}}$ but to no concept count, so rejecting foreground cannot improve purity by reducing its denominator. On CUB, counts use the grid cells containing visible keypoints.

\myparagraph{NMI, ARI and their mapped variants.} We compare the ground-truth part labels $y$ with concept identities $c$ over the annotated foreground cells of the evaluated split, following part-discovery evaluation~\citep{choudhury2021unsupervised,van2023pdisconet,aniraj2024pdiscoformer}. Unassigned foreground cells receive a separate cluster label. NMI normalizes mutual information by the arithmetic mean of the two entropies,
\begin{equation*}
  \mathrm{NMI}=\frac{2I(y;c)}{H(y)+H(c)}.
\end{equation*}
The Rand index (RI) measures the fraction of cell pairs that both labelings place together or both place apart. ARI corrects this agreement for chance,
\begin{equation*}
  \mathrm{ARI}=\frac{\mathrm{RI}-\mathbb E[\mathrm{RI}]}{1-\mathbb E[\mathrm{RI}]},
\end{equation*}
where the expectation assumes random assignments with fixed cluster sizes, giving chance an expected score of zero.

Raw NMI and ARI penalize splitting one part across several pure concepts. For NMI$_m$ and ARI$_m$, we first map each concept to its dominant label on the evaluated split, $\arg\max_j C_{nj}$, including background where annotated. We then compute NMI and ARI on the mapped labels, retaining a separate cluster for unassigned cells. On PartImageNet, this mapping uses at most $41$ semantic labels, comprising $40$ parts and background, in addition to the unassigned cluster. Mapping removes the penalty for splitting a part into concepts that receive the same label. The mapping is fitted on the evaluated split, so these are descriptive clustering scores.

\myparagraph{mIoU and FG-ARI.} Our mIoU~\citep{he2022partimagenet} is a per-image matched-part score. We upsample the $K$ concept maps to input resolution, assign each pixel to its strongest concept, and use Hungarian one-to-one matching to annotated parts. IoU is averaged over matched pairs, then images, excluding unmatched parts. This differs from dataset-level semantic mIoU. FG-ARI~\citep{locatello2020objectcentric} averages per-image ARI over annotated foreground pixels, rather than the split-wide grid cells used above.

\myparagraph{Locality, consistency and impurity.} We follow CFM's metric implementation~\citep{wittenmayer2026cfm}. We sort nonnegative map activations and find the cutoff where cumulative mass reaches $90\%$. All pixels at or above this cutoff are retained, including ties. Thus a uniform hard region retains its entire support. Part masks are dilated by a $3\times3$ maximum filter.

Locality takes the best IoU between each part and any concept support, averages over images containing that part, and then over the $40$ PartImageNet part labels. Consistency first averages each concept--part IoU over images in which either the concept or the part is present. It then selects the best concept for each part and averages over parts. Impurity measures natural-log entropy of the retained activation mass over the dilated label masks, including background, after normalizing those label masses. We average entropy over images within each concept, then over concepts active in at least five images.

As an approximate implementation check under CFM's original center-crop/$14\times14$ protocol, its unrestricted released dictionary gives locality $44.3$, consistency $15.3$ and impurity $0.333$ (published: $44.3$, $15.3$, $0.330$). This differs from our common square-resize/$28\times28$ evaluation.

\myparagraph{PIP purity.} We adapt PIP-Net's top-image purity evaluation~\citep{nauta2023pip} to each method's concept maps on the CUB test split, merging left and right versions of each part.
\begin{enumerate}
  \item For each concept, retain up to $10$ images with its highest peak activations.
  \item In each image, center a $32\times32$-pixel window on the map's peak. If several cells share the maximum, choose the maximal cell nearest their centroid. Shift windows at image borders to retain their size. The released PDiscoFormer CUB models use the scaled $74\times74$ window described in Section~\ref{app:evaluation}.
  \item For each part, compute the fraction of retained images whose window contains its keypoint, using only images in which that part is visible.
  \item Take the largest part-specific fraction as the concept's purity, then average over concepts active in at least one image.
\end{enumerate}

\myparagraph{Kp NME.} Keypoint regression tests whether concept maps locate the $15$ CUB keypoints, following PDiscoNet and PDiscoFormer~\citep{van2023pdisconet,aniraj2024pdiscoformer}. Each concept map supplies its two centroid coordinates, normalized by image size, and its mass. Features are indexed by global concept identity, with zeros for an absent concept. We rank concepts by training-set usage and select the basis size from $\{16,32,64,128,256,512\}$ using a $20\%$ holdout drawn from the training split with seed $0$. Candidate sizes exceeding the available basis are skipped. If fewer than $16$ concepts are available, we use all of them. After selection, the regressor is refitted on the full training split. For each keypoint, ridge regression with regularization $10^{-3}$ is fitted using training images where that keypoint is visible.

Kp NME is $100$ times the mean Euclidean error over all visible test keypoints in image-normalized coordinates. Our PDiscoFormer entries in Tables~\ref{tab:fine-grained} and~\ref{tab:cub-concepts} are recomputed with this evaluator. The original paper's bounding-box-normalized errors are a different quantity.

\myparagraph{Part-to-part retrieval.} The benchmark in Section~\ref{sec:setup} compares regions through globally indexed concept identities.
\begin{enumerate}
  \item \emph{Regions.} On PartImageNet, each part label covering at least two cells of the scoring grid defines one region per image, giving $3{,}524$ regions at $32\times32$. On CUB, each visible keypoint defines a one-cell region. To bound the quadratic cost of ranking, we sample $8{,}000$ regions uniformly without replacement using seed $0$. The $224\times224$ CUB evaluation of \ours{} yields $68{,}224$ regions before sampling. Counts can differ slightly for the $518\times518$ PDiscoFormer inputs.
  \item \emph{Descriptors.} A region is represented by a histogram of the global concept IDs assigned to its cells, normalized to sum to one. For \ours{}, the histogram counts the globally indexed identities $p_u$.
  \item \emph{Ranking.} Each region queries regions from other images by cosine similarity, with random tie-breaking using seed $0$. The same region set supplies queries and gallery candidates.
  \item \emph{Scoring.} A retrieved region is relevant when it has the query's part label. R@$r$ and mAP follow the definitions above. Queries without a relevant gallery candidate contribute zero to recall and are excluded from mAP.
\end{enumerate}
Rows marked $^{\ddagger}$ in Tables~\ref{tab:part-retrieval} and~\ref{tab:retrieval-full} use different grids and region sets with their own oracles. CFM's $28\times28$ grid gives $3{,}466$ PartImageNet regions and oracle R@1 $44.5$. Part-to-image retrieval compares the region histogram with a whole-image concept histogram, also normalized to sum to one. It excludes the query image and treats an image as relevant when it contains the queried part.

\myparagraph{Oracle and chance.} For part retrieval, the category-only reference is implemented as a co-occurrence-group oracle. It ranks regions sharing the query's label group before other regions, with random ordering within both sets. We construct groups from the evaluated split by visiting part labels in label order. Each unassigned label starts a group and adds the remaining unassigned labels whose image-presence sets have Jaccard overlap of at least $0.6$ with that seed label. This is a greedy grouping around the seed, without transitive merging.

For part-to-image retrieval, each gallery image receives the group of its lowest-index retained part label, or a separate sentinel group if it has no retained parts. Images sharing the query's group are ranked first, with random ordering within both sets.

On PartImageNet, these groups capture supercategory co-occurrence, with $38$ of the $40$ part labels in $12$ groups. The construction also assigns groups to the remaining labels. This measures retrieval explained by co-occurrence without identifying the particular part. It does not use bird-species labels on CUB. Chance ranks all candidates randomly. Both reference scores average five random rankings. Classification and image-to-image oracle entries instead use the ground-truth image class directly.

\myparagraph{Concept contribution.} The qualitative explanations in Figure~\ref{fig:teaser}, Section~\ref{sec:qualitative} and Appendix~\ref{app:qualitative-ours} use a separate linear classifier on $\hat c(x)$. It is fitted to features standardized using training-set means and standard deviations. Training uses AdamW for $200$ epochs, learning rate $10^{-3}$ with cosine decay, weight decay $10^{-2}$, batches of $1024$ and label smoothing $0.1$. Its predicted class defines the logit used for explanations, while the quantitative LP results use the probe in Section~\ref{app:pin-protocol}.

We measure a concept's contribution by removing it from the discrete description and decoding the modified description again. This is an occlusion test~\citep{zeiler2014visualizing} at the concept level, related to concept interventions~\citep{koh2020concept}. The removed concept's cells are reassigned to the strongest remaining concept among the already selected IDs $i_{1:K}$. The concept embeddings and attribute codes remain fixed, so only the spatial identity map $p$ changes. We measure the drop in the original predicted-class logit and normalize the positive drops across the selected concepts to sum to one. These weights describe relative positive support under the intervention, rather than an additive decomposition of the classifier's logit.
\clearpage
\section{Ablation studies}
\label{app:ablations}

\myparagraph{Shared protocol.}
We organize the ablations around the components of Section~\ref{sec:architecture} and Appendix~\ref{app:method}.

To reduce computational cost, most component studies train the continuous first stage with bilinear feature upsampling instead of AnyUp.

Unless specified otherwise, the component control uses PartImageNet, frozen DINOv2-B/14 with registers, $224\times224$ inputs, a $32\times32$ assignment/decoder grid, $N=1024$, $K=16$, and a three-block width-768 decoder. It follows the 150-epoch stage-1 schedule with batch size 16 in Table~\ref{tab:method-schedules}. Each alternative changes the named component, retaining the remaining settings. Quantization uses a separate, equally trained continuation control, described in Appendix~\ref{abl:quantization}. Backbone and concept-budget comparisons specify their own references below.

\myparagraph{Reading the results.} Unless stated otherwise, gray reference rows report absolute scores and colored entries report variant-minus-reference differences. Percentage-valued metrics are shown on a 0--100 scale, and their differences are in percentage points (pp). Backbone impurity and quantization loss use raw units. Attribute rates are absolute bits per selected concept in every row. The backbone figure's final panel reports each model's LP and kNN differences from its own frozen target, without subtracting the DINOv2 reference again. Blue indicates improvement and red degradation, with separate color scales whose ranges are shown below each panel.

The fusion, upsampling, selection, appearance-removal, auxiliary-readout, decoder-depth, gradient, objective, and quantization comparisons report means over seeds 42, 43, and 44. Smaller numbers give sample standard deviations across absolute reference scores or paired variant-minus-reference differences. The backbone, bank-size, and concept-budget sweeps report single runs. We interpret small differences descriptively and do not infer significance from mean rankings. LP denotes linear-probe accuracy, and mAP denotes full-ranking image-retrieval mAP. CFM locality and consistency in the component panels use energy-threshold maps. Metric definitions, evaluation procedures, and training schedules are given in Appendices~\ref{app:metrics}, \ref{app:evaluation}, and~\ref{app:train-ours}, respectively.

\subsection{Feature extraction}
\label{abl:features}

\phantomsection
\myparagraph{Backbone and global target.}
\label{abl:backbones}
Figure~\ref{fig:pin-full-heatmap-backbones} evaluates the quantized model across frozen feature extractors. The sweep follows the canonical two-stage settings, with the backbone-specific feature, target, and grid adaptations described in Appendix~\ref{app:backbones} and Table~\ref{tab:app-backbones}. Decoder width remains 768, including for CLIP-DINOiser's 512-dimensional projected features. Boundary-sensitive scores should be compared within a grid.

DINOv2 retains LP accuracy within $0.1$ pp of its frozen target and achieves the strongest foreground purity in its grid group. DINOv3 leads most part-quality measures in the patch-16 group, although CLIP-DINOiser has higher CFM consistency ($-3.0$ versus $-3.7$ pp relative to DINOv2). Thus the metric ordering is not universal. SigLIP2's pooler improves LP by $2.7$ pp over GAP ($-2.4$ versus $-5.1$ pp relative to DINOv2), with mixed part-quality changes. The target definition therefore matters as well as the patch features. The final panel separates representation retention from each frozen target's own probe accuracy.

\input{\RepoFigureRoot/tables/heatmap_tables/fig_partimagenet_backbones_heatmap.tex}

\phantomsection
\myparagraph{Layer fusion.}
\label{abl:fusion}
Removing the shared-query fusion of Appendix~\ref{app:features} uses the final-block patch features alone. Figure~\ref{fig:ablation-fusion} shows essentially unchanged downstream performance ($+0.17$ pp LP and $-0.05$ pp mAP), but foreground purity falls by $3.16$ pp and mIoU by $1.44$ pp. Multi-layer fusion therefore contributes primarily to concept quality in this comparison, rather than increasing downstream accuracy.
\input{\RepoFigureRoot/tables/heatmap_tables/fusion_heatmap.tex}

\phantomsection
\myparagraph{Upsampling.}
\label{abl:upsampling}
Figure~\ref{fig:ablation-upscaling} uses continuous AnyUp training as its reference. Replacing image-guided upsampling with bilinear interpolation at the same $32\times32$ resolution changes LP by $+0.22$ pp and mAP by $-0.03$ pp. Bilinear improves foreground purity by $2.07$ pp but lowers mIoU and locality by $1.39$ and $1.47$ pp. These mixed differences support its use as a cheaper component-ablation control without establishing equivalence on every metric. The canonical method retains AnyUp for its slightly more detailed qualitative maps.

\myparagraph{Native resolution and mixed-route training.}
Keeping the native $16\times16$ grid removes upsampling and changes the decoder grid accordingly. Relative to AnyUp, LP changes by $-0.04$ pp, but foreground purity falls by $6.55$ pp, while FG-ARI rises by $2.42$ pp, illustrating sensitivity to spatial granularity. The mixed alternative selects AnyUp with probability $0.2$ and bilinear interpolation otherwise, sharing the draw across both views of a local batch, and uses full AnyUp for evaluation. It remains close in LP and mAP ($+0.08$ and $-0.05$ pp) and improves foreground purity by $2.17$ pp in Figure~\ref{fig:ablation-upscaling}. Expensive image-guided upsampling on every training batch is therefore not required to retain the measured downstream performance.
\input{\RepoFigureRoot/tables/heatmap_tables/upscaling_heatmap.tex}

\FloatBarrier
\subsection{Concept bank, selection, and maintenance}
\label{abl:bank-sec}

\phantomsection
\myparagraph{Bank size.}
\label{abl:bank}
Figure~\ref{fig:ablation-bank} varies the global vocabulary $N$ from 1024 to 8192 at fixed $K=16$. The budget sweeps use continuous DINOv2-B/14 discovery with AnyUp on a $32\times32$ grid and the 150-epoch stage-1 schedule in Table~\ref{tab:method-schedules}, sharing the $N=1024,K=16$ reference. Each setting is a single completed run. Increasing $N$ to 8192 raises foreground purity by $4.05$ pp, but FG-ARI and mIoU fall by $1.85$ and $2.53$ pp, while consistency and locality fall by $10.47$ and $3.23$ pp. LP increases by $0.41$ pp. Thus higher purity with a larger vocabulary does not translate into better spatial segmentation or more consistent concepts across images. The canonical bank balances these competing properties. Interpret purity alongside consistency using the metric definitions in Appendix~\ref{app:metrics}, since finer fragmentation can increase purity.
\input{\RepoFigureRoot/tables/heatmap_tables/bank_heatmap.tex}

\phantomsection
\myparagraph{Concepts per image.}
\label{abl:budget}
Figure~\ref{fig:ablation-k} varies $K$ from 4 to 64 at fixed $N=1024$. The $K=32$ and $K=64$ configurations also increase the equivariance-map budget to 64 and 128, respectively, so these comparisons vary both budgets. The treatment of unmatched parts in mIoU follows Appendix~\ref{app:metrics}. With $K=4$, LP remains close to the reference ($+0.50$ pp), but FG-ARI and mIoU fall by $14.37$ and $15.31$ pp, and consistency and locality fall by $12.49$ and $16.00$ pp. Increasing $K$ beyond 16 also fails to improve these concept metrics. At $K=32$, FG-ARI, mIoU, and foreground purity fall by $1.83$, $2.17$, and $3.84$ pp. The per-image budget controls the spatial decomposition, and downstream accuracy alone does not identify a useful concept budget. At $N=1024$, $K=16$ gives the strongest reported spatial and semantic scores among the tested values.
\input{\RepoFigureRoot/tables/heatmap_tables/k_heatmap.tex}

\begin{figure}[!ht]
\centering
\includegraphics[width=\textwidth]{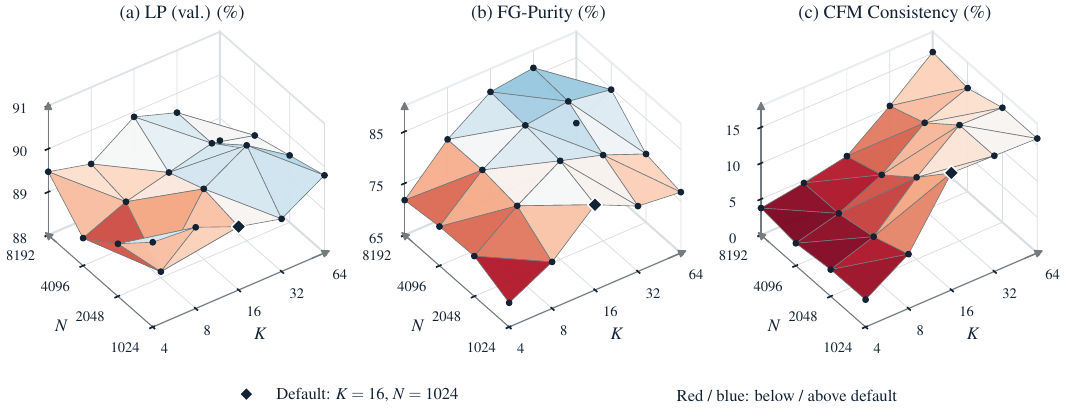}
\caption{Joint sweep of selected concept count $K$ and bank size $N$ on PartImageNet with DINOv2-B/14, using the continuous AnyUp setup and equivariance budgets in Appendix~\ref{abl:bank-sec}. Heights give absolute LP validation accuracy, foreground purity, and CFM consistency in percent. Vertical ranges differ across panels. Both horizontal axes use logarithmic (base-2) spacing. Dots mark all 20 measured configurations and the diamond marks the default $(K,N)=(16,1024)$. Surfaces connect adjacent observations without smoothing. Red and blue indicate scores below and above the default, respectively, with color intensity scaled separately for each metric. Each configuration is a single completed run.}
\label{fig:ablation-bank-k-grid}
\end{figure}

\myparagraph{Joint bank and concept-budget sweep.}
Figure~\ref{fig:ablation-bank-k-grid} extends these two slices to all 20 combinations of $K\in\{4,8,16,32,64\}$ and $N\in\{1024,2048,4096,8192\}$. LP varies from $88.14\%$ to $90.22\%$, while foreground purity and CFM consistency vary more strongly. The highest observed purity is $86.60\%$ at $(K,N)=(32,8192)$, whereas the default $(16,1024)$ has the highest observed consistency ($16.30\%$). Increasing the bank size therefore does not uniformly improve concept quality, and the effect depends on the selected concept budget. These single-run comparisons are descriptive.

\phantomsection
\myparagraph{Peak rather than mean scoring.}
\label{abl:selection}
The canonical selector averages each full-bank map over space (Appendix~\ref{app:selection}). The peak alternative instead takes its maximum, retaining Sinkhorn balancing and all other settings. Figure~\ref{fig:ablation-selection} shows small downstream changes ($+0.06$ pp LP, $+0.05$ pp mAP) but a $3.45$ pp decrease in foreground purity. Selecting concepts by their most active cell does not improve the overall region semantics in this setting.

\myparagraph{Coverage bias and direct mass selection.}
Setting the coverage-bias coefficient to zero retains Sinkhorn balancing and changes foreground purity by $-0.79\pm1.46$ pp, with little mean change in FG-ARI or mIoU (Figure~\ref{fig:ablation-selection}). These small differences relative to seed variation do not establish an independent benefit of the coverage term. Appendix~\ref{app:selection} explains its cancellation under ideal balancing. To compare the selectors, direct top-$K$ selection by mean activation is contrasted with the zero-bias Sinkhorn row. LP differs by $+0.03$ pp, while consistency and locality fall by $1.38$ and $1.63$ pp.

\phantomsection
\myparagraph{Prototype resampling.}
\label{abl:maintenance}
Disabling the inactive-prototype replacement of Appendix~\ref{app:curricula}, while retaining usage tracking and selection, lowers mAP by $7.51$ pp, foreground purity by $13.43$ pp, and consistency by $8.52$ pp (Figure~\ref{fig:ablation-selection}). These are substantially larger changes than those from the scoring alternatives. Prototype replacement supports both retrieval and concept quality.
\input{\RepoFigureRoot/tables/heatmap_tables/selection_heatmap.tex}

\FloatBarrier
\subsection{Appearance attributes and quantization}
\label{abl:attributes-sec}

\phantomsection
\myparagraph{Attributes during concept discovery.}
\label{abl:attributes}
Removing the continuous appearance contribution sets $q_{i_k}=0$ in the decoder of Appendix~\ref{app:decoder}, while learning concepts and retaining identity, geometry, and the auxiliary branch. Figure~\ref{fig:ablation-attributes} shows losses of $8.98$ pp LP and $6.22$ pp mAP, although FG-ARI and mIoU rise slightly. Concept identities and geometry alone do not retain the same downstream information. This discovery-stage experiment differs from removing attributes after the concept bank has been learned, considered next.
\input{\RepoFigureRoot/tables/heatmap_tables/attributes_heatmap.tex}

\phantomsection
\myparagraph{Quantization protocol and rate.}
\label{abl:quantization}
Figure~\ref{fig:ablation-quantisation} compares attribute encoders with the same seed-matched bilinear stage-1 parent. Each arm freezes concept assignment, retains the decoder, and receives the same 200-epoch continuation under the PartImageNet stage-2 schedule in Table~\ref{tab:method-schedules}. The continuous arm retains its learned projection, while the discrete arms initialize fresh quantization modules. The reference is private PQ with $G=16$ groups and $M=64$ entries per group, as defined in Appendix~\ref{app:quantization}. A private codebook is indexed by concept identity. A shared codebook serves all concepts. Rates count only attribute indices per selected concept, excluding identities, geometry, and model parameters. All discrete alternatives use nearest-codeword forward values and differentiable soft-mixture surrogates. Their codebooks and input projections are learned through distillation. These arms compare codebook structures, using our common training objective and surrogate rather than the original VQ-VAE or SoundStream training losses.

\myparagraph{Continuous and absent attributes.}
Continuing with continuous attributes instead of PQ improves LP and kNN by $0.46$ and $0.71$ pp, while mAP falls by $0.31$ pp (Figure~\ref{fig:ablation-quantisation}). Both arms receive the same continuation, controlling for additional training time. Removing attributes from this frozen-assignment bilinear control lowers LP by $12.24$ pp and mAP by $11.14$ pp relative to private PQ. Both comparisons support retaining appearance information.

\myparagraph{Private low-rate quantizers.}
Vector quantization (VQ) represents a full vector by a single codeword~\citep{linde1980algorithm,oord2017neural}. Private VQ64 selects one full-width vector from 64 entries (6 bits per concept). Private PQ2$\times$8~\citep{jegou2011product} splits the projected residual into two channel groups with eight entries each, also using 6 bits. Relative to private PQ16$\times$64, they lose $1.69$ and $2.05$ pp LP, respectively, and both increase distillation loss by approximately $0.05$ (Figure~\ref{fig:ablation-quantisation}). Very compact attributes retain much more information than removing attributes, but incur a larger cost than the 96-bit reference. These comparisons jointly change the attribute-index rate and quantizer structure and capacity, rather than isolating bit rate alone.

\myparagraph{Shared 12-bit quantizers.}
Shared VQ4096 selects a single full-width vector. Residual vector quantization (RVQ), as used in SoundStream~\citep{zeghidour2022soundstream}, instead quantizes successive residuals. Shared RVQ64+64 sums a first 64-entry codeword and a second 64-entry codeword fitted to the remaining residual. Shared PQ2$\times$64 concatenates two channel-group codewords~\citep{jegou2011product}. Each uses 12 bits per concept. Their LP changes are $-1.84$, $-2.23$, and $-2.41$ pp, respectively, with mAP changes of $-0.40$, $-0.89$, and $-0.94$ pp (Figure~\ref{fig:ablation-quantisation}). At this rate, full-vector VQ has the highest mean LP and mAP among these three alternatives, but all trail the higher-rate reference.

\myparagraph{Shared versus private product codebooks.}
Shared PQ16$\times$64 matches the reference's 96-bit attribute-index rate while sharing its codebooks across concepts. It changes LP by $-0.17$ pp, mAP by $+0.03$ pp, and kNN by $+0.07$ pp (Figure~\ref{fig:ablation-quantisation}). Thus private codebooks do not provide a uniform downstream advantage in this comparison. They allow concept-specific appearance dictionaries, at the parameter cost specified in Appendix~\ref{app:decoder}. Sharing is a competitive alternative when reducing codebook storage is important.
\input{\RepoFigureRoot/tables/heatmap_tables/quantisation_heatmap.tex}

\FloatBarrier
\subsection{Concept decoder}
\label{abl:decoder}

\myparagraph{Pooling instead of spatial attention.}
The zero-block alternative mean-pools the spatial tokens $t_u$ and applies the same normalization and output projection, omitting the CLS token and all attention blocks. Figure~\ref{fig:ablation-depth} shows losses of $1.26$ pp LP and $2.06$ pp mAP relative to the three-block decoder, with smaller concept-quality changes. This is the pooling-only alternative to Appendix~\ref{app:decoder}. The attention blocks improve integration of the spatial concept description.

\myparagraph{Transformer depth.}
Figure~\ref{fig:ablation-depth} varies depth from zero to six blocks at fixed width 768. Moving from one to three blocks improves LP by $0.51$ pp and mAP by $0.31$ pp. Increasing from three to six roughly doubles decoder parameters ($23.42$M to $44.68$M) for $0.11$ pp higher mean scores in each downstream metric, while consistency falls by $2.07$ pp. Three blocks offer a useful capacity trade-off. Deeper decoding does not uniformly improve concept quality.
\input{\RepoFigureRoot/tables/heatmap_tables/depth_heatmap.tex}

\myparagraph{Auxiliary-only readout.}
Removing the spatial decoder also removes its global loss and appearance branch. The pooled-region auxiliary head becomes the downstream representation, and checkpoints are selected by auxiliary validation distillation loss. Figure~\ref{fig:ablation-decoder} shows $-0.75$ pp LP and $+0.35$ pp mAP, with relatively small concept-quality changes. This system comparison changes both training and readout. Comparing the same auxiliary readout in both systems instead gives approximately $0.00\pm0.82$ pp LP and $-0.05\pm0.15$ pp mAP. Concepts can therefore be learned through auxiliary distillation and the assignment objectives alone. The auxiliary readout consumes continuous pooled region features. The stage-1 spatial decoder instead consumes concept identities, spatial assignments, and continuous appearance attributes, which are quantized in the final model.
\input{\RepoFigureRoot/tables/heatmap_tables/decoder_heatmap.tex}

\FloatBarrier
\subsection{Gradient routes}
\label{abl:gradients}

\myparagraph{Identity surrogate.}
Disabling Equation~\ref{eq:identity-ste} preserves hard forward identities but removes the decoder's gradient to soft assignments through identity embeddings. Other training objectives remain active. Figure~\ref{fig:ablation-gradients} shows $-0.28$ pp LP, $-0.79$ pp FG-ARI, and $-1.00$ pp consistency. The surrogate has modestly higher mean scores under this control, alongside the other sources of concept-learning gradients.

\myparagraph{Residual gradients.}
The canonical residual in Appendix~\ref{app:attributes} is detached. Restoring its derivative lets global distillation update pooled features and matching prototypes through the appearance projection as well as through the identity surrogate. LP and mAP change by $-0.08$ and $+0.04$ pp, while FG-ARI, mIoU, and foreground purity fall by $1.10$, $0.67$, and $0.69$ pp (Figure~\ref{fig:ablation-gradients}). These results do not motivate adding this gradient route to the canonical model.
\input{\RepoFigureRoot/tables/heatmap_tables/gradients_heatmap.tex}

\FloatBarrier
\subsection{Training objectives}
\label{abl:objectives}
We study the contribution of each auxiliary or regularization objective by setting one non-global loss weight in Table~\ref{tab:method-losses} to zero during continuous training, while retaining the global decoder loss and all remaining weights. Figure~\ref{fig:objective-differences} reports the resulting changes. The objectives and weights are defined in Appendix~\ref{app:objectives}. Across these ablations, several objectives affect region quality more strongly than downstream performance. When auxiliary distillation ($\mathcal L_{\mathrm{aux}}$) is removed, LP and mAP change little, but foreground purity and locality fall by $1.38$ and $1.51$ pp. The auxiliary signal supports region quality even when the retained spatial decoder preserves downstream information. Removing feature-weighted Potts regularization ($\mathcal L_{\mathrm{Potts}}$) lowers LP by only $0.12$ pp, but mIoU and consistency fall by $6.94$ and $7.91$ pp. Feature-aware spatial regularization chiefly supports concept organization rather than downstream accuracy. Removing $\mathcal L_{\mathrm{orth}}$ lowers LP by $0.11$ pp and both mIoU and consistency by $1.25$ pp. Its effects are modest relative to Potts, but generally favor retaining distinct modulated region features.

The assignment and cross-view objectives reveal trade-offs between foreground purity and spatial and grounding metrics. Without assignment entropy regularization ($\mathcal L_{\mathrm{ent}}$), FG-ARI and mIoU rise by $2.45$ and $1.65$ pp, and consistency and locality rise by $1.28$ and $2.16$ pp, while foreground purity falls by $0.87$ pp. The regularizer favors purity in this comparison, while several spatial and grounding metrics favor its removal. Similarly, without affine equivariance ($\mathcal L_{\mathrm{eq}}$), mAP is unchanged and FG-ARI rises by $0.34$ pp, but foreground purity and consistency fall by $4.76$ and $2.14$ pp. Cross-view agreement supports semantic stability beyond the quality of a single-image partition. Removing selected-region presence ($\mathcal L_{\mathrm{slot}}$) raises FG-ARI by $1.28$ pp and mIoU by $0.66$ pp but lowers foreground purity by $2.70$ pp. Local activation of selected regions favors purity, with mixed effects on partition metrics.

When usage diversity ($\mathcal L_{\mathrm{div}}$) is removed, LP changes by $+0.01$ pp, but foreground purity and locality fall by $7.04$ and $4.82$ pp. Regularizing full-bank activation therefore remains useful alongside balancing the subset-selection scores. In comparison, removing bank presence ($\mathcal L_{\mathrm{bank}}$) lowers foreground purity by $0.68$ pp and locality by $0.19$ pp.
\input{\RepoFigureRoot/tables/heatmap_tables/objective_heatmap.tex}
\clearpage
\providecommand{\sidblue}{\iftoggle{comments}{\color[HTML]{1F5FBF}}{}}
\clearpage
\section{Additional results}
\label{app:additional-results}

This section gives the full versions of the main-paper tables, in their order (Sections~\ref{app:classification-full}--\ref{app:retrieval-full}), explains how we visualize quantized attributes (Section~\ref{app:qualitative-quant}), and reports the user-study details and results (Section~\ref{app:user-study}).

\subsection{Large-scale classification in full}
\label{app:classification-full}

Table~\ref{tab:classification-full} lists every method of the comparison summarized in Figure~\ref{fig:classification-user-study} (left) of the main paper, with the same properties and the same ranking.

Figure~\ref{fig:classification-user-study} (left) keeps one supervised method and the three vision-language methods with the highest ImageNet accuracy, preferring methods that are also baselines elsewhere in the paper; Table~\ref{tab:classification-full} lists all of them. In the supervised block, InfoDisent and ProtoQuant train their heads on a frozen, label-supervised ViT-B/16. The other rows report a linear probe on the backbone or on the concept representation, except DCLIP, which scores classes by their descriptions~\citep{menon2023visual}. Following CFM~\citep{wittenmayer2026cfm}, LF-CBM~\citep{oikarinen2023label}, LaBo~\citep{yang2023language}, CDM~\citep{panousis2023sparse}, DCLIP, DN-CBM~\citep{rao2024discover} and D-CBM~\citep{prasse2024dcbm} learn global concepts, CF-CBM~\citep{panousis2024coarse} region-level ones, and InfoDisent~\citep{struski2024infodisent}, SALF-CBM~\citep{benou2025show}, ProtoQuant~\citep{janusz2026protoquant} and CFM spatial ones.

\begin{table}[!ht]
  \centering
  \definecolor{propyes}{HTML}{2E7D32}\definecolor{proppart}{HTML}{B7791F}\definecolor{propno}{HTML}{B4B4B4}%
  \providecommand{\cmark}{\textcolor{propyes}{\checkmark}}%
  \providecommand{\pmark}{\textcolor{proppart}{(\checkmark)}}%
  \providecommand{\xmark}{\textcolor{propno}{$\times$}}%
  \providecommand{\suprow}[1]{\textcolor{black!65}{#1}}%
  \caption{\textbf{Large-scale classification, all methods} (linear probe, top-1). \cmark~yes,
  \pmark~partially, \xmark~no, as in Table~\ref{tab:related-properties}; $\varnothing$: not applicable;
  \textemdash: not reported. \textbf{Best} and \underline{second best} among self-supervised methods; gray
  rows are unranked references. Vision-language results follow Fig.~6 of CFM~\citep{wittenmayer2026cfm};
  supervised ViT-B/16 results are from Table~4 of ProtoQuant~\citep{janusz2026protoquant}, using the CLS
  backbone baseline. $^{\dagger}$DCLIP uses description-based scoring, not a linear probe.}
  \label{tab:classification-full}
  \footnotesize
  \setlength{\tabcolsep}{3pt}
  \renewcommand{\arraystretch}{1.08}
  \begin{tabular}{@{}lcccc@{}}
    \toprule
    & \multicolumn{2}{c}{Concept representation} & \multicolumn{2}{c}{Top-1 accuracy} \\
    \cmidrule(lr){2-3}\cmidrule(l){4-5}
    Method & \shortstack{Spatial\\grounding} & \shortstack{Discrete\\bottleneck} & ImageNet-1k\,$\uparrow$ & Places\,$\uparrow$ \\
    \midrule
    \multicolumn{5}{@{}l}{\suprow{\textit{Fully supervised, ViT-B/16 (reference)}}} \\
    \suprow{Backbone} & \suprow{$\varnothing$} & \suprow{$\varnothing$} & \suprow{81.1} & \suprow{\textemdash} \\
    \suprow{InfoDisent~\citep{struski2024infodisent}} & \suprow{(\checkmark)} & \suprow{$\times$} & \suprow{79.2} & \suprow{\textemdash} \\
    \suprow{ProtoQuant~\citep{janusz2026protoquant}} & \suprow{(\checkmark)} & \suprow{(\checkmark)} & \suprow{80.1} & \suprow{\textemdash} \\
    \addlinespace
    \midrule
    \multicolumn{5}{@{}l}{\textit{Vision-language self-supervised, CLIP ViT-B/16}} \\
    \suprow{Backbone} & \suprow{$\varnothing$} & \suprow{$\varnothing$} & \suprow{80.2} & \suprow{55.1} \\
    LF-CBM~\citep{oikarinen2023label} & \xmark & \xmark & 75.4 & 50.6 \\
    LaBo~\citep{yang2023language} & \xmark & \xmark & 78.9 & \textemdash \\
    CDM~\citep{panousis2023sparse} & \xmark & \pmark & 79.3 & 52.6 \\
    DCLIP$^{\dagger}$~\citep{menon2023visual} & \xmark & \xmark & 68.0 & 40.3 \\
    DN-CBM~\citep{rao2024discover} & \xmark & \xmark & \underline{79.5} & 55.1 \\
    D-CBM~\citep{prasse2024dcbm} & \xmark & \xmark & 70.5 & 50.9 \\
    CF-CBM~\citep{panousis2024coarse} & \pmark & \pmark & 78.5 & \textemdash \\
    SALF-CBM~\citep{benou2025show} & \pmark & \xmark & 76.3 & 49.4 \\
    CFM~\citep{wittenmayer2026cfm} & \pmark & \xmark & 78.9 & \textbf{55.4} \\
    \addlinespace
    \midrule
    \multicolumn{5}{@{}l}{\textit{Vision-only self-supervised, DINOv2-B/14}} \\
    \suprow{Backbone} & \suprow{$\varnothing$} & \suprow{$\varnothing$} & \suprow{83.4} & \suprow{55.4} \\
    \rowcolor{oursrow} \textbf{\ours{}} (ours) & \cmark & \cmark & \textbf{83.2} & \underline{55.3} \\
    \bottomrule
  \end{tabular}
\end{table}

\myparagraph{Properties.} The two property columns follow Table~\ref{tab:related-properties}. \emph{Spatial grounding} is full when every location carries exactly one concept, partial for region-level or overlapping concepts, and absent for global concepts. A \emph{discrete bottleneck} is full when discrete indices alone define what the predictor reads, and partial when a discrete selection or codebook is combined with continuous scores, as in CDM, CF-CBM and ProtoQuant. \ours{} is the only method with both. The columns describe how each method is built, not how well it localizes.

\clearpage
\subsection{Full PartImageNet comparison}
\label{app:pin-full}
\label{app:pin-all}

We extend Table~\ref{tab:pin-concepts} by reporting every evaluated configuration under the same protocol, as differences to \ours{} on DINOv2-B/14. Figure~\ref{fig:pin-full-heatmap-supervised} covers the \textbf{supervised methods}: all part counts $K$ of PDiscoFormer, released and retrained, and PIP-Net and ProtoQuant with their classifier stage on every backbone we trained them on. Figure~\ref{fig:pin-full-heatmap-ssl} covers the \textbf{self-supervised methods}: every concept budget of CFM including its own unbudgeted setting, the sparse-autoencoder baselines across dictionary sizes, budgets and backbones, and the label-free first stage of PIP-Net and ProtoQuant. Figure~\ref{fig:pin-full-heatmap-msae} shows the Matryoshka SAE across dictionary sizes, concept budgets, loss weightings and its own backbone. \ours{} on the other backbones in Figure~\ref{fig:pin-full-heatmap-backbones}, with the backbone study of Appendix~\ref{abl:backbones}.

\input{\RepoFigureRoot/tables/heatmap_tables/fig_partimagenet_full_heatmap.tex}

\clearpage%
\subsection{Full CUB concept metrics}
\label{app:cub-full}
\label{app:finegrained}

Table~\ref{tab:cub-concepts} lists every concept metric we measure on CUB. Since CUB marks parts with keypoints, the clustering metrics are computed on the cells that hold a visible keypoint, and mIoU, FG-ARI and the grounding metrics are undefined (Appendix~\ref{app:metrics}). \ours{} leads the self-supervised methods on every column, by $3.5$ to $8.5$ points on the clustering metrics and by $17.2$ on PIP purity. The supervised PDiscoFormer, trained with class labels and evaluated at $518\times518$, scores higher on the clustering metrics for $K\geq8$. Its CUB checkpoints read $518\times518$ images, so we resample their $37\times37$ part maps to the $32\times32$ grid and scale the PIP purity window from $32$ to $74$ pixels (Appendix~\ref{app:metrics}). It also has a far lower keypoint error at every $K$ ($4.44$, $2.93$ and $2.87$ for $K{=}4$, $8$ and $16$, against $9.36$ for \ours{}), since its background channel concentrates its parts on the bird and it reads $518\times518$ images. Its PIP purity is $64.3$ at $K{=}4$ and $8$ and $40.4$ at $K{=}16$, against $54.5$ for \ours{} with $16$ concepts.

\begin{table*}[!hbp]%
  \centering
  \definecolor{deltapos}{HTML}{1F5FBF}\definecolor{deltaneg}{HTML}{C62828}%
  \providecommand{\dpos}[1]{\textsubscript{\textcolor{deltapos}{\textbf{#1}}}}%
  \providecommand{\dneg}[1]{\textsubscript{\textcolor{deltaneg}{\textbf{#1}}}}%
  \providecommand{\suprow}[1]{\textcolor{black!65}{#1}}%
  \caption{\textbf{All concept metrics on CUB-200-2011} (test split, frozen DINOv2-B/14; CFM: CLIP-DINOiser;
  MSAE\textsubscript{CLIP-L/14}: MSAE on the model it was published on;
  PDiscoFormer: the released checkpoints, at $518\times518$); $\times100$ except Kp NME. CUB is annotated with
  keypoints, not part masks, so mIoU, FG-ARI and the grounding triple of Table~\ref{tab:pin-concepts} are
  undefined. Marks, subscripts and gray rows as in Table~\ref{tab:fine-grained}; $\varnothing$: undefined for
  that method, \textemdash: not measured.}
  \label{tab:cub-concepts}
  \footnotesize
  \setlength{\tabcolsep}{4pt}
  \renewcommand{\arraystretch}{1.08}
  \begin{tabular}{@{}llccccccc@{}}
    \toprule
    & & \multicolumn{2}{c}{Downstream} & \multicolumn{5}{c}{Concept quality} \\
    \cmidrule(lr){3-4}\cmidrule(l){5-9}
    Method & $K/N$ & LP\,$\uparrow$ & kNN\,$\uparrow$ & Pur$_{\mathrm{fg}}$\,$\uparrow$ & NMI$_m$\,$\uparrow$ & ARI$_m$\,$\uparrow$ & PIP pur.\,$\uparrow$ & Kp NME\,$\downarrow$ \\
    \midrule
    \multicolumn{9}{@{}l}{\suprow{\textit{Supervised (reference)}}} \\
    \suprow{PDiscoFormer} & \suprow{4/4} & \suprow{\textemdash} & \suprow{\textemdash} & \suprow{28.6} & \suprow{55.1} & \suprow{27.3} & \suprow{64.3} & \suprow{4.44} \\
    \suprow{PDiscoFormer} & \suprow{8/8} & \suprow{\textemdash} & \suprow{\textemdash} & \suprow{51.9} & \suprow{67.2} & \suprow{42.5} & \suprow{64.3} & \suprow{2.93} \\
    \suprow{PDiscoFormer} & \suprow{16/16} & \suprow{86.8} & \suprow{74.8} & \suprow{53.6} & \suprow{65.6} & \suprow{46.0} & \suprow{40.4} & \suprow{2.87} \\
    \suprow{PCA} & \suprow{16/3200} & \suprow{$\varnothing$} & \suprow{$\varnothing$} & \suprow{14.3} & \suprow{3.5} & \suprow{0.4} & \suprow{\textemdash} & \suprow{\textemdash} \\
    \suprow{PIP-Net\textsubscript{stage 2}} & \suprow{16/768} & \suprow{88.9} & \suprow{88.9} & \suprow{42.0} & \suprow{37.0} & \suprow{21.3} & \suprow{47.0} & \suprow{10.16} \\
    \suprow{ProtoQuant\textsubscript{stage 2}} & \suprow{16/4000} & \suprow{84.9} & \suprow{83.0} & \suprow{41.0} & \suprow{33.3} & \suprow{21.2} & \suprow{25.2} & \suprow{12.16} \\
    \addlinespace
    \midrule
    \multicolumn{9}{@{}l}{\suprow{\textit{Frozen backbone, no concept bottleneck (reference)}}} \\
    \suprow{CLS token} & \suprow{\textemdash} & \suprow{90.4} & \suprow{86.7} & \suprow{\textemdash} & \suprow{\textemdash} & \suprow{\textemdash} & \suprow{\textemdash} & \suprow{\textemdash} \\
    \addlinespace
    \midrule
    \multicolumn{9}{@{}l}{\textit{Self-supervised}} \\
    CFM & 16/8192 & 54.8 & 42.8 & 12.5 & 4.8 & 0.6 & 16.1 & 12.84 \\
    TopK SAE & 16/2048 & 36.9 & 24.3 & 21.2 & 17.0 & 7.4 & 8.7 & 10.92 \\
    PatchSAE & 16/49152 & 62.0 & 33.4 & 18.3 & 5.0 & 2.3 & 6.9 & 19.12 \\
    MSAE & 16/6144 & 73.5 & 42.2 & 13.8 & 3.2 & 1.1 & 15.1 & 12.21 \\
    MSAE\textsubscript{CLIP-L/14} & 16/8192 & 71.5 & 36.5 & 18.3 & 15.2 & 4.4 & 8.8 & 11.44 \\
    PIP-Net\textsubscript{stage 1} & 16/768 & 77.8 & \underline{68.4} & 39.7 & 32.7 & 16.4 & \underline{37.3} & \underline{9.87} \\
    ProtoQuant\textsubscript{stage 1} & 16/4000 & \underline{80.1} & 67.7 & \underline{44.0} & \underline{36.4} & \underline{24.3} & 24.6 & 11.59 \\
    \addlinespace
    \midrule
    \rowcolor{oursrow} \textbf{\ours{}} (ours) & 16/1024 & \textbf{89.4}\dpos{+9.3} & \textbf{85.5}\dpos{+17.1} & \textbf{47.5}\dpos{+3.5} & \textbf{44.9}\dpos{+8.5} & \textbf{29.6}\dpos{+5.3} & \textbf{54.5}\dpos{+17.2} & \textbf{9.36}\dpos{-0.51} \\
    \bottomrule
  \end{tabular}
\end{table*}

\subsection{PDiscoFormer on Cars and Dogs}
\label{app:pdisco-repro}

PDiscoFormer releases no checkpoint for Stanford Cars or Stanford Dogs, so Table~\ref{tab:fine-grained} leaves those cells empty. Table~\ref{tab:pdisco-repro} gives our retraining with the authors' code and recipe (Appendix~\ref{app:baselines}), at $224\times224$ with $K{=}16$ parts, on the same training splits as \ours{}. As in Table~\ref{tab:pin-concepts}, the probes read the mean of its modulated part embeddings. Trained with class labels, PDiscoFormer matches \ours{} on the linear probe on Cars ($91.1$ vs.\ $91.0$) and trails it by $2.3$ to $4.9$~p.p.\ on the other three columns. Its recipe keeps the snapshot with the best test accuracy, which is that of epoch $27$ of $28$ on both datasets, so this selection hardly touches the test split.

\begin{table}[!htbp]
  \centering
  \providecommand{\suprow}[1]{\textcolor{black!65}{#1}}%
  \caption{\textbf{PDiscoFormer reproduced on Cars and Dogs} (test, $\times100$): the authors' code retrained at $224\times224$ with $K{=}16$. The other rows are from Table~\ref{tab:fine-grained}.}
  \label{tab:pdisco-repro}
  \small
  \setlength{\tabcolsep}{5pt}
  \begin{tabular}{@{}llcccc@{}}
    \toprule
    & & \multicolumn{2}{c}{Stanford Cars} & \multicolumn{2}{c}{Stanford Dogs} \\
    \cmidrule(lr){3-4}\cmidrule(l){5-6}
    Method & $K/N$ & LP\,$\uparrow$ & kNN\,$\uparrow$ & LP\,$\uparrow$ & kNN\,$\uparrow$ \\
    \midrule
    \suprow{PDiscoFormer, reproduced} & \suprow{16/16} & \suprow{91.1} & \suprow{79.9} & \suprow{85.5} & \suprow{80.9} \\
    \suprow{CLS token} & \suprow{\textemdash} & \suprow{93.2} & \suprow{85.6} & \suprow{88.6} & \suprow{86.8} \\
    \rowcolor{oursrow} \textbf{\ours{}} (ours) & 16/1024 & 91.0 & 83.4 & 87.8 & 85.8 \\
    \bottomrule
  \end{tabular}
\end{table}

\subsection{ProtoQuant and PIP-Net: frozen vs. fine-tuned backbone}
\label{app:backbone-freeze}

 Two of our baselines fine-tune the backbone in their own protocols. ProtoQuant trains it for classification on the target dataset and then freezes it, so the per-backbone accuracies its Table~3 reports are those of a fine-tuned network, and PIP-Net's recipe trains its backbone end to end. We report both on the frozen DINOv2-B/14 that every other method reads (Appendix~\ref{app:baselines}), and we also ran their protocols as published. Table~\ref{tab:backbone-freeze} gives the pair, for completeness and as a control on that choice.

Fine-tuning improves classification accuracy and hurts part quality. ProtoQuant's stage-1 linear probe on PartImageNet rises from $85.2$ to $89.1$ and its kNN from $83.9$ to $88.7$, while its foreground purity falls from $82.1$ to $61.2$, its part mIoU from $33.7$ to $9.3$ and its part-to-part R@1 from $79.2$ to $30.7$. On CUB its purity falls from $44.0$ to $16.3$ and its keypoint error rises from $11.59$ to $14.47$, against the $15.50$ of predicting the mean keypoint. PIP-Net loses the same part columns and, unlike ProtoQuant, loses accuracy with them: its stage-1 probe drops from $77.8$ to $57.6$ and its kNN from $70.6$ to $31.8$. What the backbone is fine-tuned \emph{on} accounts for the difference. ProtoQuant's first stage trains it with class labels, so its features are class-discriminative before any codebook is learned, while PIP-Net trains it through its own label-free objective, which supplies no such signal and degrades the pretrained features in this configuration.

One column improves: PIP-Net's keypoint error on CUB, from $9.91$ to $9.07$. Kp NME regresses keypoints from part centroids, so it rewards parts that lie on the bird even when those parts are less pure, and it is the one column in the table that does not read a part as a region. Further arms are left out of the table because they move more than one thing at a time. Using PIP-Net's official protocol, which unfreezes the last two blocks after a frozen warm-up and scores a bare softmax over the backbone's channels, yields lower part scores on PartImageNet: foreground purity $20.8$ and mIoU $12.5$ at stage 1 on DINOv2-B/14, against $50.7$ and $21.8$ frozen, and $15.8$ and $7.9$ on the ConvNeXt and $27.0$ and $17.3$ on the ResNet from its own paper. Of the fine-tuned ProtoQuant arms we ran with three seeds, on their DeiT-S, accuracy varies by at most $0.6$, which is why the table carries one seed per cell.

Thus, the frozen rows of the main tables are not a weakened form of either method. They are the stronger arm on the columns the paper measures, and they keep the comparison to one variable, since a fine-tuned arm sees the target dataset's labels through its backbone as well as through its classifier. The fine-tuned runs are usable as a control because they reproduce their authors' reported accuracy: ProtoQuant's fine-tuned DeiT-S reaches a top-1 of $87.0$ on bounding-box-cropped CUB, against the $86.9$ of their Table~3.

\begin{table*}[!htbp]
  \centering
  \providecommand{\suprow}[1]{\textcolor{black!65}{#1}}%
  \caption{\textbf{A frozen against a fine-tuned backbone}, for the two baselines whose own protocols
  fine-tune it (DINOv2-B/14; $\times100$ except Kp NME; \emph{stage 2} rows are gray, as they use class
  labels). The main paper reports the frozen arms, so that every method reads the same features; these rows
  reproduce them. Fine-tuning raises classification accuracy on every dataset and lowers part quality
  throughout, with one exception: PIP-Net's keypoint error on CUB, which improves.
  PIP-Net's fine-tuned arms exist only at the native $16\times16$ grid, so its pair is the second and third
  rows of its block and the first row is the AnyUp arm of Tables~\ref{tab:pin-concepts}
  and~\ref{tab:fine-grained}. CUB carries Pur$_{\mathrm{fg}}$ and Kp NME rather than PIP purity, whose
  window rule changed after the fine-tuned arms were scored. \textemdash: not measured.}
  \label{tab:backbone-freeze}
  \footnotesize
  \setlength{\tabcolsep}{4pt}
  \renewcommand{\arraystretch}{1.08}
  \resizebox{\textwidth}{!}{%
  \begin{tabular}{@{}lllccccccccccc@{}}
    \toprule
    & & & \multicolumn{5}{c}{PartImageNet} & \multicolumn{3}{c}{CUB} & Cars & Dogs & Flowers \\
    \cmidrule(lr){4-8}\cmidrule(lr){9-11}\cmidrule(lr){12-12}\cmidrule(lr){13-13}\cmidrule(l){14-14}
    Method & Backbone & Stage & LP\,$\uparrow$ & kNN\,$\uparrow$ & Pur$_{\mathrm{fg}}$\,$\uparrow$ & mIoU\,$\uparrow$ & p2p R@1\,$\uparrow$ & LP\,$\uparrow$ & Pur$_{\mathrm{fg}}$\,$\uparrow$ & Kp NME\,$\downarrow$ & LP\,$\uparrow$ & LP\,$\uparrow$ & LP\,$\uparrow$ \\
    \midrule
    PIP-Net & frozen, AnyUp & 1 & 76.9 & 71.5 & 68.5 & 29.0 & 66.6 & 77.8 & 39.7 & 9.87 & 82.3 & 77.8 & 99.1 \\
    \suprow{PIP-Net} & \suprow{frozen, AnyUp} & \suprow{2} & \suprow{87.5} & \suprow{86.8} & \suprow{59.3} & \suprow{24.9} & \suprow{65.4} & \suprow{88.9} & \suprow{42.0} & \suprow{10.16} & \suprow{91.4} & \suprow{87.1} & \suprow{99.6} \\
    PIP-Net & frozen & 1 & 77.8 & 70.6 & 50.7 & 21.8 & 61.3 & 79.0 & 38.8 & 9.91 & 85.3 & 76.4 & 99.0 \\
    \suprow{PIP-Net} & \suprow{frozen} & \suprow{2} & \suprow{87.2} & \suprow{87.8} & \suprow{37.1} & \suprow{20.5} & \suprow{51.5} & \suprow{88.8} & \suprow{39.4} & \suprow{10.23} & \suprow{91.5} & \suprow{86.8} & \suprow{99.6} \\
    PIP-Net & fine-tuned & 1 & 57.6 & 31.8 & 26.4 & 17.2 & 31.7 & 67.6 & 33.5 & 9.07 & 70.8 & 51.8 & 91.4 \\
    \suprow{PIP-Net} & \suprow{fine-tuned} & \suprow{2} & \suprow{88.2} & \suprow{85.6} & \suprow{22.1} & \suprow{15.3} & \suprow{29.1} & \suprow{90.0} & \suprow{37.2} & \suprow{8.42} & \suprow{93.2} & \suprow{86.0} & \suprow{99.5} \\
    \addlinespace
    ProtoQuant & frozen & 1 & 85.2 & 83.9 & 82.1 & 33.7 & 79.2 & 80.1 & 44.0 & 11.59 & 87.0 & 82.2 & 99.7 \\
    \suprow{ProtoQuant} & \suprow{frozen} & \suprow{2} & \suprow{87.8} & \suprow{89.5} & \suprow{71.7} & \suprow{26.5} & \suprow{62.6} & \suprow{84.9} & \suprow{41.0} & \suprow{12.16} & \suprow{91.3} & \suprow{86.0} & \suprow{99.6} \\
    ProtoQuant & fine-tuned & 1 & 89.1 & 88.7 & 61.2 & 9.3 & 30.7 & 89.1 & 16.3 & 14.47 & 93.8 & 87.2 & 99.5 \\
    \suprow{ProtoQuant} & \suprow{fine-tuned} & \suprow{2} & \suprow{88.1} & \suprow{89.0} & \suprow{62.8} & \suprow{13.3} & \suprow{35.2} & \suprow{89.4} & \suprow{21.2} & \suprow{14.85} & \suprow{94.1} & \suprow{86.5} & \suprow{99.5} \\
    \bottomrule
  \end{tabular}}
\end{table*}

\subsection{Retrieval results}
\label{app:retrieval-full}

Table~\ref{tab:retrieval-full} is Table~\ref{tab:part-retrieval} with CUB and the part\,$\to$\,image setting added, plus two rows for: PCA (per-image and class-wise), and \ours{} described by something other than its concept identities. The attribute head quantizes a \emph{residual}, the normalized region feature minus its normalized prototype, and each concept owns a private codebook, so an attribute vector carries no concept identity by construction, and its codes are only comparable between occurrences of the same concept. The three descriptor rows show what follows: the codes alone retrieve a same-part neighbor about as often as the concept ids do (R@1 $78.9$ against $79.7$ on PartImageNet) but rank the rest of the list far worse (mAP $16.3$ against $30.5$); concatenating the two changes nothing; and using the attribute to order candidates that already share a concept raises mAP to $57.3$ on PartImageNet and from $9.3$ to $28.1$ on CUB. Appearance is orthogonal to part identity, and only useful once identity has selected the candidates. No baseline has an attribute, so these rows are ours alone and unranked.

Table~\ref{tab:retrieval-full} also holds the rows that Table~\ref{tab:part-retrieval} leaves out. The category-only oracle and uniform chance are two reference rankers that use no concepts (Appendix~\ref{app:metrics}). The oracle uses ground-truth classes for accuracy and image-to-image retrieval, solving those columns by construction. For part retrieval, it uses only part-label co-occurrence groups and reaches an R@1 of $43.9$, against $79.7$ for \ours{}. The components of per-image PCA (Appendix~\ref{app:baselines}) differ from image to image, so part-to-part retrieval with them is close to chance (R@1 $5.3$, against $4.6$). Class-wise PCA, which selects its components with the class label and is therefore a supervised reference, shows why R@1 must be read beside mAP: ranking by category, it reaches an R@1 of $52.4$ on its own query set at an mAP of only $8.9$.

PDiscoFormer is scored at its native feature resolution, whose $16\times16$ grid gives a query set of $3{,}111$ regions and an oracle R@1 of $46.2$, against $3{,}524$ regions and $43.9$ on the $32\times32$ grid (Appendix~\ref{app:metrics}). Resampling its part maps to our $32\times32$ grid, as we do for its CUB models, puts it on our query set and changes its scores by at most $3.5$: part-to-part R@1 is $45.8$, $63.5$ and $71.8$ for $K{=}16$, $25$ and $50$ (natively $49.3$, $63.6$ and $71.8$), and mAP is $32.1$, $40.8$ and $43.9$ (natively $34.2$, $41.9$ and $45.9$). On either grid, PDiscoFormer trails \ours{} on R@1 ($79.7$) and exceeds it on mAP ($30.5$).

On CUB, \ours{} leads part-to-part R@1 ($36.8$) and R@5 ($73.7$), against an R@1 of $22.7$ for the oracle and $7.1$ for chance, and trails the TopK SAE by $0.2$ on mAP. Part-to-image retrieval is at chance on CUB, since nearly every image contains nearly every keypoint, so it is omitted ($\varnothing$). Appendix~\ref{app:metrics} defines the regions, the oracle and the query sets.

MSAE is trained under two loss weightings. The MSAE rows of Table~\ref{tab:retrieval-full} use the reverse-weighted dictionary, the configuration selected on PartImageNet (Table~\ref{tab:pin-concepts}), which is the only one scored on this benchmark. Table~\ref{tab:fine-grained} and Table~\ref{tab:cub-concepts} instead report the uniform-weighted dictionary selected jointly across the four fine-grained datasets, whose CUB linear probe and $k$-NN read $73.5$ and $42.2$ against the $72.5$ and $41.7$ reported there.

\begin{table*}[t]
  \centering
  \definecolor{deltapos}{HTML}{1F5FBF}\definecolor{deltaneg}{HTML}{C62828}%
  \providecommand{\dpos}[1]{\textsubscript{\textcolor{deltapos}{\textbf{#1}}}}%
  \providecommand{\dneg}[1]{\textsubscript{\textcolor{deltaneg}{\textbf{#1}}}}%
  \providecommand{\suprow}[1]{\textcolor{black!65}{#1}}%
  \caption{\textbf{Retrieval on PartImageNet and CUB} ($\times100$). Among self-supervised methods, \ours{} has the best part-to-part R@1 on both datasets. Conventions follow Table~\ref{tab:pin-concepts}. A double dagger marks a different grid and oracle (Section~\ref{app:retrieval-full}).}
  \label{tab:retrieval-full}
  \footnotesize
  \setlength{\tabcolsep}{3pt}
  \renewcommand{\arraystretch}{1.08}
  \resizebox{\textwidth}{!}{%
  \begin{tabular}{@{}llccccccccccc@{}}
    \toprule
    & & \multicolumn{2}{c}{Accuracy} & \multicolumn{3}{c}{Image $\to$ image} & \multicolumn{3}{c}{Part $\to$ image} & \multicolumn{3}{c}{Part $\to$ part} \\
    \cmidrule(lr){3-4}\cmidrule(lr){5-7}\cmidrule(lr){8-10}\cmidrule(lr){11-13}
    Method & $K/N$ & LP\,$\uparrow$ & kNN\,$\uparrow$ & R@1\,$\uparrow$ & R@5\,$\uparrow$ & mAP\,$\uparrow$ & R@1\,$\uparrow$ & R@5\,$\uparrow$ & mAP\,$\uparrow$ & R@1\,$\uparrow$ & R@5\,$\uparrow$ & mAP\,$\uparrow$ \\
    \midrule
    \multicolumn{13}{@{}l}{\textbf{PartImageNet}} \\
    \addlinespace[1pt]
    \multicolumn{13}{@{}l}{\suprow{\textit{Supervised (reference)}}} \\
    \suprow{PDiscoFormer$^{\ddagger}$} & \suprow{16/16} & \suprow{89.6} & \suprow{88.5} & \suprow{78.8} & \suprow{92.9} & \suprow{70.2} & \suprow{38.8} & \suprow{63.5} & \suprow{36.3} & \suprow{49.3} & \suprow{78.3} & \suprow{34.2} \\
    \suprow{PDiscoFormer$^{\ddagger}$} & \suprow{25/25} & \suprow{89.5} & \suprow{87.5} & \suprow{78.4} & \suprow{93.4} & \suprow{69.1} & \suprow{52.0} & \suprow{74.6} & \suprow{42.5} & \suprow{63.6} & \suprow{86.2} & \suprow{41.9} \\
    \suprow{PDiscoFormer$^{\ddagger}$} & \suprow{50/50} & \suprow{89.8} & \suprow{88.1} & \suprow{78.4} & \suprow{93.4} & \suprow{69.3} & \suprow{61.9} & \suprow{80.4} & \suprow{51.4} & \suprow{71.8} & \suprow{89.0} & \suprow{45.9} \\
    \suprow{PCA, class-wise$^{\ddagger}$} & \suprow{16/2528} & \suprow{$\varnothing$} & \suprow{$\varnothing$} & \suprow{$\varnothing$} & \suprow{$\varnothing$} & \suprow{$\varnothing$} & \suprow{83.0} & \suprow{98.4} & \suprow{20.1} & \suprow{52.4} & \suprow{88.1} & \suprow{8.9} \\
    \suprow{PIP-Net\textsubscript{stage 2}} & \suprow{16/768} & \suprow{87.5} & \suprow{86.8} & \suprow{77.6} & \suprow{91.9} & \suprow{64.2} & \suprow{77.5} & \suprow{93.8} & \suprow{45.9} & \suprow{65.4} & \suprow{85.2} & \suprow{21.1} \\
    \suprow{ProtoQuant\textsubscript{stage 2}} & \suprow{16/2048} & \suprow{87.8} & \suprow{89.5} & \suprow{82.3} & \suprow{93.8} & \suprow{76.6} & \suprow{86.2} & \suprow{98.5} & \suprow{33.0} & \suprow{62.6} & \suprow{89.6} & \suprow{15.5} \\
    \addlinespace
    \midrule
    \multicolumn{13}{@{}l}{\suprow{\textit{Frozen backbone, no concept bottleneck (reference)}}} \\
    \suprow{CLS token} & \suprow{\textemdash} & \suprow{89.3} & \suprow{88.9} & \suprow{81.0} & \suprow{92.9} & \suprow{75.1} & \suprow{\textemdash} & \suprow{\textemdash} & \suprow{\textemdash} & \suprow{\textemdash} & \suprow{\textemdash} & \suprow{\textemdash} \\
    \suprow{Patch tokens, mean-pooled} & \suprow{\textemdash} & \suprow{85.7} & \suprow{79.3} & \suprow{57.5} & \suprow{82.2} & \suprow{44.7} & \suprow{\textemdash} & \suprow{\textemdash} & \suprow{\textemdash} & \suprow{\textemdash} & \suprow{\textemdash} & \suprow{\textemdash} \\
    \multicolumn{13}{@{}l}{\suprow{\textit{Ground-truth controls}}} \\
    \suprow{category-only oracle} & \suprow{\textemdash} & \suprow{100.0} & \suprow{100.0} & \suprow{100.0} & \suprow{100.0} & \suprow{100.0} & \suprow{82.2} & \suprow{91.5} & \suprow{81.5} & \suprow{43.9} & \suprow{89.0} & \suprow{44.7} \\
    \suprow{uniform chance} & \suprow{\textemdash} & \suprow{\textemdash} & \suprow{\textemdash} & \suprow{0.6} & \suprow{2.9} & \suprow{0.6} & \suprow{13.7} & \suprow{47.7} & \suprow{14.1} & \suprow{4.6} & \suprow{20.1} & \suprow{4.9} \\
    \addlinespace
    \midrule
    \multicolumn{13}{@{}l}{\textit{Self-supervised}} \\
    PCA, per-image & 16/16 & $\varnothing$ & $\varnothing$ & $\varnothing$ & $\varnothing$ & $\varnothing$ & 11.0 & 39.0 & 14.5 & 5.3 & 23.6 & 5.2 \\
    CFM$^{\ddagger}$ & 16/8192 & 72.6 & 69.2 & 47.0 & 75.4 & 34.6 & 63.4 & 74.3 & 40.8 & 31.6 & 68.8 & 16.0 \\
    TopK SAE & 16/1024 & 78.1 & 73.0 & 52.6 & 79.2 & 42.7 & 84.7 & 97.4 & 34.9 & 56.2 & 86.5 & 17.2 \\
    TopK SAE & 16/2048 & 80.5 & 73.5 & 53.8 & 77.5 & 43.3 & 85.6 & \underline{98.0} & 35.7 & 58.3 & 86.9 & 18.1 \\
    PatchSAE & 16/1024 & 82.9 & 76.5 & 56.8 & 80.8 & 44.6 & 83.3 & 97.4 & 29.3 & 41.9 & 83.6 & 11.8 \\
    PatchSAE & 16/49152 & 84.1 & 77.2 & 56.8 & 81.1 & 44.4 & 85.0 & 95.5 & 20.6 & 48.6 & 82.9 & 8.0 \\
    MSAE & 16/1024 & 85.2 & 79.9 & 58.9 & 81.7 & 46.1 & 82.9 & 95.8 & 30.9 & 41.4 & 82.2 & 12.5 \\
    MSAE & 16/8192 & \underline{85.6} & 79.5 & 58.3 & 82.3 & 45.4 & 79.8 & 93.4 & 30.3 & 41.2 & 79.9 & 12.6 \\
    PIP-Net\textsubscript{stage 1} & 16/768 & 76.9 & 71.5 & 49.1 & 78.6 & 36.3 & 77.9 & 94.6 & \underline{47.6} & 66.6 & 87.7 & \underline{27.0} \\
    ProtoQuant\textsubscript{stage 1} & 16/2048 & 85.2 & \underline{83.9} & \underline{72.5} & \underline{90.4} & \underline{61.8} & \textbf{89.8} & \textbf{98.8} & 40.3 & \underline{79.2} & \textbf{92.8} & 22.7 \\
    \addlinespace
    \rowcolor{oursrow} \textbf{\ours{}} (ours) & 16/1024 & \textbf{89.2}\dpos{+3.6} & \textbf{88.7}\dpos{+4.8} & \textbf{80.6}\dpos{+8.1} & \textbf{92.9}\dpos{+2.5} & \textbf{76.1}\dpos{+14.3} & \underline{86.9}\dneg{-2.9} & 97.8\dneg{-1.0} & \textbf{48.5}\dpos{+0.9} & \textbf{79.7}\dpos{+0.5} & \underline{92.4}\dneg{-0.4} & \textbf{30.5}\dpos{+3.5} \\
    \multicolumn{13}{@{}l}{\suprow{\textit{\ours{}, the same model described differently (ours only, unranked)}}} \\
    \suprow{\quad attribute codes only} & \suprow{16/1024} & \suprow{89.2} & \suprow{88.7} & \suprow{80.6} & \suprow{92.9} & \suprow{76.1} & \suprow{89.8} & \suprow{98.1} & \suprow{32.1} & \suprow{78.9} & \suprow{93.8} & \suprow{16.3} \\
    \suprow{\quad concept id $+$ attribute} & \suprow{16/1024} & \suprow{89.2} & \suprow{88.7} & \suprow{80.6} & \suprow{92.9} & \suprow{76.1} & \suprow{86.9} & \suprow{97.8} & \suprow{48.5} & \suprow{79.9} & \suprow{92.4} & \suprow{30.5} \\
    \suprow{\quad concept id $+$ same-concept appearance} & \suprow{16/1024} & \suprow{89.2} & \suprow{88.7} & \suprow{80.6} & \suprow{92.9} & \suprow{76.1} & \suprow{74.8} & \suprow{91.9} & \suprow{37.7} & \suprow{82.4} & \suprow{93.2} & \suprow{57.3} \\
    \addlinespace
    \midrule
    \multicolumn{13}{@{}l}{\textbf{CUB-200-2011}} \\
    \addlinespace[1pt]
    \multicolumn{13}{@{}l}{\suprow{\textit{Supervised (reference)}}} \\
    \suprow{PDiscoFormer} & \suprow{16/16} & \suprow{86.8} & \suprow{74.8} & \suprow{69.8} & \suprow{89.5} & \suprow{38.2} & \suprow{$\varnothing$} & \suprow{$\varnothing$} & \suprow{$\varnothing$} & \suprow{45.8} & \suprow{84.6} & \suprow{38.6} \\
    \suprow{PCA, class-wise} & \suprow{16/3200} & \suprow{$\varnothing$} & \suprow{$\varnothing$} & \suprow{$\varnothing$} & \suprow{$\varnothing$} & \suprow{$\varnothing$} & \suprow{$\varnothing$} & \suprow{$\varnothing$} & \suprow{$\varnothing$} & \suprow{9.0} & \suprow{34.0} & \suprow{7.1} \\
    \suprow{PIP-Net\textsubscript{stage 2}} & \suprow{16/768} & \suprow{88.9} & \suprow{88.9} & \suprow{85.6} & \suprow{94.9} & \suprow{75.0} & \suprow{$\varnothing$} & \suprow{$\varnothing$} & \suprow{$\varnothing$} & \suprow{32.3} & \suprow{69.8} & \suprow{8.5} \\
    \suprow{ProtoQuant\textsubscript{stage 2}} & \suprow{16/4000} & \suprow{84.9} & \suprow{83.0} & \suprow{76.8} & \suprow{93.5} & \suprow{52.5} & \suprow{$\varnothing$} & \suprow{$\varnothing$} & \suprow{$\varnothing$} & \suprow{30.1} & \suprow{65.9} & \suprow{8.4} \\
    \addlinespace
    \midrule
    \multicolumn{13}{@{}l}{\suprow{\textit{Frozen backbone, no concept bottleneck (reference)}}} \\
    \suprow{CLS token} & \suprow{\textemdash} & \suprow{90.4} & \suprow{86.7} & \suprow{85.1} & \suprow{95.4} & \suprow{65.9} & \suprow{$\varnothing$} & \suprow{$\varnothing$} & \suprow{$\varnothing$} & \suprow{\textemdash} & \suprow{\textemdash} & \suprow{\textemdash} \\
    \multicolumn{13}{@{}l}{\suprow{\textit{Ground-truth controls}}} \\
    \suprow{category-only oracle} & \suprow{\textemdash} & \suprow{100.0} & \suprow{100.0} & \suprow{100.0} & \suprow{100.0} & \suprow{100.0} & \suprow{$\varnothing$} & \suprow{$\varnothing$} & \suprow{$\varnothing$} & \suprow{22.7} & \suprow{60.2} & \suprow{22.9} \\
    \suprow{uniform chance} & \suprow{\textemdash} & \suprow{\textemdash} & \suprow{\textemdash} & \suprow{0.5} & \suprow{2.4} & \suprow{0.5} & \suprow{$\varnothing$} & \suprow{$\varnothing$} & \suprow{$\varnothing$} & \suprow{7.1} & \suprow{30.2} & \suprow{7.1} \\
    \addlinespace
    \midrule
    \multicolumn{13}{@{}l}{\textit{Self-supervised}} \\
    PCA, per-image & 16/16 & $\varnothing$ & $\varnothing$ & $\varnothing$ & $\varnothing$ & $\varnothing$ & $\varnothing$ & $\varnothing$ & $\varnothing$ & 7.6 & 32.6 & 7.2 \\
    CFM & 16/8192 & 54.8 & 42.8 & 35.9 & 63.3 & 11.9 & $\varnothing$ & $\varnothing$ & $\varnothing$ & 8.9 & 35.1 & 7.4 \\
    TopK SAE & 16/1024 & 36.6 & 24.5 & 22.0 & 44.0 & 10.0 & $\varnothing$ & $\varnothing$ & $\varnothing$ & 15.5 & 42.9 & \textbf{9.5} \\
    TopK SAE & 16/2048 & 36.9 & 24.3 & 21.9 & 43.4 & 10.0 & $\varnothing$ & $\varnothing$ & $\varnothing$ & 15.9 & 47.2 & \underline{9.4} \\
    PatchSAE & 16/1024 & 56.6 & 33.5 & 30.1 & 55.1 & 12.6 & $\varnothing$ & $\varnothing$ & $\varnothing$ & 9.3 & 34.3 & 7.2 \\
    PatchSAE & 16/49152 & 62.0 & 33.4 & 30.2 & 54.1 & 11.4 & $\varnothing$ & $\varnothing$ & $\varnothing$ & 10.9 & 37.2 & 7.2 \\
    MSAE\textsubscript{RW} & 16/1024 & 71.0 & 41.3 & 35.9 & 61.9 & 14.4 & $\varnothing$ & $\varnothing$ & $\varnothing$ & 10.5 & 37.1 & 7.7 \\
    MSAE\textsubscript{RW} & 16/6144 & 72.5 & 41.7 & 36.3 & 62.2 & 14.4 & $\varnothing$ & $\varnothing$ & $\varnothing$ & 9.5 & 34.1 & 7.5 \\
    PIP-Net\textsubscript{stage 1} & 16/768 & 77.8 & 68.4 & 60.7 & 86.4 & 35.2 & $\varnothing$ & $\varnothing$ & $\varnothing$ & 29.8 & 64.3 & \underline{9.4} \\
    ProtoQuant\textsubscript{stage 1} & 16/4000 & 80.1 & 67.7 & 59.4 & 84.7 & 31.5 & $\varnothing$ & $\varnothing$ & $\varnothing$ & 30.9 & 67.5 & 8.2 \\
    \addlinespace
    \rowcolor{oursrow} \textbf{\ours{}} (ours) & 16/1024 & \textbf{89.4}\dpos{+9.3} & \textbf{85.5}\dpos{+17.1} & \textbf{84.5}\dpos{+23.8} & \textbf{95.1}\dpos{+8.7} & \textbf{67.8}\dpos{+32.6} & $\varnothing$ & $\varnothing$ & $\varnothing$ & \textbf{36.8}\dpos{+5.9} & \textbf{73.7}\dpos{+6.2} & 9.3\dneg{-0.2} \\
    \multicolumn{13}{@{}l}{\suprow{\textit{\ours{}, the same model described differently (ours only, unranked)}}} \\
    \suprow{\quad attribute codes only} & \suprow{16/1024} & \suprow{89.4} & \suprow{85.5} & \suprow{84.5} & \suprow{95.1} & \suprow{67.8} & \suprow{$\varnothing$} & \suprow{$\varnothing$} & \suprow{$\varnothing$} & \suprow{37.6} & \suprow{71.5} & \suprow{7.9} \\
    \suprow{\quad concept id $+$ attribute} & \suprow{16/1024} & \suprow{89.4} & \suprow{85.5} & \suprow{84.5} & \suprow{95.1} & \suprow{67.8} & \suprow{$\varnothing$} & \suprow{$\varnothing$} & \suprow{$\varnothing$} & \suprow{38.6} & \suprow{74.0} & \suprow{9.3} \\
    \suprow{\quad concept id $+$ same-concept appearance} & \suprow{16/1024} & \suprow{89.4} & \suprow{85.5} & \suprow{84.5} & \suprow{95.1} & \suprow{67.8} & \suprow{$\varnothing$} & \suprow{$\varnothing$} & \suprow{$\varnothing$} & \suprow{39.6} & \suprow{76.5} & \suprow{28.1} \\
    \bottomrule
  \end{tabular}}
\end{table*}

\clearpage%
\subsection{What the quantized attributes keep}
\label{app:qualitative-quant}

For a selected concept $i_k$ in an image, the attribute encoder sees the residual $r_{i_k}=\nu(\bar z_{i_k})-\nu(w_{i_k})$ between the normalized region mean $\bar z_{i_k}$ and the normalized concept prototype $w_{i_k}$ (Section~\ref{subsec:attribute-quant}), so $\nu(w_{i_k})+r_{i_k}$ is the normalized region mean. Stage~2 projects the residual to $y_{i_k}=W_{\mathrm{PQ}}\,r_{i_k}$ and encodes it as $G{=}16$ indices $j_{i_k,1:G}$ into the concept-specific codebooks $\mathcal B_{i_k}$, with $M{=}64$ codewords per group (Appendix~\ref{app:quantization}). The attribute $q_{i_k}$ that the decoder receives is learned for decoding and is not trained to reconstruct $y_{i_k}$.

To see what the code keeps, we therefore decode it back into feature space: each index $j_{i_k,g}$ is replaced by the mean of group $g$ of $y_{i_k}$ over the training occurrences of concept $i_k$ that received it, and a ridge regression fitted on the training split maps the result back to feature space, giving $r^{\mathrm{dec}}_{i_k}$. The images of Figures~\ref{fig:app-quant-pin}--\ref{fig:app-quant-flowers} (Appendix~\ref{app:qualitative-quant-examples}) are held out from both.

The quantized attribute also keeps which class an occurrence of a concept comes from: in Appendix~\ref{app:qualitative-attrspace}, the occurrences of one concept in the evaluation images group by the class of their image (Figures~\ref{fig:app-attrspace-pin}--\ref{fig:app-attrspace-dogs}).

\subsection{User study details and results}
\label{app:user-study}

\begin{figure}[t]
    \centering
        \centering
        \includegraphics[width=0.8\linewidth]{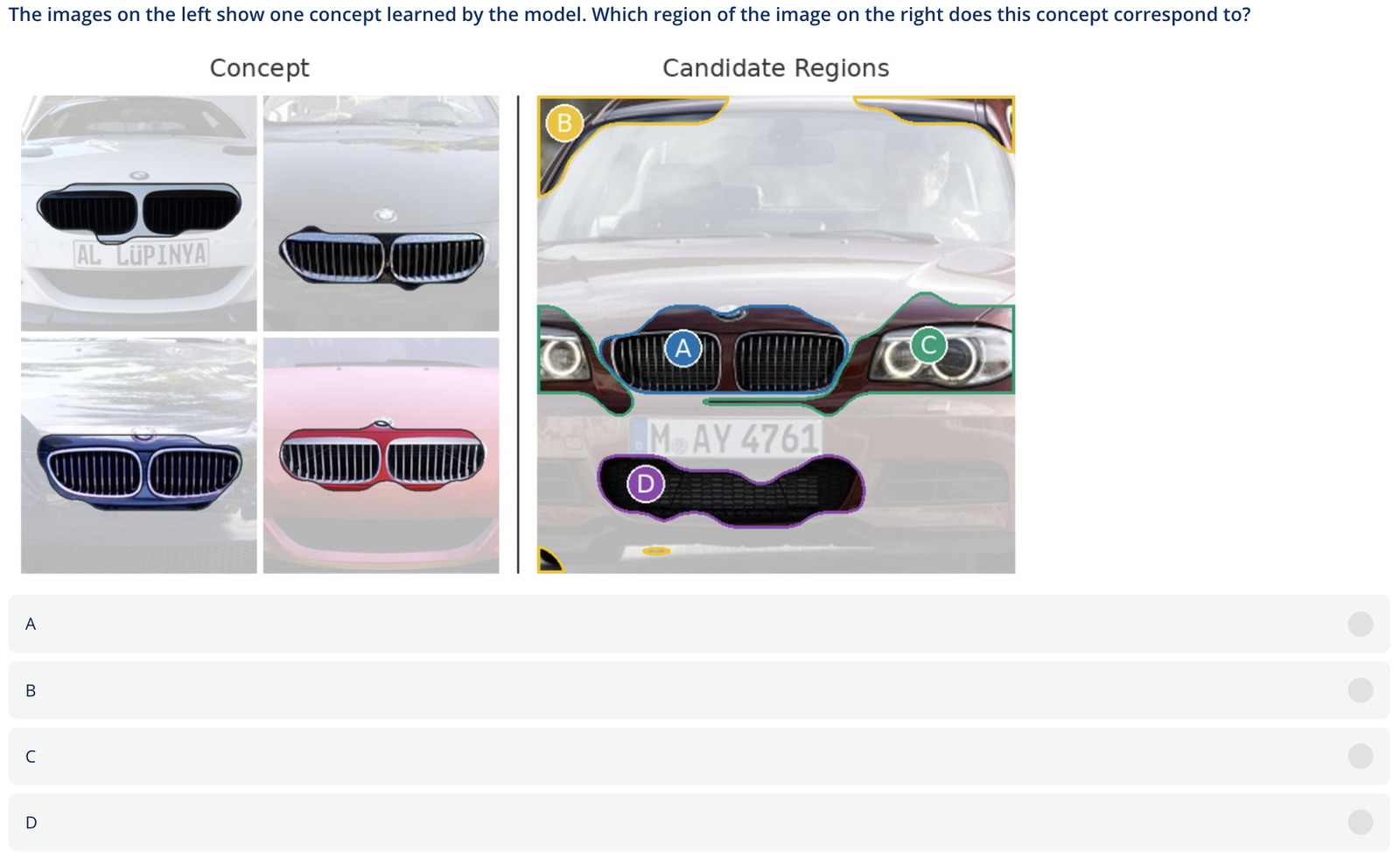}
        \caption{Example questions shown in the survey briefing for Q2: part retrieval.}
        \label{fig:us_examples2}
\end{figure}

\begin{figure}[t]
        \centering
        \includegraphics[width=0.8\linewidth]{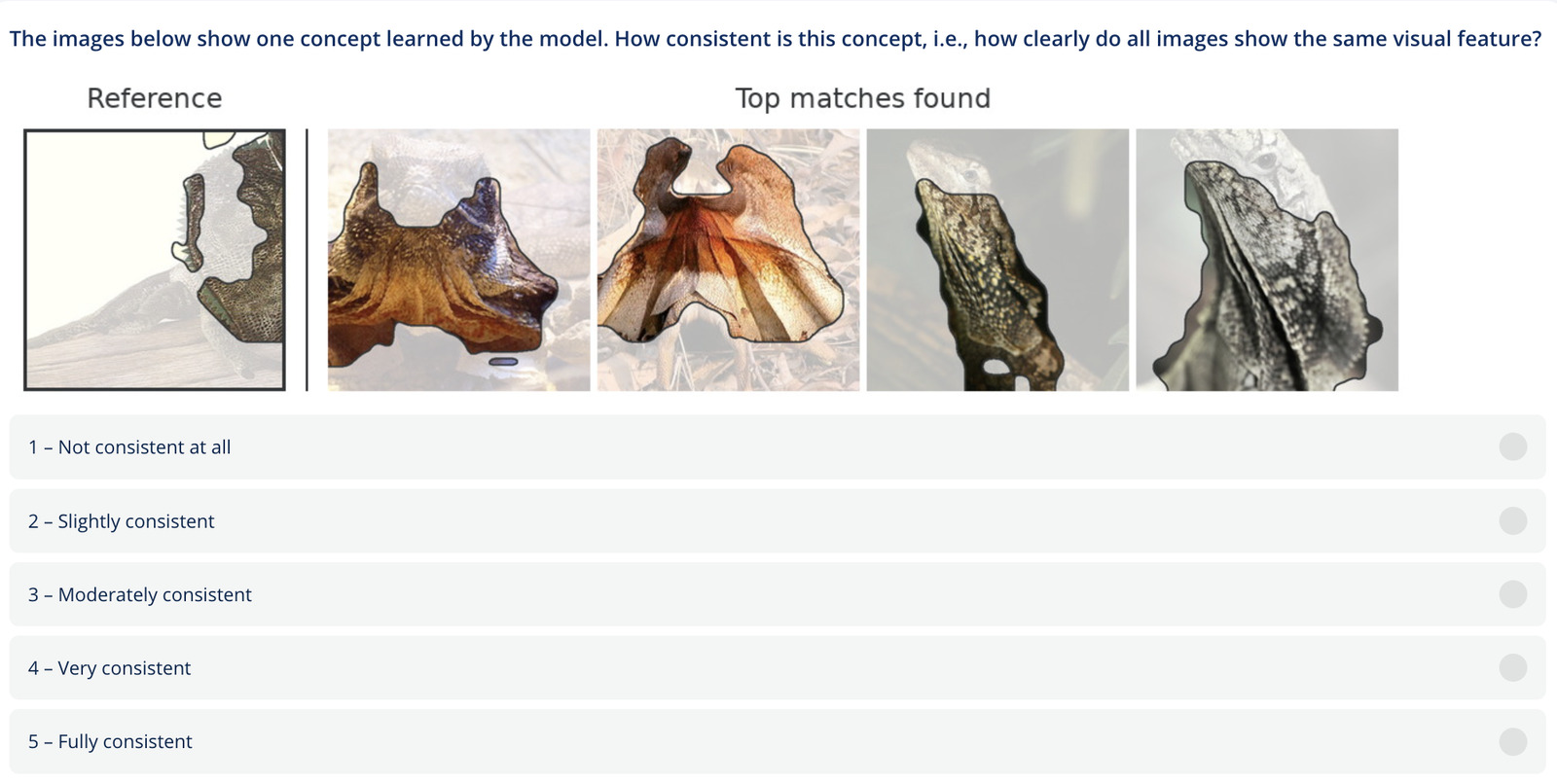}
        \caption{Example questions shown in the survey briefing for Q1: part consistency.}
    \label{fig:us_examples1}
\end{figure}

\paragraph{Setup.}
We recruited 120 native English speakers through the \url{clickworker.com} platform, balanced in terms of gender and age (18-64). The study was fully anonymous, no personal data was collected, and it followed our institution's guidelines for research involving human participants. %
The median completion time was 8 minutes 42 seconds. Before starting, participants read a short briefing describing both question types, with an annotated example for each (Fig.~\ref{fig:us_examples1} and Fig.~\ref{fig:us_examples2}). Each participant then answered 42 questions: 21 part-retrieval questions (Q2) and 21 part-consistency questions (Q1). For both question types, the 21 questions contained 7 prototypes from each of DisParQ, PIP-Net, and CFM, drawn from all three datasets, presented in random order and with the method identity hidden. We used three survey versions with different prototype sets, each completed by 40 participants. For Q1, the per-method CUB/Cars/PartImageNet counts were 3/3/1, 3/1/3 and 1/3/3 across versions. Prototypes were selected by sampling random test images and taking one of the four most important prototypes for the image's prediction. Here, four specifies the candidate set for stimulus selection, not the model's concept budget. The same images were used for all three methods, and all prototypes were visualized in the same way, by their top activating image patches. To ensure data quality, we screened responses for inconsistent answering and excluded 4\% of participants, recruiting replacements until 120 valid responses were collected.


\begin{samepage}
\paragraph{Q1: Part consistency.}
Participants were shown the visualization of a single prototype and asked: \emph{``The images below show one concept learned by the model. How consistent is this concept, i.e., how clearly do all images show the same visual feature?''} They answered on a 5-point scale (1: not consistent at all, 2: slightly, 3: moderately, 4: very, 5: fully consistent) and were told that there are no right or wrong answers. For each participant, we average ratings per method, overall and per dataset. We compare DisParQ with each baseline using two-sided paired Wilcoxon signed-rank tests, with Holm correction across all eight comparisons (two baselines, overall and three datasets).
\par
\end{samepage}

Participants rated DisParQ prototypes as the most consistent overall, with a significant advantage over both PIP-Net ($p<0.001$) and CFM ($p<0.001$). Per dataset (Fig.~\ref{fig:us_consistency_per_dataset}), the gap is largest on CUB, where DisParQ is significantly better than both baselines. On Cars it significantly outperforms CFM, while on PartImageNet all three methods are rated similarly low, suggesting that consistent parts are harder to judge on this more diverse dataset.

\begin{figure}[t]
    \centering
    \begingroup
\definecolor{ink}{HTML}{303437}
\definecolor{studyblue}{HTML}{8AAEC3}
\definecolor{studypeach}{HTML}{C49A76}
\definecolor{studygray}{HTML}{929799}
\begin{tikzpicture}[x=12.5mm,y=9mm,
  every node/.style={font=\scriptsize,text=ink,inner sep=1pt}]
  \node[rotate=90] at (-1.05,3) {Mean score per rater};
  \foreach \offset/\dataset/\slug in {0/CUB/cub,3.7/Cars/cars,7.4/PartImageNet/partimagenet} {
    \begin{scope}[shift={(\offset,0)}]
      \node[font=\scriptsize\bfseries,anchor=south] at (.975,6.05) {\dataset};
      \foreach \y in {1,...,5} {
        \draw[black!12,line width=.3pt] (-.5,\y) -- (2.45,\y);
        \node[anchor=east] at (-.55,\y) {\y};
      }
      \draw[ink,line width=.45pt] (-.5,5) -- (-.5,1) -- (2.45,1);
      \input{\RepoFigureRoot/figures/user_study/violin_\slug _data.tex}
      \foreach \x/\method in {0/\ours{},1/PIP-Net,2/CFM}
        \node[anchor=north] at (\x,.85) {\method};
      \input{\RepoFigureRoot/figures/user_study/violin_\slug _stats.tex}
    \end{scope}
  }
\end{tikzpicture}
\endgroup
    \caption{\textbf{Part consistency ratings per dataset.} Each point is one participant's mean rating (1: not consistent at all, 5: fully consistent) over the prototypes of a given method shown from that dataset ($n{=}120$). Violins show the distribution, diamonds the mean, and black bars the interquartile range. Brackets mark comparisons in which DisParQ is rated significantly higher than a baseline (two-sided paired Wilcoxon signed-rank test, Holm-corrected): $^{*}p<0.05$, $^{**}p<0.01$, $^{***}p<0.001$; unmarked comparisons are not significant.}
    \label{fig:us_consistency_per_dataset}
\end{figure}

\begin{samepage}
\paragraph{Q2: Part retrieval.}
Inspired by HIVE~\citep{kim2022hive}, this task measures how well humans can predict the model's reasoning from its explanations. Participants were shown the visualization of a prototype on the left and a new image with four lettered regions on the right, all identified by the model, and asked: \emph{``The images on the left show one concept learned by the model. Which region of the image on the right does this concept correspond to?''} Participants could select one of the four regions or ``None of the above''. The correct answer is the region in which the model localizes the prototype. For each participant, we compute the accuracy per method, overall and per dataset, and compare DisParQ with each baseline using a two-sided paired Wilcoxon signed-rank test with Holm correction. We test each method against a chance level of 25\% with a one-sided Wilcoxon signed-rank test; this is conservative, as five answer options were available.
\par
\end{samepage}

All methods lead to above-chance accuracy, but participants were most accurate with DisParQ prototypes on every dataset, reaching 69\% overall compared to 60\% for PIP-Net and 52\% for CFM (Tab.~\ref{tab:us_retrieval}). The advantage is significant against both baselines overall and on every dataset except CUB, where DisParQ and PIP-Net perform comparably. It is most pronounced on PartImageNet, where DisParQ outperforms PIP-Net by 16 percentage points. A logistic regression with standard errors clustered by participant confirms the result: with DisParQ, the odds of a correct answer are 1.5 and 2.1 times higher than with PIP-Net and CFM, respectively ($p<10^{-4}$).

\begin{table}[t]
\centering
\caption{Part retrieval accuracy (\%). Participants chose which of four model-identified regions matches the visualized concept, or ``None of the above'' (chance: 25\% or lower); 120 participants. All methods are significantly above chance ($p<10^{-4}$, one-sided Wilcoxon). Stars mark baselines that DisParQ significantly outperforms (paired Wilcoxon signed-rank test on per-participant accuracy, Holm-corrected): $^{*}p<0.05$, $^{**}p<0.01$, $^{***}p<0.001$.}
\label{tab:us_retrieval}
\begin{tabular}{lcccc}
\toprule
Method & All & CUB & Cars & PartImageNet \\
\midrule
Random  & 25.00 & 25.00 & 25.00 & 25.00 \\
\midrule
CFM     & 51.50$^{***}$ & 50.00$^{**}$ & 56.67$^{***}$ & 48.57$^{**}$ \\
PIP-Net  & 60.12$^{***}$ & 63.57 & 63.57$^{*}$ & 53.21$^{***}$ \\
DisParQ & \textbf{69.05} & \textbf{65.00} & \textbf{72.86} & \textbf{69.29} \\
\bottomrule
\end{tabular}
\end{table}

\providecommand{\sidblue}{\iftoggle{comments}{\color[HTML]{1F5FBF}}{}}
\section{Qualitative results}
\label{app:qualitative}
This section first describes our interactive concept visualizer to browse concepts from \ours{} (Sections~\ref{app:qualitative-visualizer}). It then compares the concept maps of \ours{} with those of the baselines (Sections~\ref{app:qualitative-pin}--\ref{app:retrieval}), and shows what the description of \ours{} contains: the concepts behind a prediction (Section~\ref{app:qualitative-ours}) and their attributes (Sections~\ref{app:qualitative-quant-examples} and~\ref{app:qualitative-attrspace}).

\subsection{Interactive Concept Visualizer}
\label{app:qualitative-visualizer}

\begin{figure*}[!htbp]
  \centering
  \includegraphics[width=\linewidth]{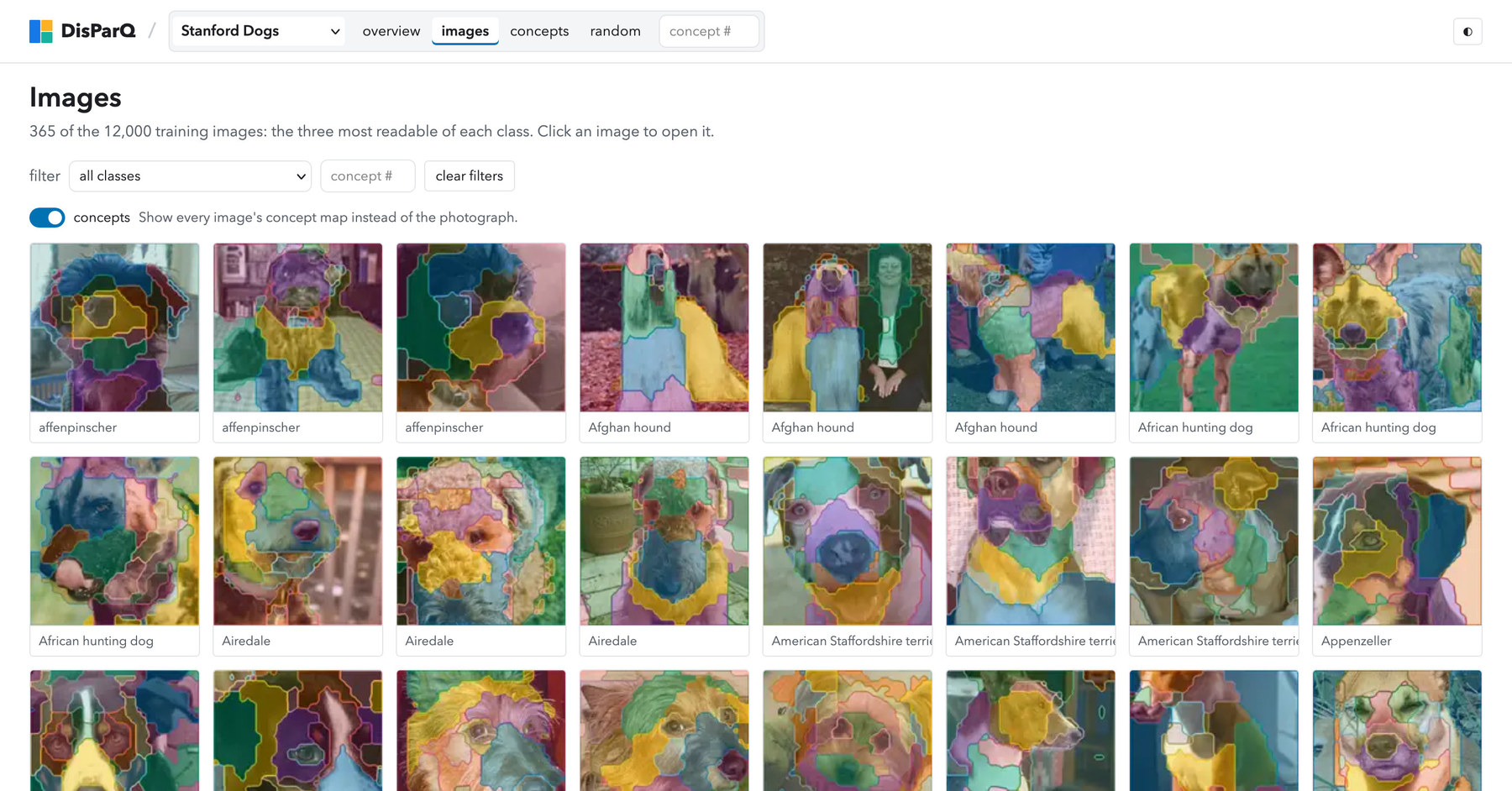}\\[2mm]
  \includegraphics[width=\linewidth]{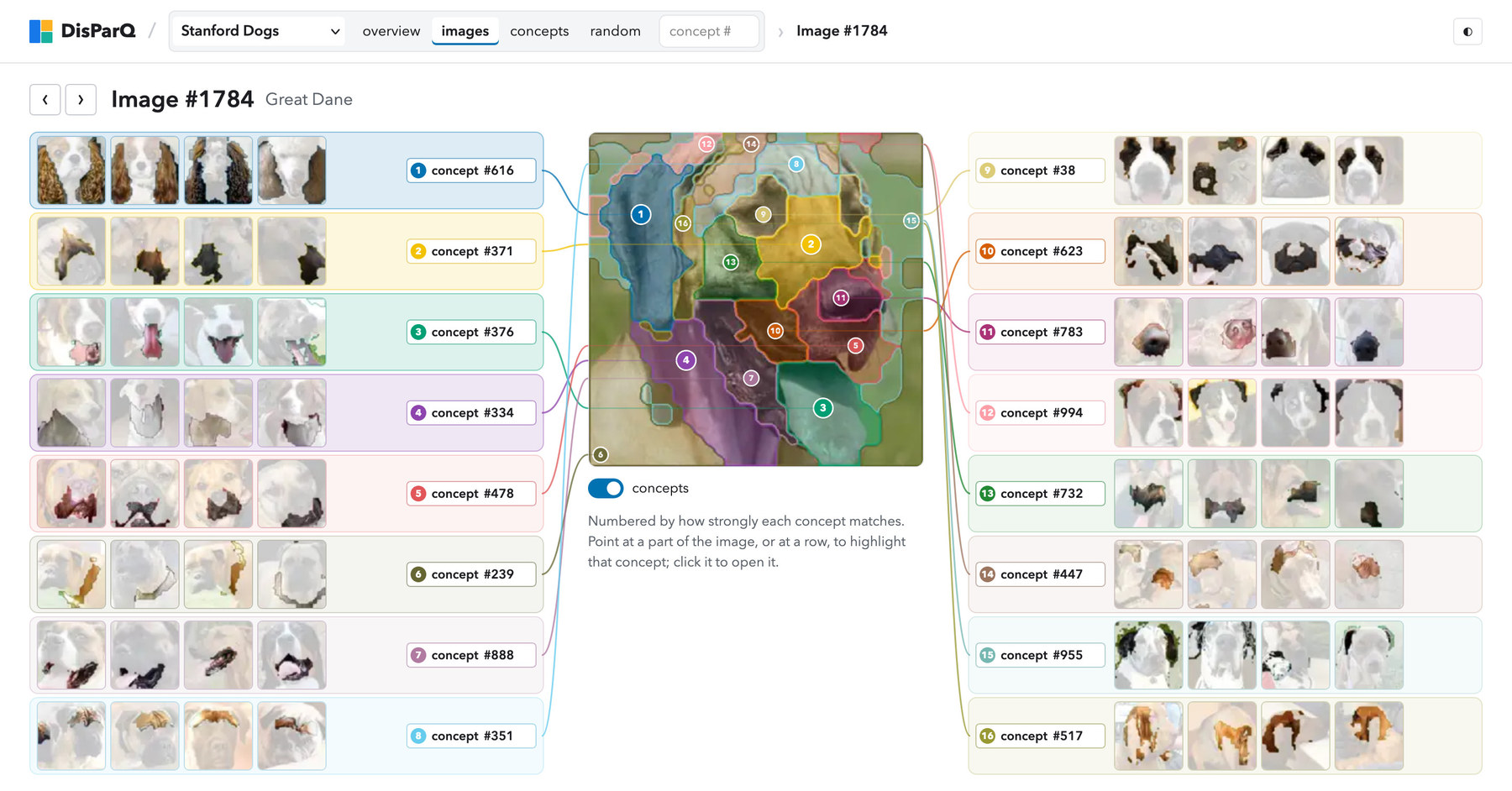}%
  \vspace{-2.3mm}%
  \caption{\textbf{Visualizer: the image gallery and one image.} \emph{Top:}~The training images of one dataset, filtered by class or by concept, with an overlay of each image's concept map. \emph{Bottom:}~One image, opened from the gallery: its concept map in the middle, its concepts numbered by match strength. Each concept's row shows the training images where it matches most strongly, and an arrow joins the row to the concept's region.}
  \label{fig:app-visualizer-images}
\end{figure*}

We provide an interactive concept visualizer at \url{https://disparq.gmum.net/} that allows browsing concepts from \ours{}. The visualizer includes concept banks for five datasets (PartImageNet, CUB, Dogs, Cars, and Flowers), along with an image gallery that shows what concepts activate for an image and where. We describe the two galleries below:

\myparagraph{Image gallery.} This consists of three images per class (Figure~\ref{fig:app-visualizer-images}) for each dataset. The gallery view (Figure~\ref{fig:app-visualizer-images}, top) shows images with a concept overlay, and can be filtered by class or concept ID. Each image can be selected to open the per-image view, that shows the top-16 concepts, where they activate, and what they are, characterized by other exemplars from the dataset that contain the same concept (Figure~\ref{fig:app-visualizer-images}, bottom). The visualizer also contains an option to view the per-image view of a random image from the selected dataset.

\begin{figure*}[!htbp]
  \centering
  \includegraphics[width=\linewidth]{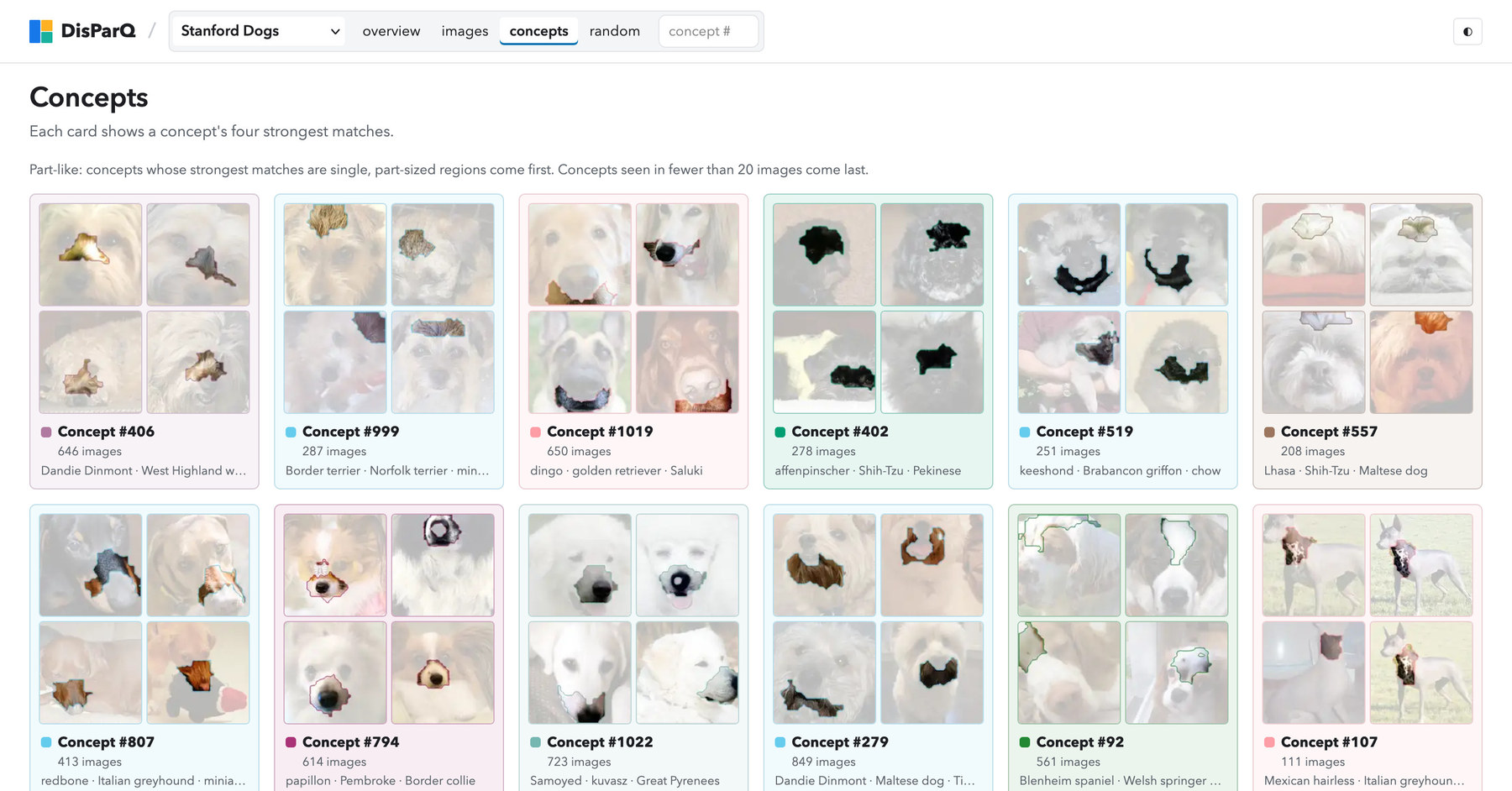}\\[2mm]
  \includegraphics[width=\linewidth]{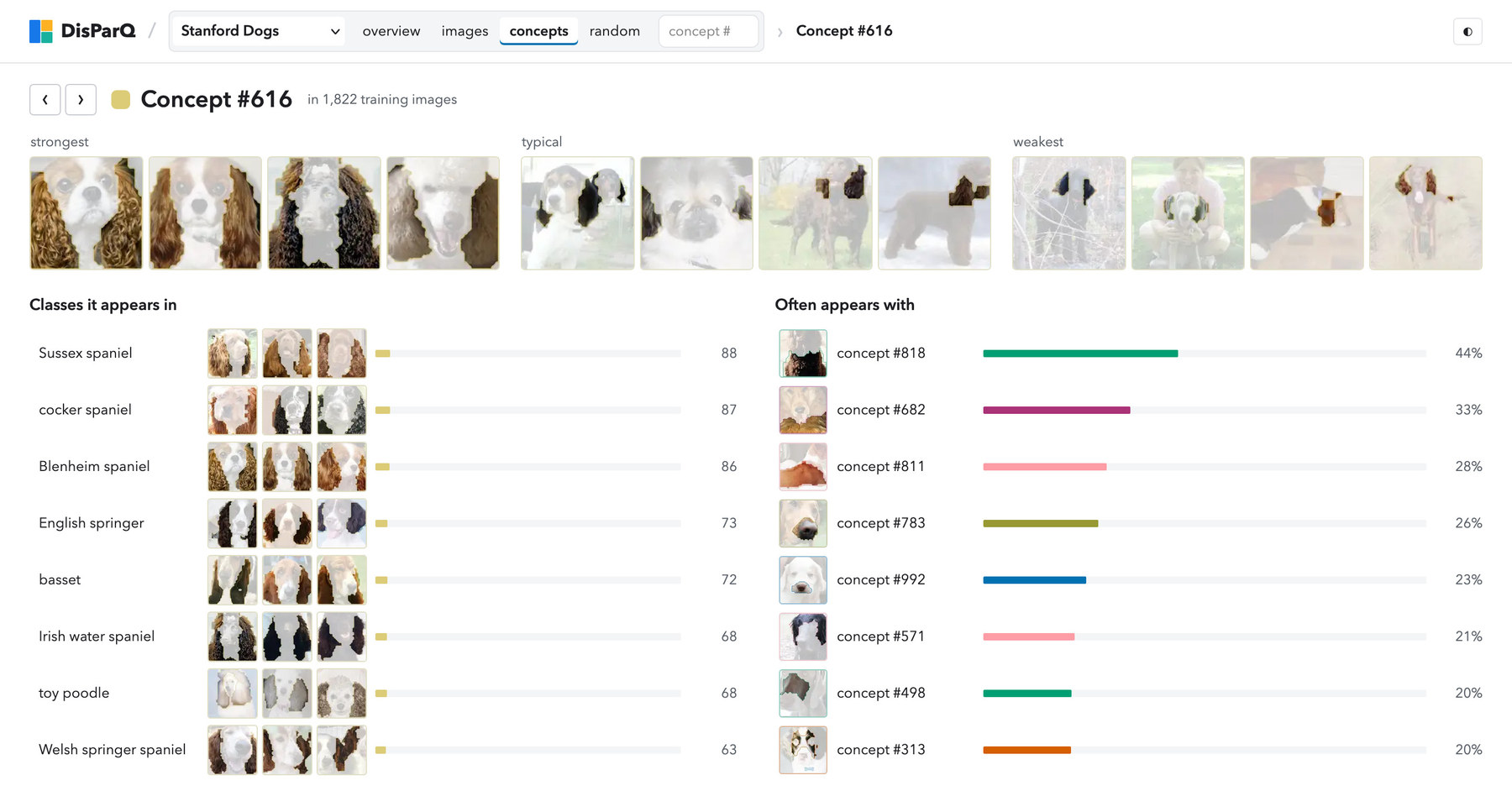}%
  \vspace{-2.3mm}%
  \caption{\textbf{Visualizer: the concept gallery and one concept.} \emph{Top:}~Every concept used on a dataset, each shown by its four strongest matches. \emph{Bottom:}~One concept, opened from the gallery: its strongest, typical and weakest matches, the classes it appears in with examples, and the concepts it most often appears with.}
  \label{fig:app-visualizer-concepts}
\end{figure*}

\myparagraph{Concept gallery.} This shows the full set of concepts that activate on at least one image in the dataset (Figure~\ref{fig:app-visualizer-concepts}). The gallery view (Figure~\ref{fig:app-visualizer-concepts}, top) shows all concepts from the dataset that are found in at least one image. Each concept card shows exemplars of images and where the concept activates on it, along with the number of images the concept activates on the primary classes they belong to. The per-concept view (Figure~\ref{fig:app-visualizer-concepts}, bottom) shows the images and regions within them that correspond to the strongest, typical, and weakest activations for the concept in the dataset. It also shows the top classes the concept appears in, and other concepts that most frequently co-occur with the concept.

\clearpage%

\subsection{PartImageNet}
\label{app:qualitative-pin}

\begin{figure*}[!htbp]
  \centering
  \includegraphics[width=\linewidth]{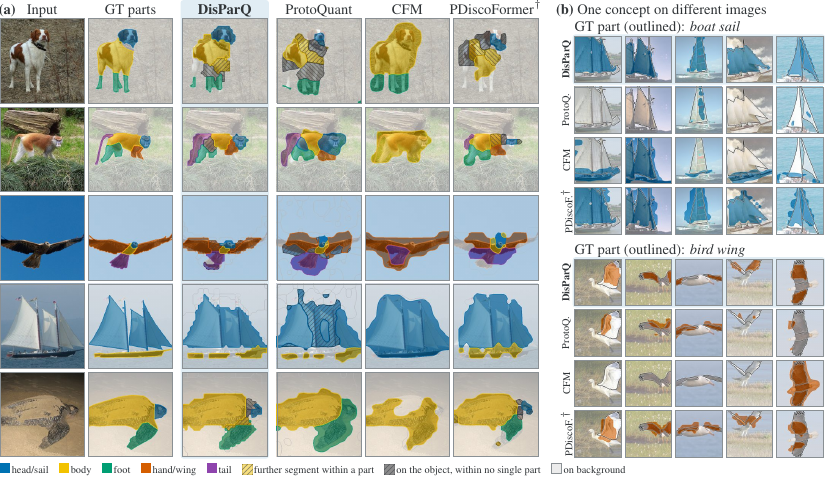}%
  \vspace{-2.3mm}%
  \caption{\textbf{Part maps, and one concept across images}, on PartImageNet. \emph{(a)}~\QualMapsLegendTextSub{}. \emph{(b)}~For a ground-truth part (outlined), the one concept of each method that is most consistent with that part over the validation split, shown on the same five images. The concept of \ours{} marks the part in every image; a baseline's best concept is absent from most of them, or covers the whole object.}
  \label{fig:app-qual-composite}
\end{figure*}

This subsection extends Figure~\ref{fig:qualitative}. All figures use the PartImageNet validation split and the models of Table~\ref{tab:pin-concepts}; every method sees the same center-cropped $224\times224$ input, and the ground-truth mask is cropped identically. Colors follow the rule of Figure~\ref{fig:qualitative}.
Figure~\ref{fig:app-qual-composite} shows five further images and, for two parts, the one concept of each method that is most consistent with that part over the validation split.

\begin{figure*}[!htbp]
  \centering
  \includegraphics[width=0.97\linewidth]{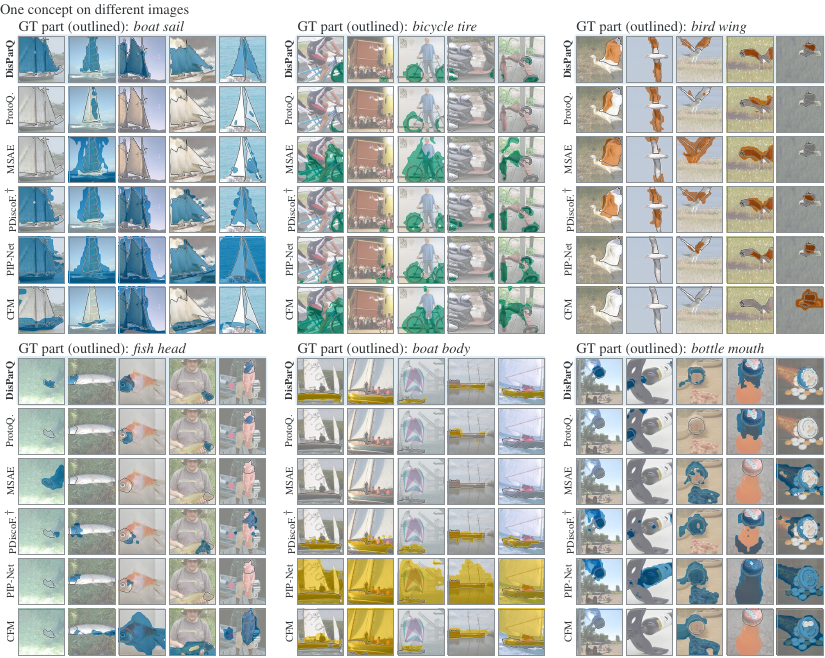}%
  \vspace{-2.3mm}%
  \caption{\textbf{One concept, different images.} Each row shows, for the outlined ground-truth part, the concept of that method most consistent with it over the validation split. $^{\dagger}$\,trained with class labels.}
  \label{fig:app-qual-concepts}
\end{figure*}

\begin{figure*}[p]%
  \centering
  \includegraphics[width=0.73\linewidth]{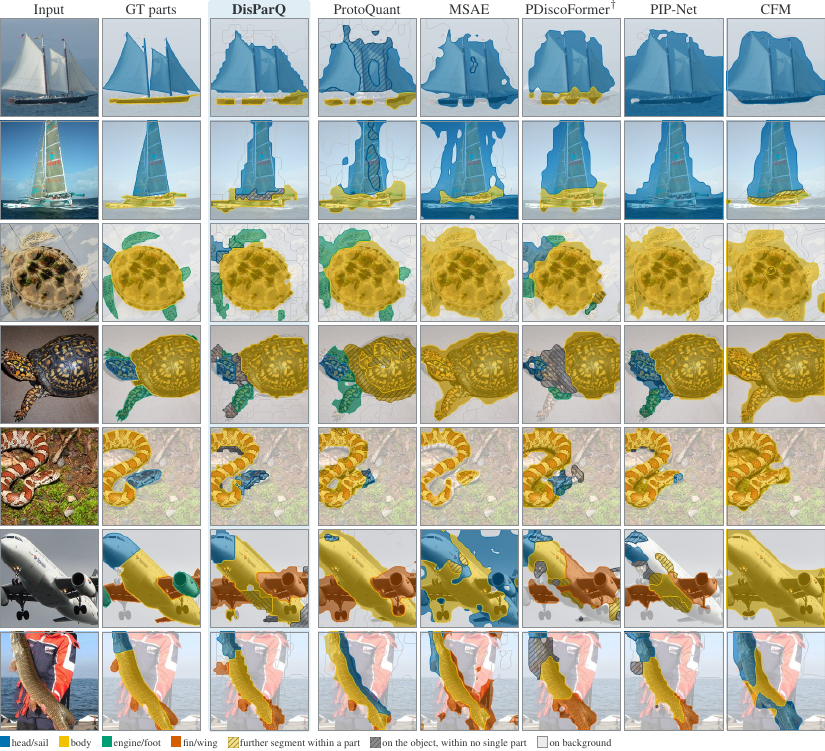}%
  \vspace{-2.3mm}%
  \caption{\textbf{Further examples} on PartImageNet, all five baselines. Maps and colors as in Figure~\ref{fig:qualitative}. $^{\dagger}$\,trained with class labels.}
  \label{fig:app-qual-selected}
\end{figure*}

\begin{figure*}[p]%
  \centering
  \includegraphics[width=\linewidth]{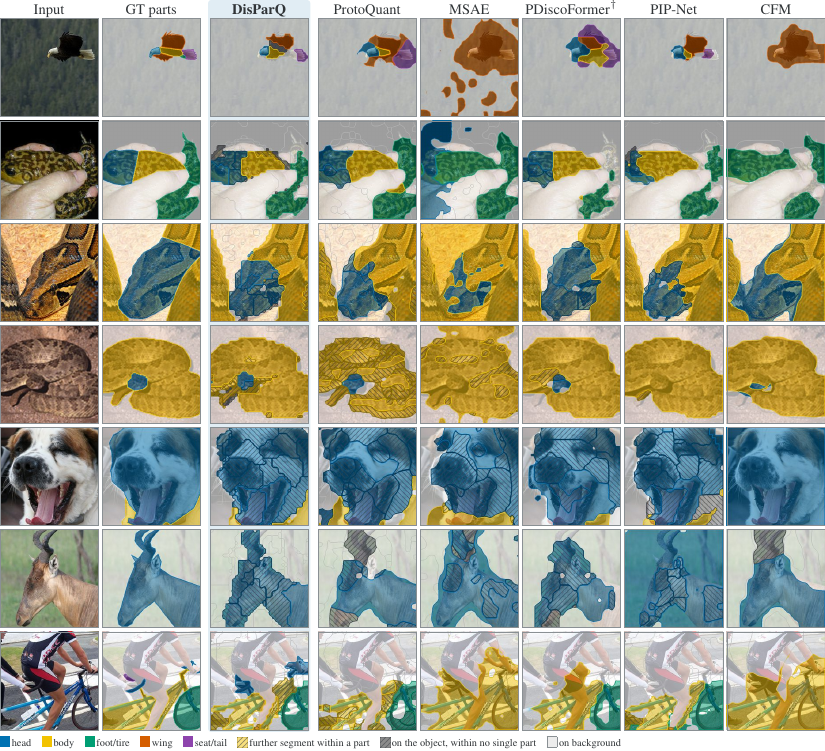}%
  \vspace{-2.3mm}%
  \caption{\textbf{Random examples} from the PartImageNet validation split. Maps and colors as in Figure~\ref{fig:qualitative}. $^{\dagger}$\,trained with class labels.}
  \label{fig:app-qual-random}
\end{figure*}

Figures~\ref{fig:app-qual-selected} and~\ref{fig:app-qual-random} compare \ours{} with five baselines: ProtoQuant~\citep{janusz2026protoquant}, MSAE~\citep{zaigrajew2025interpreting}, PDiscoFormer~\citep{aniraj2024pdiscoformer}, PIP-Net~\citep{nauta2023pip} and CFM~\citep{wittenmayer2026cfm}. \ours{} follows the annotated parts and stops at the object boundary: it separates the hull of a sailboat from its sails, and the head and flippers of a turtle from its shell. The baselines merge parts into one concept (MSAE, PIP-Net and CFM cover a turtle with the color of its shell) or spread into the background (MSAE and PIP-Net fill the sea and sky around a sailboat). On random images (Figure~\ref{fig:app-qual-random}), \ours{} still recovers small parts such as a snake's head, while MSAE marks the background and CFM covers the object with a single concept. Figure~\ref{fig:app-qual-concepts} follows one concept of each method across five images: the concept of \ours{} marks the sail, the hull and the bottle mouth in all five images and the tire, the wing and the fish head in most, whereas the best-matching concept of a baseline is missing from several images or covers the whole object.\par

\begin{figure*}[!h]
  \centering
  \includegraphics[width=\linewidth]{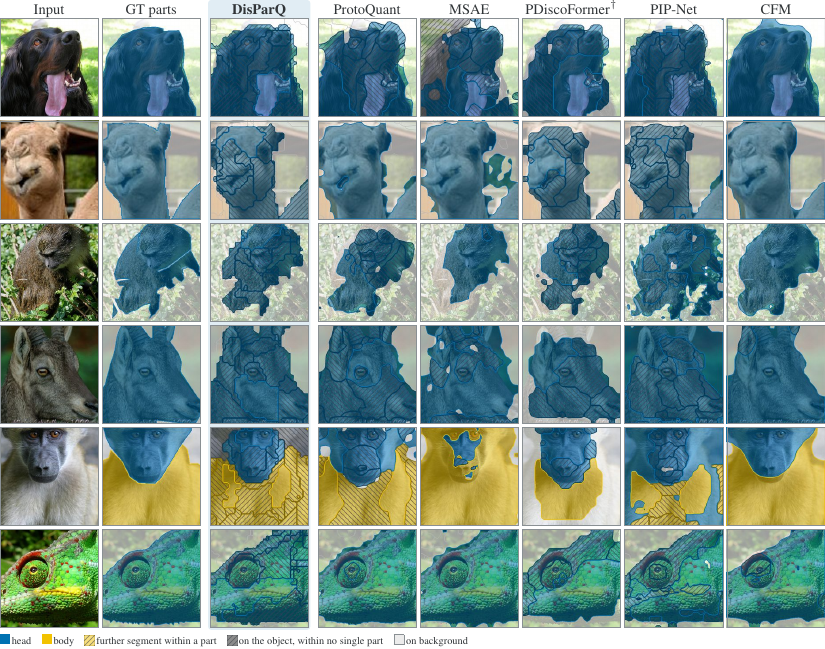}%
  \vspace{-2.3mm}%
  \caption{\textbf{Failure cases}: images on which a baseline reaches a higher per-image mIoU than \ours{}. Maps and colors as in Figure~\ref{fig:qualitative}. $^{\dagger}$\,trained with class labels.}
  \label{fig:app-qual-failures}
\end{figure*}

\myparagraph{Where \ours{} fails.}
Figure~\ref{fig:app-qual-failures} shows the images with the largest per-image deficit against the best baseline.
They share one cause: close-ups in which a single annotated part (almost always \emph{head}) fills the frame.
\ours{} describes such an image with many finer concepts like snout, eyes, ears, fur, of which only one can be matched to the single ground-truth part, while a method with one coarse concept per object matches it trivially. We see this as a limitation of the annotation granularity of human-labeled datasets rather than of the concepts: PartImageNet labels a head as one part, so a finer decomposition that is consistent across images is scored as over-segmentation.

\clearpage%

\clearpage

\subsection{Fine-grained datasets}
\label{app:qualitative-other}

This subsection repeats Figure~\ref{fig:qualitative} on four fine-grained datasets: CUB~\citep{wah2011caltech}, Stanford Cars~\citep{krause2013collecting}, Stanford Dogs~\citep{khosla2011novel} and Oxford Flowers~\citep{nilsback2008automated}, with the baselines trained on each dataset (Figures~\ref{fig:app-qual-cub}--\ref{fig:app-qual-flowers}). All images come from the test split of each dataset. These datasets have no part masks, so a segment is colored by where it sits on the object, by the same rule in every column; the reference column shows the foreground and, on CUB, the annotated keypoints. %
Panel (b) follows two concepts of \ours{}, each across five images, beside each baseline's best-matching concept.
The maps of \ours{} stay on the object and divide it into regions that recur across images, while MSAE also covers the background. The concept of \ours{} marks the same part in all five images, for example the tail or the head of a bird, while the best-matching concept of MSAE covers whole birds. PDiscoFormer is trained with class labels and shown for reference.

\newcommand{\QualOtherCaption}[1]{\textbf{#1: \ours{} and baselines.} \emph{(a)}~Hard concept maps, colored by where a segment sits on the object (same rule in every column, no part meaning); pale outlined segments lie on the background. \emph{(b)}~Two concepts of \ours{} (outlined), each beside the concept of every baseline that agrees best with it. $^{\dagger}$\,trained with class labels.}
\begin{figure*}[!hbp]\centering \includegraphics[width=0.80\linewidth]{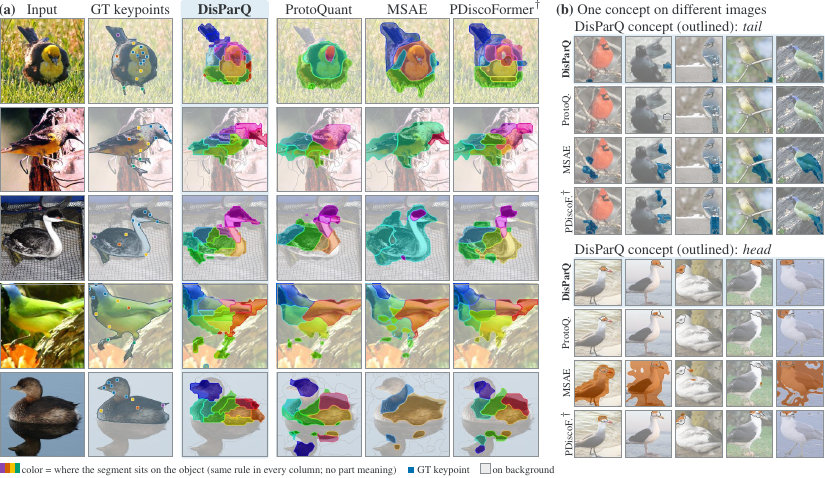}%
\vspace{-2.3mm}%
\caption{\QualOtherCaption{CUB}}             \label{fig:app-qual-cub}\end{figure*}
\begin{figure*}[!hbp]\centering \includegraphics[width=0.80\linewidth]{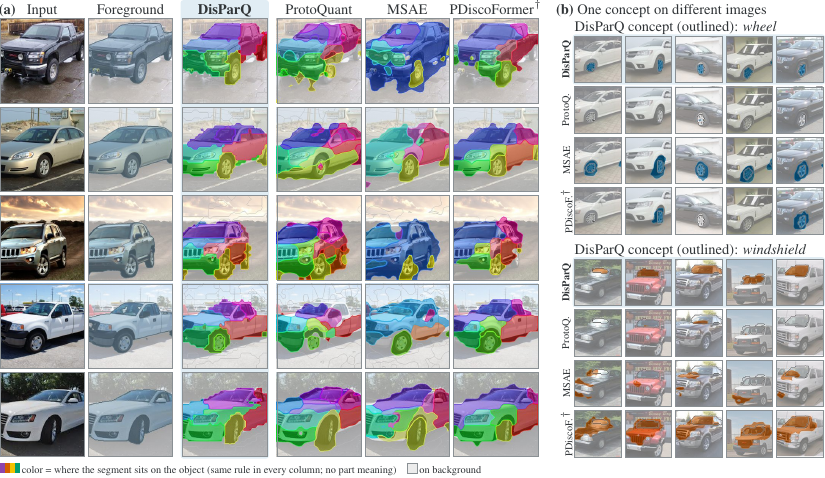}%
\vspace{-2.3mm}%
\caption{\QualOtherCaption{Stanford Cars}}   \label{fig:app-qual-cars}\end{figure*}
\begin{figure*}[!hbp]\centering \includegraphics[width=\linewidth]{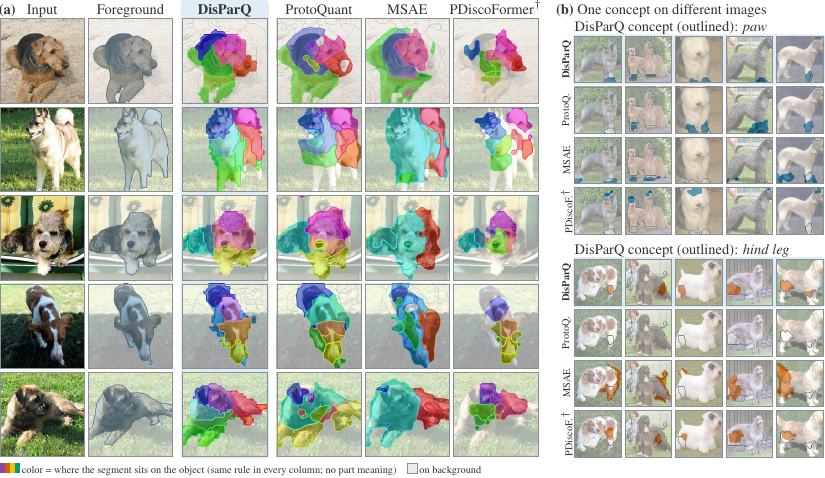}%
\vspace{-2.3mm}%
\caption{\QualOtherCaption{Stanford Dogs}}   \label{fig:app-qual-dogs}\end{figure*}
\begin{figure*}[!hbp]\centering \includegraphics[width=\linewidth]{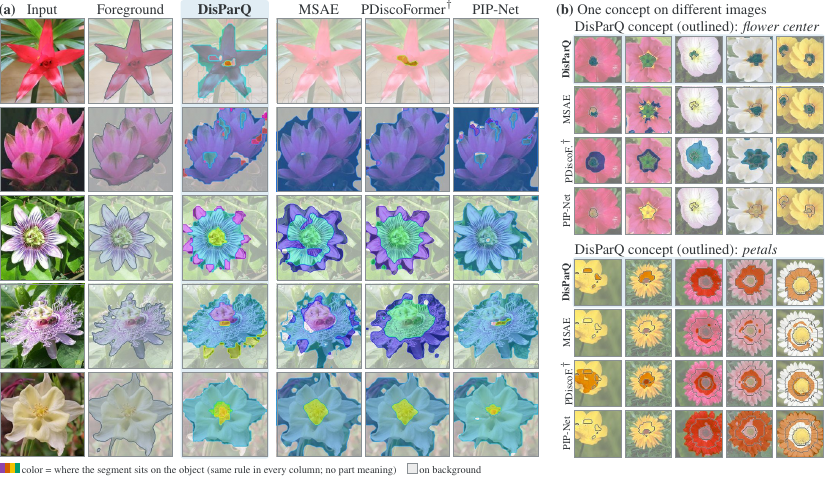}%
\vspace{-2.3mm}%
\caption{\QualOtherCaption{Oxford Flowers}}  \label{fig:app-qual-flowers}\end{figure*}

\clearpage

\subsection{Part-to-part retrieval}
\label{app:retrieval}

This subsection shows part-to-part retrieval, the benchmark of Table~\ref{tab:part-retrieval}, defined in Appendix~\ref{app:metrics}, on all five datasets (Figure~\ref{fig:app-retrieval}). For a query region (left), each row gives the four regions of other images that a method ranks highest by the concept histogram of the region. On PartImageNet and CUB, a green border marks a retrieval that carries the query's annotated part and a red border a miss. Queries and retrievals come from held-out images (the PartImageNet validation split and the test splits of the other datasets), and all methods describe the same regions. A region is an annotated part mask on PartImageNet and a keypoint group on CUB; on Cars, Dogs and Flowers, which have no part annotation, it is a concept of \ours{}.\par
All four retrievals of \ours{} carry the query's part in each of the six scored panels, while at most two of the four retrievals of ProtoQuant or MSAE do, as for the snake head and for the head and leg of a bird. PDiscoFormer, trained with class labels, also retrieves the bird parts. On Cars, Dogs and Flowers, which have no part annotation, \ours{} retrieves windshields, bumpers, grilles, muzzles and petals like the query, while the baselines return regions of mixed appearance.\par

\begin{figure*}[!hbp]%
  \centering
  \includegraphics[width=\linewidth]{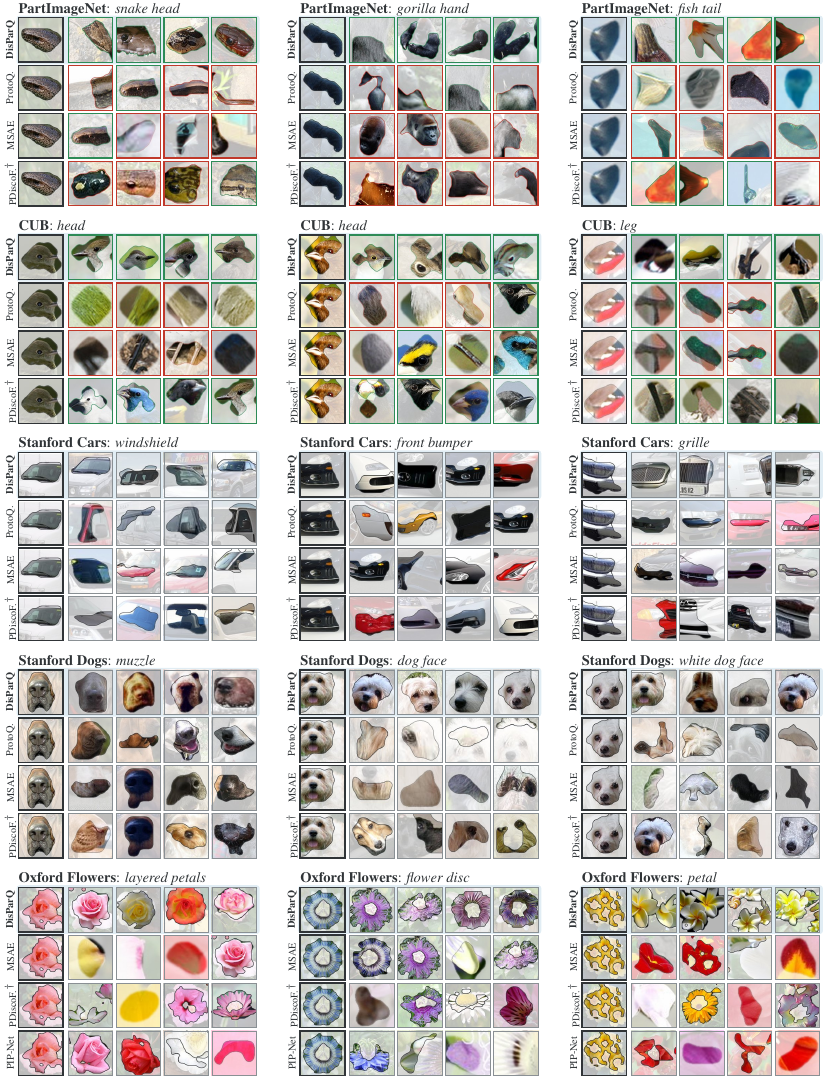}%
  \vspace{-2.3mm}%
  \caption{\textbf{Part-to-part retrieval on five datasets.} Each panel is one query: the region on the left (dark border) and, per method (rows, \ours{} first), the four highest-ranked regions of other images. Green or red border: the retrieval does or does not carry the query's annotated part (PartImageNet and CUB); Cars, Dogs and Flowers are unscored and their region names are ours. $^{\dagger}$\,trained with class labels.}
  \label{fig:app-retrieval}
\end{figure*}

\clearpage

\subsection{Class evidence of \ours{} and interpretability}
\label{app:qualitative-ours}

This subsection extends Figure~\ref{fig:teaser} to more images, for \ours{} on DINOv2-B/14~\citep{oquab2023dinov2} (Figures~\ref{fig:app-ours-nn-pin}--\ref{fig:app-ours-nn-flowers}). Predictions use the separate explanation classifier in Appendix~\ref{app:metrics}. The left panels show the four concepts with the largest positive drops in the predicted-class logit. The right panels show their nearest training occurrences of the same concept, ranked by cosine similarity between attributes. The images on the left come from the PartImageNet validation split and the test splits of the other datasets.

On PartImageNet (Figure~\ref{fig:app-ours-nn-pin}), the concepts for a Jeep, a warplane or a Komodo dragon retrieve the same part in other images and classes. On CUB (Figure~\ref{fig:app-ours-nn-cub}), the green head of a mallard retrieves the heads of other mallards, the head and wing of a blue jay account for $81\%$ of its positive logit drops, and the yellow wing bar of a European goldfinch retrieves wings with the same bar. On Stanford Cars (Figure~\ref{fig:app-ours-nn-cars}), the evidence lies on parts a person would name: the number plate and tail lights of a BMW X5 retrieve number plates and tail lights of other cars, and the headlight and wheel of a Volkswagen Beetle retrieve round headlights and wheel rims. On Stanford Dogs (Figure~\ref{fig:app-ours-nn-dogs}), it lies on the face and the coat, and the face concept of a Blenheim spaniel retrieves the faces of other spaniels. On Oxford Flowers (Figure~\ref{fig:app-ours-nn-flowers}), the corona and stamens of a passion flower account for $79\%$ of its positive logit drops, and the dark disc of a sunflower retrieves other sunflower discs.

\myparagraph{Interpretability.} Two regularities hold across the 25 examples. Positive logit drops are concentrated in a few concepts. The four largest contributions account for $56$ to $100\%$ of their total and the largest alone for $17$ to $52\%$. These normalized occlusion effects measure relative support under the intervention. The concepts are also shared across images, as each concept retrieves the same part in other images, often of other classes, so predictions for different classes are explained with one vocabulary of parts, and a prediction reads as a short list of recognizable parts with their weights. Appendix~\ref{app:metrics} (Concept contribution) defines how the contribution is measured.

\newcommand{\QualOursNNCaption}[1]{\textbf{#1: evidence concepts and what they retrieve.} \emph{Left:} the four concepts with the largest contribution to the predicted-class logit. \emph{Right:} each one cropped from this image (thick border) beside its nearest training occurrences.}
\begin{figure*}[p]\centering \QualBody{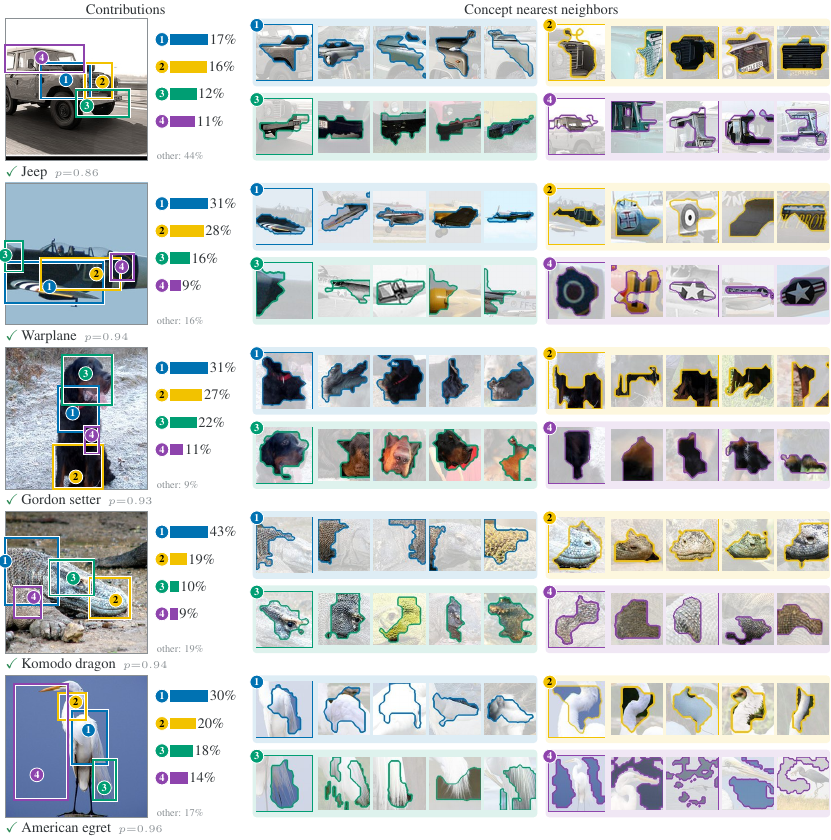}%
\vspace{-2.3mm}%
\caption{\QualOursNNCaption{PartImageNet}}   \label{fig:app-ours-nn-pin}\end{figure*}
\begin{figure*}[p]\centering \QualBody{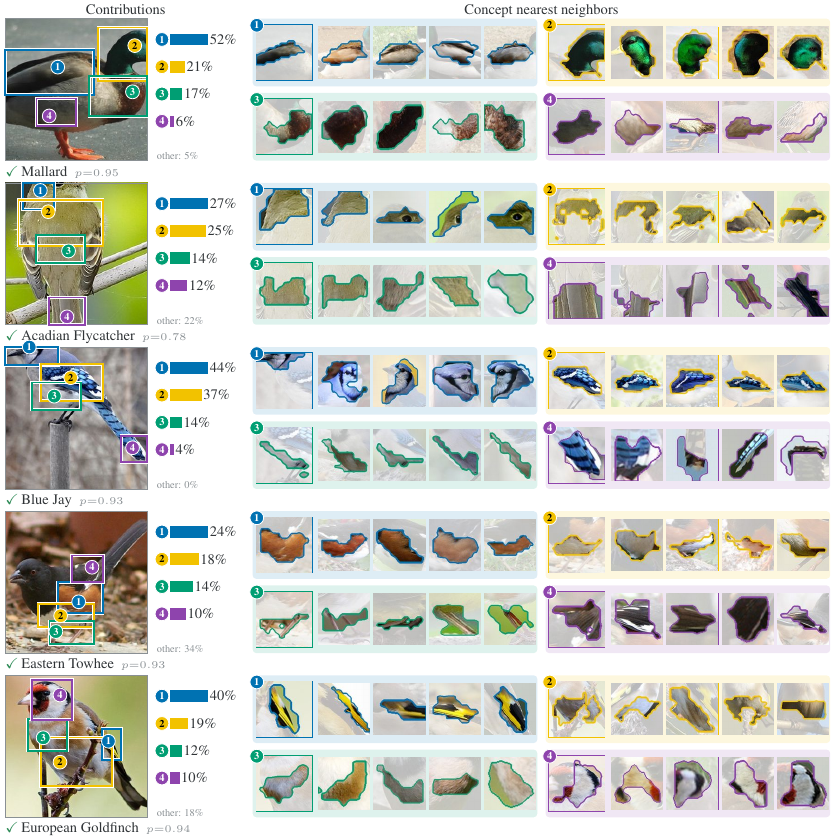}%
\vspace{-2.3mm}%
\caption{\QualOursNNCaption{CUB}}            \label{fig:app-ours-nn-cub}\end{figure*}
\begin{figure*}[p]\centering \QualBody{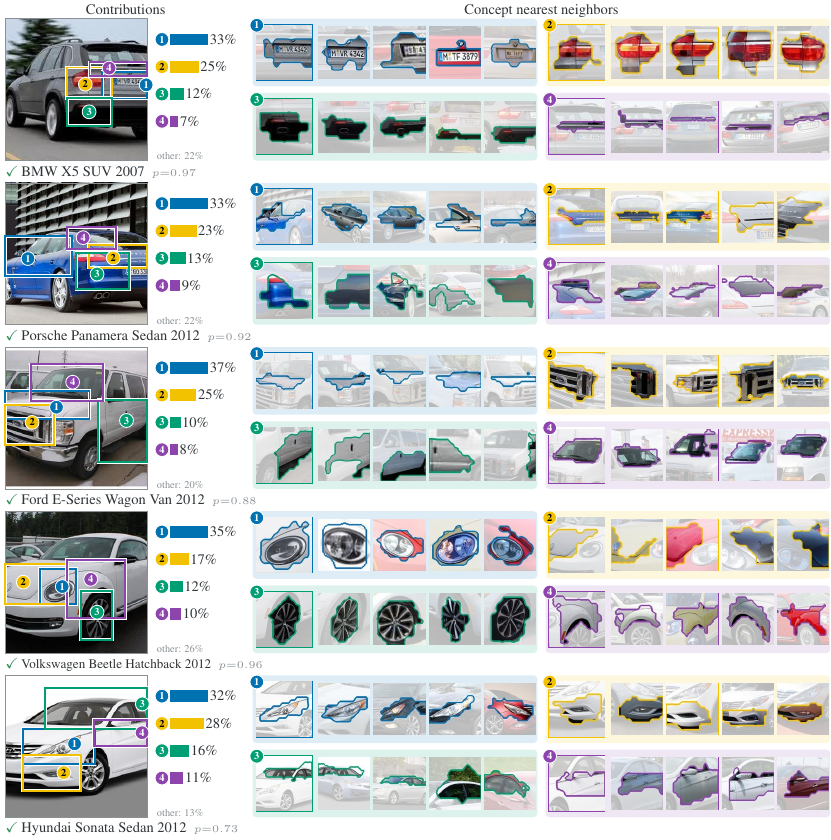}%
\vspace{-2.3mm}%
\caption{\QualOursNNCaption{Stanford Cars}}  \label{fig:app-ours-nn-cars}\end{figure*}
\begin{figure*}[p]\centering \QualBody{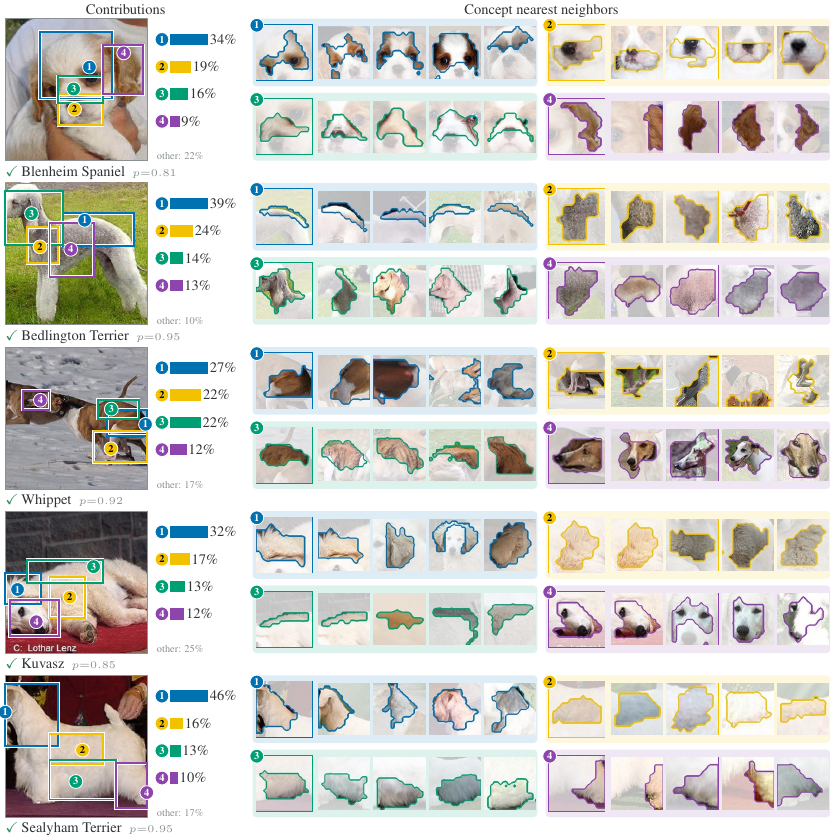}%
\vspace{-2.3mm}%
\caption{\QualOursNNCaption{Stanford Dogs}}  \label{fig:app-ours-nn-dogs}\end{figure*}
\begin{figure*}[p]\centering \QualBody{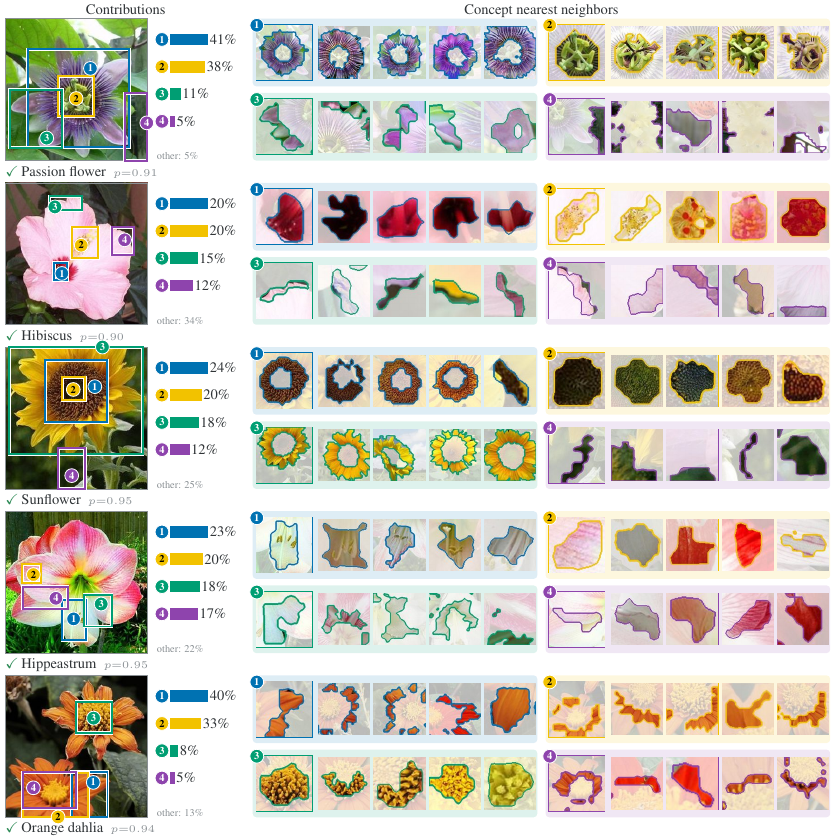}%
\vspace{-2.3mm}%
\caption{\QualOursNNCaption{Oxford Flowers}} \label{fig:app-ours-nn-flowers}\end{figure*}

\clearpage

\subsection{What quantization keeps}
\label{app:qualitative-quant-examples}

This subsection shows what the product-quantized attribute~\citep{jegou2011product} keeps of the continuous one (Figures~\ref{fig:app-quant-pin}--\ref{fig:app-quant-flowers}). To draw a code, we decode it back into feature space as $r^{\mathrm{dec}}_{i_k}$ using statistics of the training split (Appendix~\ref{app:qualitative-quant}). The images shown come from the PartImageNet validation split and the test splits of the other datasets. \emph{Top:} one PCA basis, fitted on the backbone's foreground patch features, maps features to colors, so colors compare across columns. The columns show the backbone features and each concept region drawn in three ways: as its concept key $\nu(w_{i_k})$, as key plus continuous attribute $\nu(w_{i_k})+r_{i_k}$, and as key plus quantized attribute $\nu(w_{i_k})+r^{\mathrm{dec}}_{i_k}$. \emph{Bottom:} for two concepts, the attributes of their occurrences before (left) and after (right) quantization, drawn as image crops at their coordinates in one PCA plane. The two panels are scaled independently, so what matters is which crops stay next to each other.\par
\emph{Top rows.} Drawn as its concept key alone, a region has the same color in every image: the key says which concept a region belongs to, not how the region looks. Adding the continuous attribute gives each region back the color of its own backbone features, and the quantized attribute gives nearly the same colors.
\emph{Bottom panels.} Quantization keeps the neighborhoods of the attribute space: $55$ to $78\%$ of the 5-nearest-neighbor relations (the share in each panel title) are the same before and after. The code therefore keeps which occurrences of a concept look alike, which Section~\ref{app:qualitative-attrspace} examines over all occurrences of a concept.\par

\newcommand{\QualQuantFigure}[3]{\begin{figure*}[p]\centering \QualBody{#1_pca_maps}\\[2.5mm] \QualBody{#1_pca_attr}%
\vspace{-2.3mm}%
\caption{\textbf{#2: what quantization keeps.} \emph{Top:} backbone features, and each concept region as its key, key + continuous attribute and key + quantized attribute, in one PCA basis. \emph{Bottom:} two concepts' attribute spaces before and after quantization (same crops and basis; panels scaled independently).}
\label{#3}\end{figure*}}
\QualQuantFigure{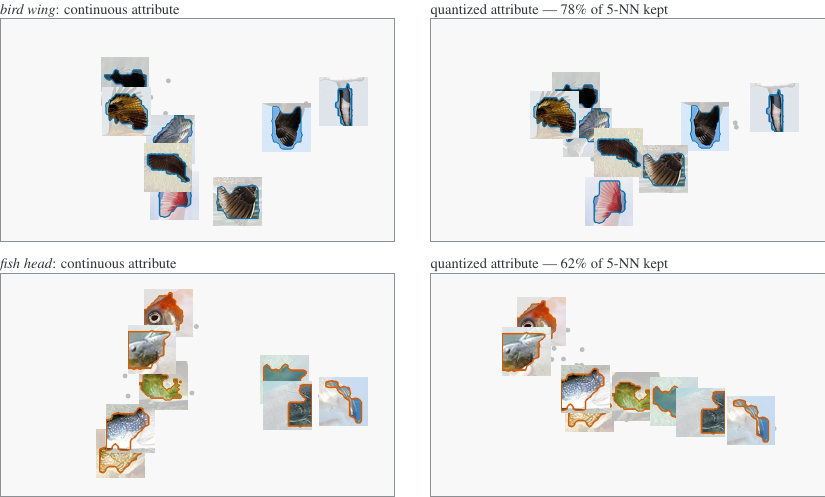}{PartImageNet}{fig:app-quant-pin}
\QualQuantFigure{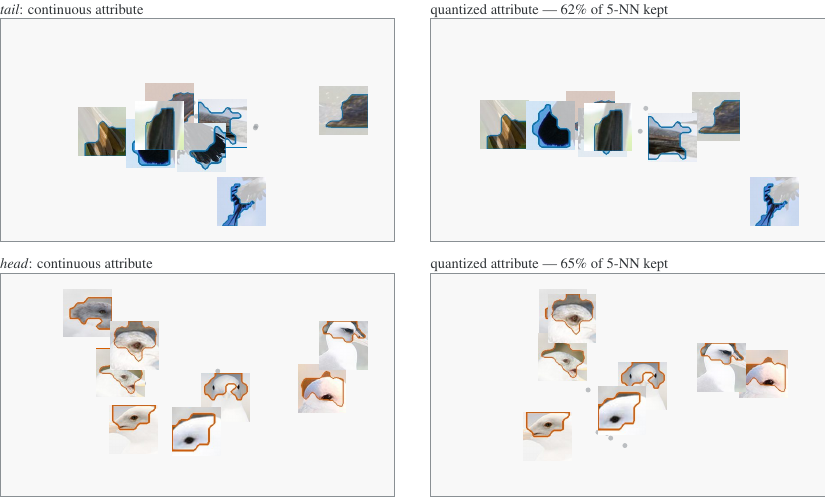}{CUB}{fig:app-quant-cub}
\QualQuantFigure{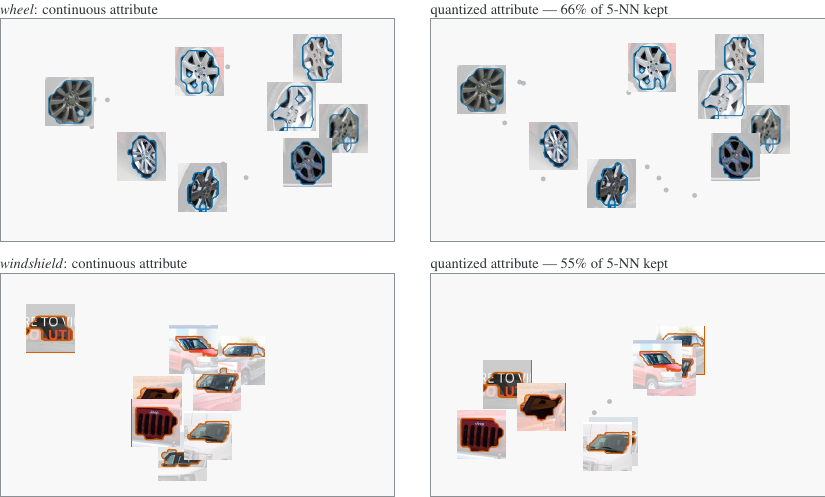}{Stanford Cars}{fig:app-quant-cars}
\QualQuantFigure{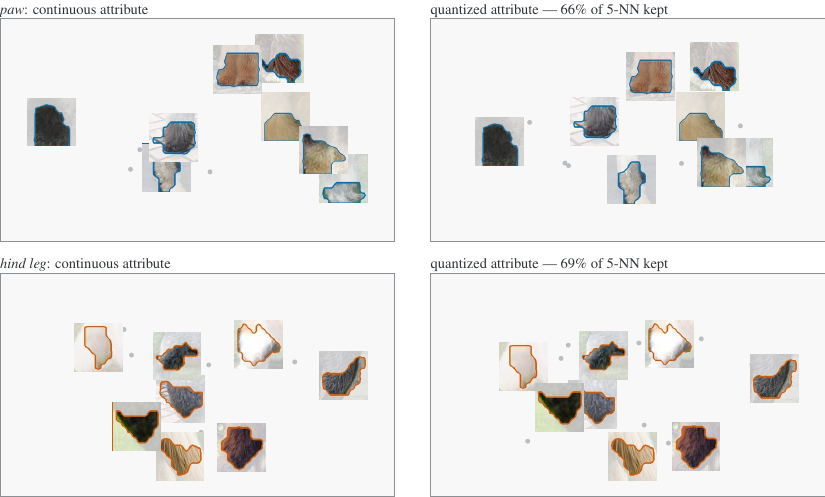}{Stanford Dogs}{fig:app-quant-dogs}
\QualQuantFigure{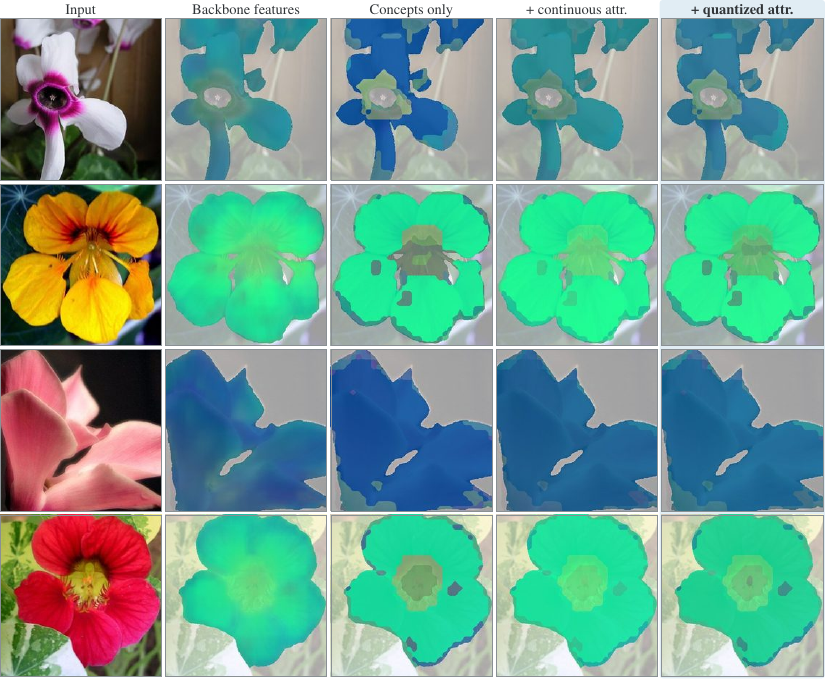}{Oxford Flowers}{fig:app-quant-flowers}

\clearpage

\subsection{How one concept varies across images}
\label{app:qualitative-attrspace}
This subsection shows how one concept varies across images (Figures~\ref{fig:app-attrspace-pin}--\ref{fig:app-attrspace-dogs}). Two occurrences of a concept share its prototype and differ in their attribute codes (Section~\ref{subsec:attribute-quant}). A concept occurs in an image when the model selects it among the $K{=}16$ concepts of that image and gives it at least $1.2\%$ of the $32\times32$ grid. Each map holds the quantized attributes $q_{i_k}$ (Equation~\ref{eq:private-pq}) of all occurrences of one concept in images not used in training: the PartImageNet validation and test splits ($3{,}614$ images) and the test splits of CUB ($5{,}794$), Stanford Cars ($8{,}041$) and Stanford Dogs ($8{,}580$). At evaluation, $q_{i_k}$ is the concatenation of the $G{=}16$ selected codewords, so it follows from the discrete code alone.

We fit a PCA to these vectors and draw each occurrence at its coordinates, as a crop with the concept's region outlined or as a dot. The plane holds $12$ to $20\%$ of the variance, so positions are approximate; the row below each map gives the exact neighbors of one occurrence (the input, the large crop), its four nearest and four farthest by cosine over the full $768$-dimensional attribute. Unlike the bottom panels of Figures~\ref{fig:app-quant-pin}--\ref{fig:app-quant-flowers}, which compare continuous and quantized attributes on a small gallery, these maps show the quantized attribute of every occurrence.

The frame of a crop gives the class of its image (the ImageNet class on PartImageNet, the species on CUB, the make on Stanford Cars and the breed on Stanford Dogs), with the five most frequent classes of the concept colored and the others gray; the outline color gives the distance to the input, from vermilion (nearest) to blue (farthest). The model never sees these labels during training. They are used here only to color the frames.\par
Within a concept, occurrences group by the kind of object, although no class label is used in training. The PartImageNet concept on the \emph{quadruped head} separates tigers, leopards, cougars, domestic cats and cheetahs, and the four nearest occurrences of the input leopard are leopards; the four nearest beaks to an evening grosbeak's belong to evening grosbeaks, the nearest grilles to a Mercedes-Benz C-Class come from the same model, and the nearest muzzles to a bull mastiff's are those of bull mastiffs.

This is the division of labor the model is built for. The concept identity is shared across classes and says \emph{which} part is present, such as a grille; the attribute says \emph{which kind} of that part it is, down to the species, the make or the breed. A description by concepts and attributes can therefore separate fine-grained classes that share all their parts, which is consistent with the small gap to the backbone on the fine-grained datasets (Table~\ref{tab:fine-grained}).\par

\myparagraph{What this does not show.} The attribute is computed from a frozen DINOv2 region feature, which already
separates classes, so part of this structure comes from the backbone. The figures show that the discrete code preserves
it within a single concept; they do not show that quantization adds it.

\newcommand{\AttrPcaFigure}[4]{\begin{figure*}[p]\centering
\includegraphics[width=\linewidth]{\RepoFigureRoot/figures/attr_pca/#1.pdf}%
\vspace{-2.3mm}%
\caption{\textbf{#2: how a concept of \ours{} varies across images.} \emph{Map:} PCA of the quantized attributes of all occurrences of the concept; each is a crop with its region outlined, or a dot; the dark-framed crop is the input. Frame color: the #3 of the image; outline color: distance to the input. \emph{Row:} the input, its four nearest and four farthest occurrences. On PartImageNet and CUB, a concept is named after the annotated part it falls on.\protect}
\label{#4}\end{figure*}}
\AttrPcaFigure{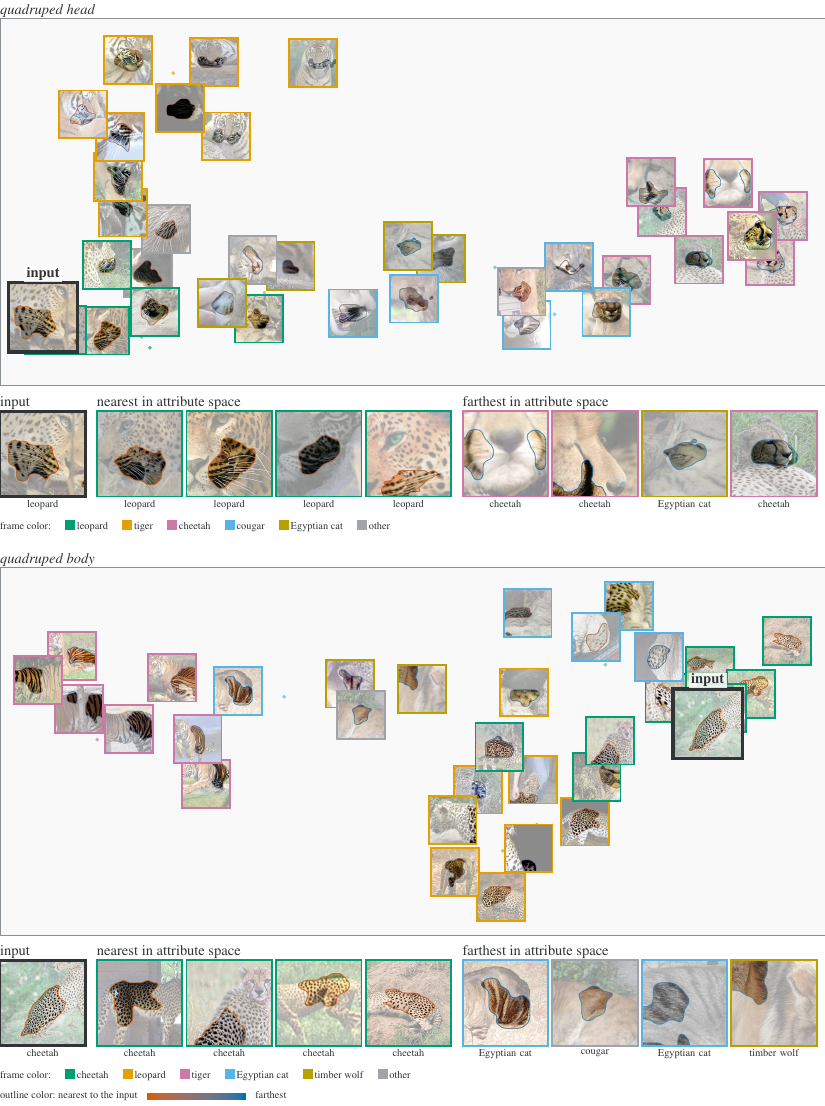}{PartImageNet, \emph{quadruped head} ($64$ images) and \emph{quadruped body} ($60$ images)}{ImageNet class}{fig:app-attrspace-pin}
\AttrPcaFigure{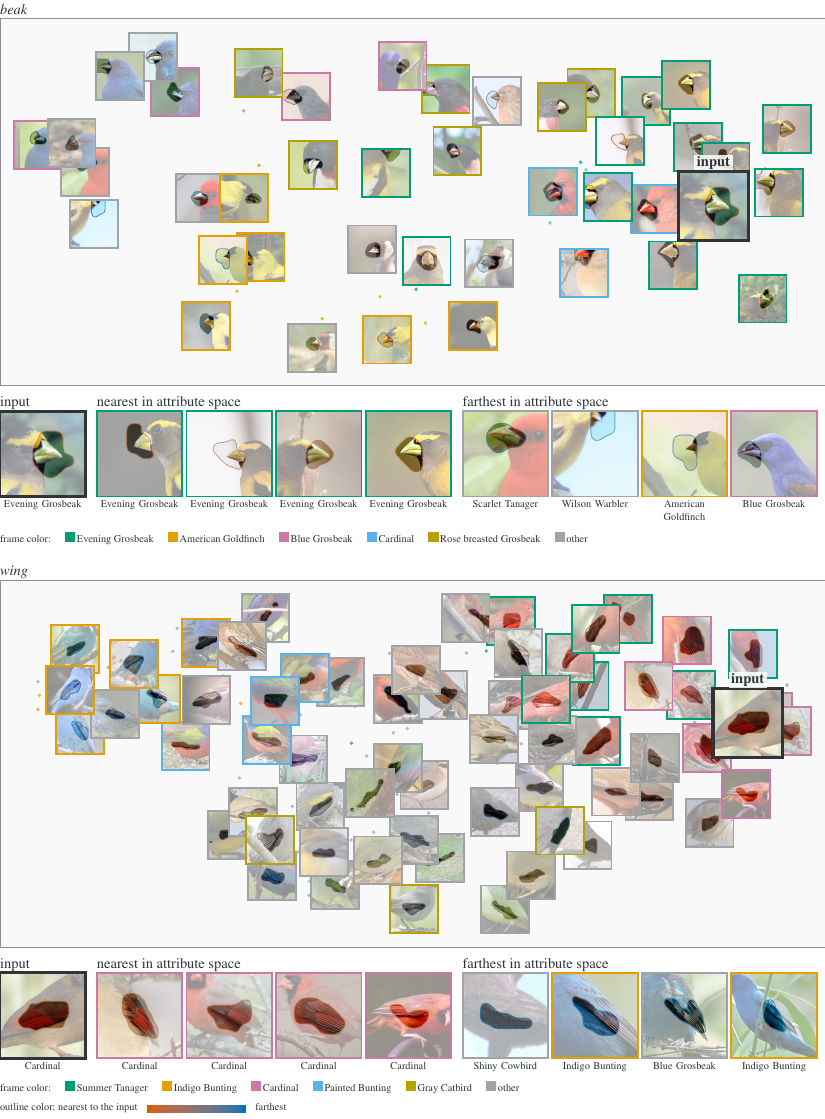}{CUB, \emph{beak} ($73$ images) and \emph{wing} ($191$ images)}{species}{fig:app-attrspace-cub}
\AttrPcaFigure{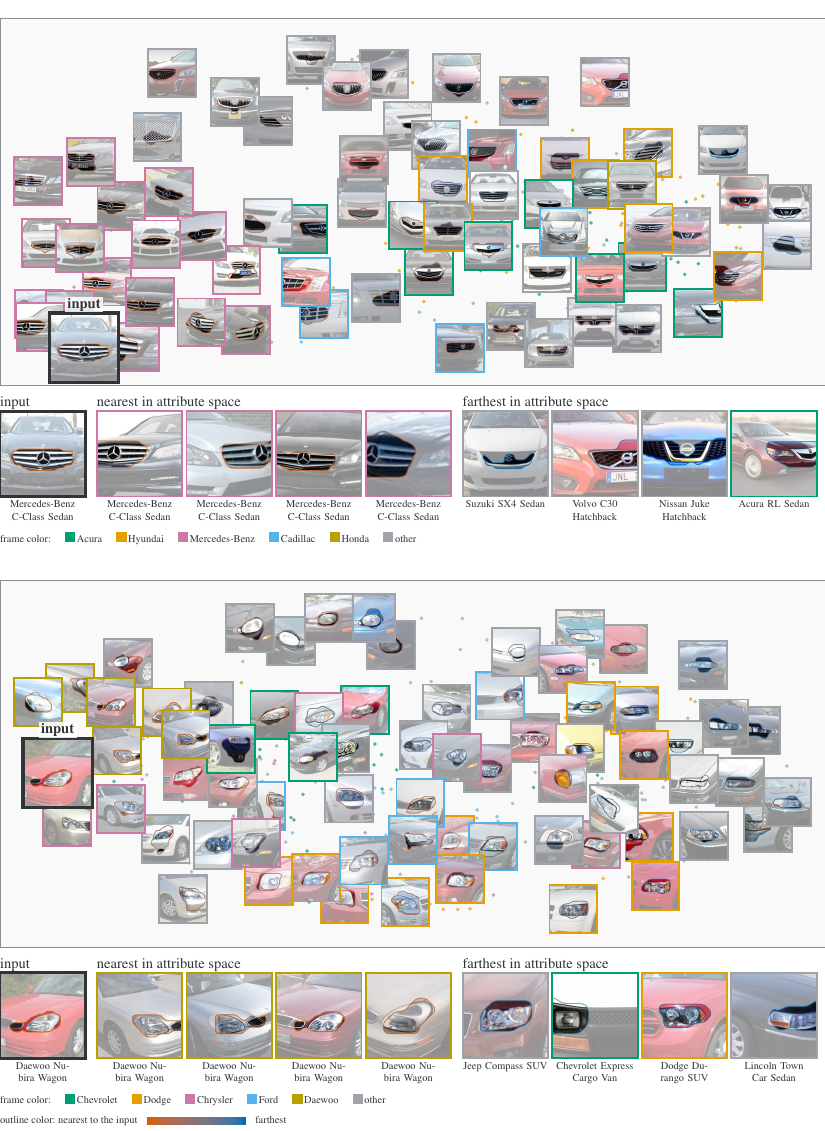}{Stanford Cars, two concepts ($288$ and $500$ images)}{make}{fig:app-attrspace-cars}
\AttrPcaFigure{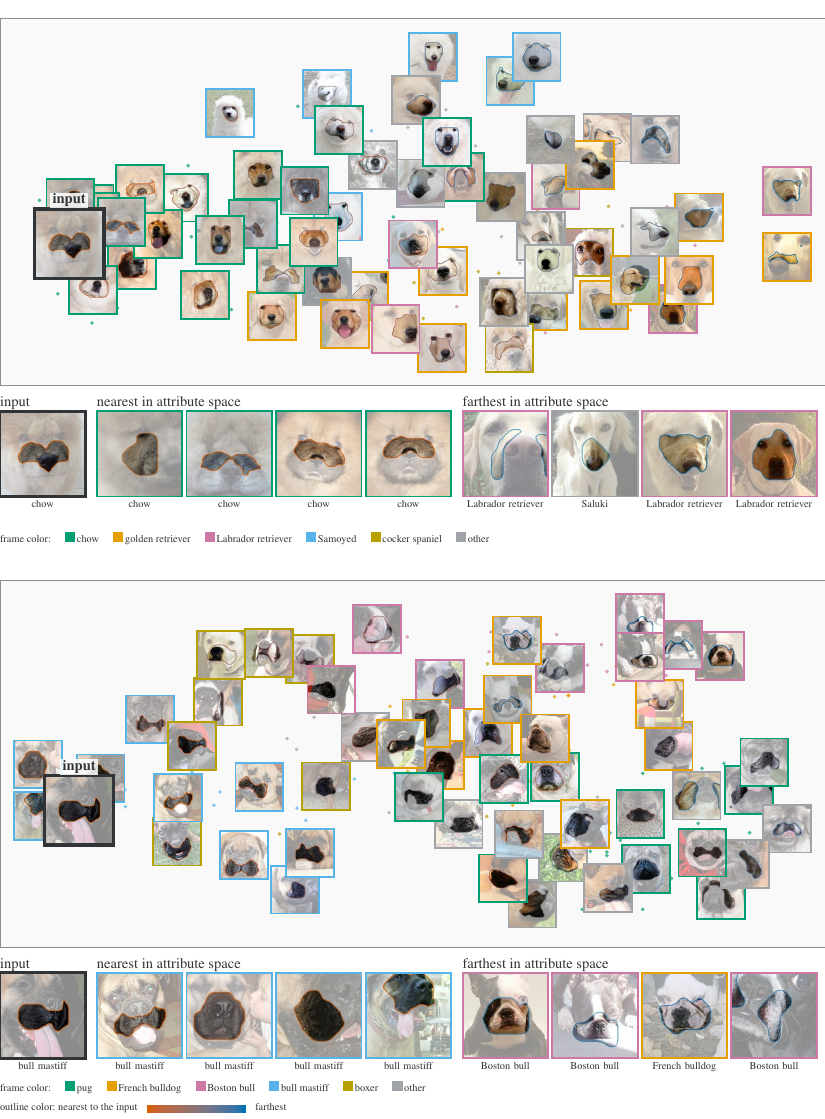}{Stanford Dogs, two concepts ($197$ and $212$ images)}{breed}{fig:app-attrspace-dogs}

\clearpage
\section{Automatic alignment of concept names}
\label{app:concept-naming}

\ours{} learns spatially grounded concepts without text supervision (Section~\ref{sec:architecture}). In this section, we investigate whether these concepts can be named after training using language-aligned features and a fixed vocabulary, and whether their quantized attributes help describe individual occurrences. We first align the frozen concept descriptions with a language-aligned visual representation, then retrieve names from a fixed vocabulary. Finally, we assess whether those names recover the object or part labels of their regions. All naming methods use the same regions within each model and evaluation dataset, so their comparison isolates the contribution of the naming rule.

\subsection{Concept descriptions and alignment targets}
\label{app:naming-alignment}

\myparagraph{Frozen concept models.}
We use three frozen \ours{} models, two based on DINOv2 and trained on ImageNet-1k and PartImageNet (PIN), respectively, and one based on CLIP-DINOiser and trained on PartImageNet. Following the backbone interfaces in Appendix~\ref{abl:backbones}, the DINOv2 models use $H=W=32$, while CLIP-DINOiser uses $H=W=28$ and omits layer fusion.

For each image $x$, we extract the selected concepts and hard identity map $p$ of Section~\ref{subsec:concept-assignment}. A concept occurrence consists of the locations assigned to a selected bank identity $n=i_k$, namely those with $p_u=n$. Its description combines the fixed prototype $w_n$ with the image-dependent attribute $q_n$ from Equation~\ref{eq:private-pq}. Naming therefore uses the same quantized attributes as the decoder. As in the method section, dependence on the current image is implicit.

\myparagraph{Language-aligned targets.}
To relate these descriptions to text, we use a frozen CLIP-DINOiser~\citep{wysoczanska2024clipdinoiser} teacher. Its projected patch features follow the language-aligned interface described in Appendix~\ref{abl:backbones}. We bilinearly resample them to the concept model's $H\times W$ grid, average over locations with $p_u=n$, and normalize the resulting vector to unit length. The resulting target $\varphi_n\in\mathbb R^{512}$ describes the image content assigned to concept $n$. Every comparison uses the same teacher and its OpenCLIP ViT-B/16 text encoder~\citep{cherti2023reproducible}, trained on LAION-2B~\citep{schuhmann2022laion}.

\myparagraph{Data and fitting split.}
For each concept model, we fit alignment independently on PartImageNet~\citep{he2022partimagenet} and VOC2012~\citep{everingham2012voc}. We split the 20,466 PartImageNet training images or 5,717 VOC2012 Main training images into 90\% for fitting and 10\% for adapter selection, using image-level split seed 45. Only the fitting subset determines adapter parameters, input statistics, and mean target vectors. No segmentation labels are used in fitting or selection. Evaluation uses the 1,206 PartImageNet validation images or 1,449 VOC2012 validation images. Each image undergoes a single shorter-side resize and center crop to $224\times224$, without sliding windows or multiple crops. Annotation masks undergo the paired transformation with nearest-neighbor interpolation. An occurrence used for fitting or adapter selection must occupy at least four assignment-grid cells.

\subsection{Concept descriptions to names}
\label{app:naming-methods}

A fixed name can describe what occurrences of a concept have in common, while an attribute-dependent name can also reflect differences between occurrences. We compare both approaches in the teacher's 512-dimensional space using the same vocabulary.

\myparagraph{Fixed concept descriptions.}
The \emph{region mean} averages $\varphi_n$ over fitting images for each concept and normalizes the mean to unit length. It provides a fixed description without learning an adapter. The \emph{linear key} and \emph{MLP key} methods instead learn a mapping from the prototype $w_n$ to the teacher space. Here ``key'' refers to the matching prototype $w_n$. The MLP has one hidden layer of width 512 with GELU. Both mappings produce one normalized vector per concept and therefore one fixed name. We also evaluate the normalized prototype directly (\emph{direct key}) for CLIP-DINOiser, whose prototypes occupy the compatible projected CLIP space. DINOv2 prototypes have no such direct correspondence to text.

\myparagraph{Attribute-dependent descriptions.}
To let names reflect individual occurrences, we add $q_n$ to the adapter input. Before fitting the adapters, we compute a mean and standard deviation for each input dimension across the occurrences in the fitting split, separately for prototypes and attributes. We subtract the corresponding mean and divide by the standard deviation for each adapter input. The \emph{additive} adapter applies a learned linear projection to each standardized input, adds the two outputs and a learned bias, and normalizes the result to unit length. The \emph{joint MLP} instead concatenates the standardized inputs and passes them through a 512-unit hidden layer with GELU, followed by a projection to the teacher space and unit-length normalization. This allows the attribute's contribution to depend on the prototype.

\myparagraph{Adapter fitting.}
Adapters minimize cosine distance to $\varphi_n$, with occurrences weighted inversely by their concept's fitting frequency. We use AdamW with learning rate $10^{-3}$, weight decay $10^{-4}$, batch size 512, and at most 100 epochs. We select the adapter with the highest holdout cosine similarity, averaged equally over concepts with at least five qualifying fitting-image occurrences. These are the supported concepts used below. Early stopping uses patience 15 and minimum improvement $10^{-5}$.

The \emph{pooled teacher} reference names each validation occurrence directly from $\varphi_n$, which is computed by running the teacher on that validation image and pooling its features over the concept region. The adapters instead predict a vector in the teacher's space from $w_n$ and, for the attribute-dependent methods, $q_n$. The reference therefore shows what names can be retrieved from the teacher's features on the same regions.

\myparagraph{Vocabulary retrieval.}
We use the filtered vocabulary supplied by the authors of CFM~\citep{wittenmayer2026cfm}, retaining 102,171 distinct strings after removing the malformed entry \texttt{(\#}. Each string has three normalized text embeddings, obtained by averaging the supplied noun, adjective, and verb prompt sets separately. Given a normalized visual description, we select the string with the largest cosine similarity across all strings and prompt groups. Every method uses this same vocabulary and prompt recipe, without CFM's hierarchy-specific naming correction. Dataset class names do not restrict the choice of a name.

\subsection{Evaluating the retrieved names}
\label{app:naming-evaluation}

\myparagraph{Name-to-class retrieval.}
We assess the selected name by asking whether it retrieves the annotated label of its region. We re-encode the name and the dataset class names with a shared prompt ensemble: the teacher's 80 ImageNet templates for VOC, and ``a photo of \{\}.'' / ``a close-up photo of \{\}.'' for PartImageNet. We then rank class embeddings by cosine similarity to the name embedding. The candidates are the 20 VOC object classes or 40 PartImageNet part classes, plus background. This evaluates the retrieved text through its embedding, rather than matching strings exactly or classifying directly from the adapter output.

\myparagraph{Eligible regions and metrics.}
We evaluate validation occurrences of supported concepts that occupy at least four assignment-grid cells. We upsample each hard region to the $224\times224$ annotation grid by nearest neighbor and ignore void pixels. For inclusion in the primary metric, the dominant ground-truth label must be foreground and cover at least 75\% of the region's valid pixels. This focuses evaluation on regions with a clear annotated label.

Let $R_n(x)$ be the rank of that label when retrieved from the predicted name, and let $\mathcal O_n$ contain eligible validation images for concept $n$. We average first over occurrences of each concept and then over the set $\mathcal C$ of concepts with at least one eligible occurrence:
\begin{equation}
 \begin{aligned}
 \mathrm{Acc}_{\mathrm{name}}&=\frac{1}{|\mathcal C|}\sum_{n\in\mathcal C}\frac{1}{|\mathcal O_n|}\sum_{x\in\mathcal O_n}\mathbb{1}[R_n(x)=1],\\
 \mathrm{MRR}&=\frac{1}{|\mathcal C|}\sum_{n\in\mathcal C}\frac{1}{|\mathcal O_n|}\sum_{x\in\mathcal O_n}\frac{1}{R_n(x)}.
 \end{aligned}
\end{equation}
Top-1 accuracy measures how often the correct label ranks first, while mean reciprocal rank (MRR) also credits labels retrieved below first place. Tables~\ref{tab:naming-pin} and~\ref{tab:naming-voc} report $100\mathrm{Acc}_{\mathrm{name}}$ and MRR, together with the eligible region and concept counts. Those subsets differ across concept models, including for the pooled-teacher reference. No human or language-model judge is used. The text encoder both selects names and scores their correspondence to class labels, so the metric is not an independent semantic judgment. The confidence intervals measure variation across sampled concepts, not across training seeds.

\begin{table}[t]
\centering\small
\setlength{\tabcolsep}{5pt}
\begin{tabular}{lrrrrrr}
\toprule
& \multicolumn{2}{c}{ImageNet / DINOv2} & \multicolumn{2}{c}{PIN / DINOv2} & \multicolumn{2}{c}{PIN / CLIP-DINOiser} \\
\cmidrule(lr){2-3}\cmidrule(lr){4-5}\cmidrule(lr){6-7}
Method & Top-1 & MRR & Top-1 & MRR & Top-1 & MRR \\
\midrule
Region mean & 15.28 & 0.2764 & 23.18 & 0.3623 & 26.88 & 0.4133 \\
Direct key & \textemdash & \textemdash & \textemdash & \textemdash & 31.95 & 0.4486 \\
Linear key & 12.29 & 0.2468 & 22.95 & 0.3598 & 27.40 & 0.4245 \\
MLP key & 14.82 & 0.2722 & 22.29 & 0.3544 & 26.30 & 0.4074 \\
Additive & 16.04 & 0.2905 & 22.53 & 0.3651 & 30.71 & 0.4534 \\
Joint MLP & 23.74 & 0.3691 & 21.74 & 0.3553 & 26.86 & 0.4136 \\
\midrule
Pooled teacher (reference) & 29.75 & 0.4314 & 26.02 & 0.3959 & 30.97 & 0.4544 \\
\bottomrule
\end{tabular}
\caption{PartImageNet name-to-class retrieval on validation regions. Top-1 is a percentage, while MRR lies in $[0,1]$. Both are macro-averaged over concepts. Column groups identify concept-training dataset / backbone. Adapters are fitted independently on PartImageNet training images. Eligible region / concept counts are 5,541 / 467, 4,758 / 416, 1,208 / 171, respectively. The teacher reference pools features from each model's own regions.}\label{tab:naming-pin}
\end{table}
\begin{table}[t]
\centering\small
\setlength{\tabcolsep}{5pt}
\begin{tabular}{lrrrrrr}
\toprule
& \multicolumn{2}{c}{ImageNet / DINOv2} & \multicolumn{2}{c}{PIN / DINOv2} & \multicolumn{2}{c}{PIN / CLIP-DINOiser} \\
\cmidrule(lr){2-3}\cmidrule(lr){4-5}\cmidrule(lr){6-7}
Method & Top-1 & MRR & Top-1 & MRR & Top-1 & MRR \\
\midrule
Region mean & 30.09 & 0.4643 & 41.28 & 0.5490 & 39.96 & 0.5531 \\
Direct key & \textemdash & \textemdash & \textemdash & \textemdash & 44.97 & 0.5878 \\
Linear key & 27.90 & 0.4384 & 41.45 & 0.5515 & 44.32 & 0.5794 \\
MLP key & 29.13 & 0.4537 & 41.54 & 0.5512 & 39.40 & 0.5481 \\
Additive & 41.56 & 0.5515 & 54.07 & 0.6396 & 58.50 & 0.6895 \\
Joint MLP & 53.14 & 0.6438 & 61.54 & 0.7004 & 58.17 & 0.6936 \\
\midrule
Pooled teacher (reference) & 73.82 & 0.8128 & 80.21 & 0.8630 & 73.33 & 0.8103 \\
\bottomrule
\end{tabular}
\caption{VOC2012 name-to-class retrieval on validation regions. Top-1 is a percentage, while MRR lies in $[0,1]$. Both are macro-averaged over concepts. Column groups identify concept-training dataset / backbone. Adapters are fitted independently on VOC training images. Eligible region / concept counts are 7,567 / 633, 7,088 / 368, 3,576 / 190, respectively. The teacher reference pools features from each model's own regions.}\label{tab:naming-voc}
\end{table}

\subsection{Results}

\myparagraph{Object names on VOC.}
The joint MLP improves top-1 retrieval over region means by $23.05$, $20.26$, and $18.21$ percentage points for ImageNet/DINOv2, PartImageNet/DINOv2, and PartImageNet/CLIP-DINOiser, respectively (Table~\ref{tab:naming-voc}). Paired concept-bootstrap 95\% intervals are $[20.35,25.48]$, $[16.86,23.56]$, and $[13.75,22.33]$ points, using 1,000 resamples. The additive adapter performs slightly better than the joint MLP for PartImageNet/CLIP-DINOiser ($58.50$ versus $58.17$). Attribute-dependent adapters therefore improve object-name retrieval over fixed region-mean names in all three settings.

\myparagraph{Part names on PartImageNet.}
The gains do not extend uniformly to part labels (Table~\ref{tab:naming-pin}). The joint MLP improves over region means by 8.46 points for ImageNet/DINOv2 (95\% interval $[5.96,10.83]$), but changes PartImageNet/DINOv2 and PartImageNet/CLIP-DINOiser by $-1.44$ and $-0.02$ points, with intervals $[-3.22,0.40]$ and $[-3.53,2.98]$. Direct CLIP-DINOiser prototypes remain competitive. Better agreement with the teacher also need not yield better part names: for PartImageNet/DINOv2, joint-MLP cosine similarity to the pooled teacher increases from the region mean's $0.8169$ to $0.8615$, while top-1 retrieval falls from $23.18$ to $21.74$. The pooled teacher itself achieves only $26--31\%$ top-1 on PartImageNet part labels.

\subsection{Qualitative examples}

\myparagraph{Gallery construction.}
For each supported concept, the gallery contains its closest qualifying validation image and up to nine additional, distinct images sampled reproducibly. A displayed occurrence must be among the image's $K$ selected concepts and occupy at least 1\% of the assignment grid. The closest image minimizes the prototype distance $\|z_u-w_n\|_2^2$ from Equation~\ref{eq:bank-competition} over image-grid locations. The gallery separates fixed names from attribute-dependent names and includes pooled-teacher predictions, with five candidate names available per method. The 1\% threshold applies only to visualization. Quantitative evaluation retains the four-cell threshold above.

\begin{figure*}[p]
  \centering
  \includegraphics[width=\textwidth]{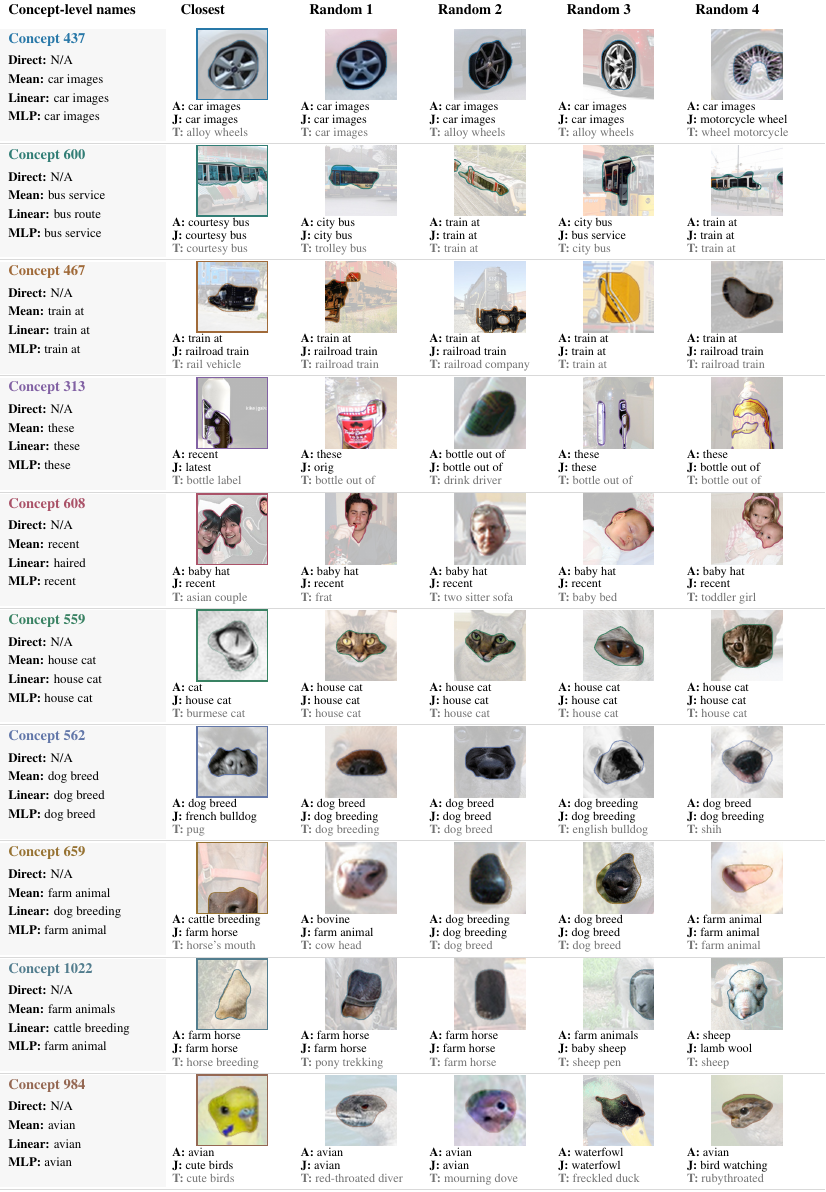}
  \caption{\textbf{PartImageNet-trained DINOv2 concepts named on VOC (selected examples).}
  Each row shows the closest occurrence and four saved random examples.
  Left: fixed direct-key, region-mean, linear-key, and MLP-key names. Direct CLIP naming is undefined for DINOv2 (N/A).
  Below each image: \textbf{A}, additive, \textbf{J}, joint MLP, gray \textbf{T}, teacher.}
  \label{fig:naming-dino-voc}
\end{figure*}

\begin{figure*}[p]
  \centering
  \includegraphics[width=\textwidth]{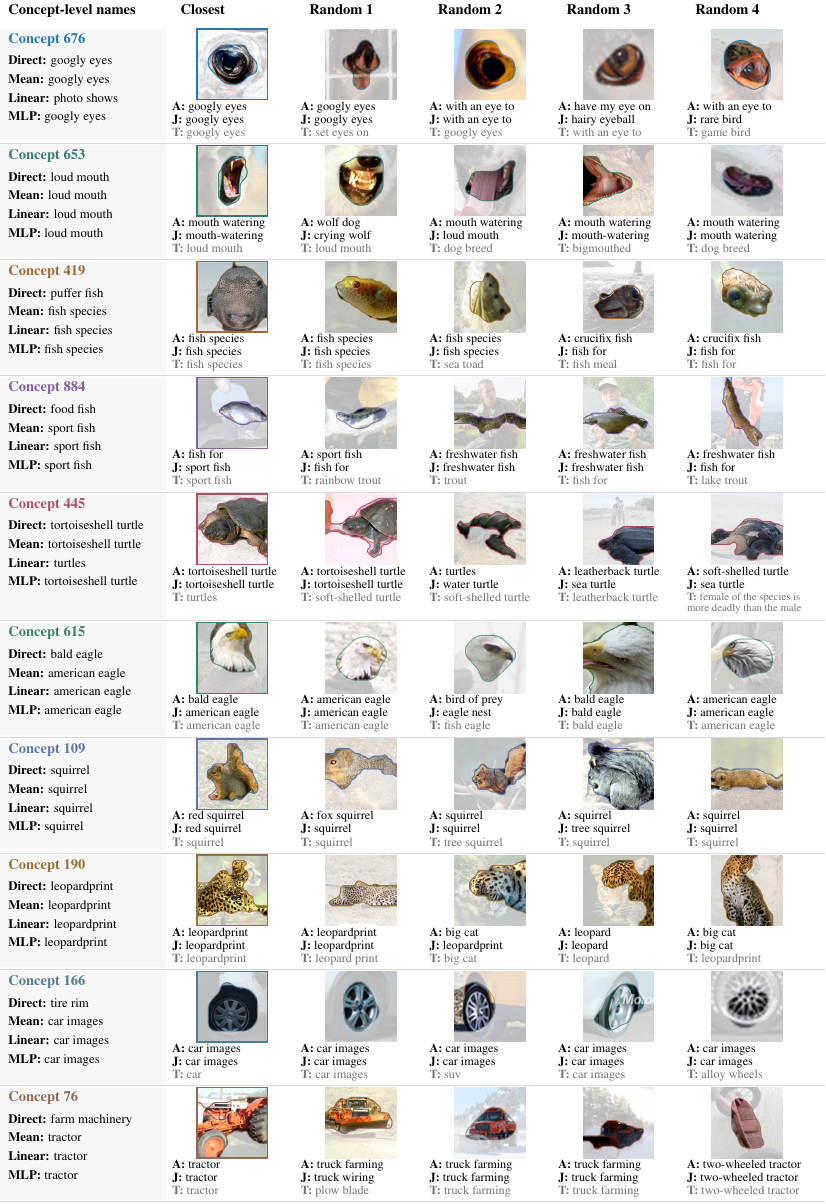}
  \caption{\textbf{PartImageNet-trained CLIP-DINOiser concepts named on PartImageNet (selected examples).}
  Each row shows the closest occurrence and four saved random examples.
  Direct names use the compatible projected CLIP prototype space. Labels denote \textbf{A}, additive, \textbf{J}, joint MLP, gray \textbf{T}, teacher.}
  \label{fig:naming-clip-pin}
\end{figure*}

\myparagraph{Selection for the figures.}
Figures~\ref{fig:naming-dino-voc} and~\ref{fig:naming-clip-pin} each show ten concepts. To select them, we exclude background-associated concepts by requiring at least 85\% mean foreground overlap across their saved gallery examples and at least 75\% in each of the five displayed examples. We then inspect contact sheets to select concepts with diverse classes, excluding rows with potentially inappropriate displayed labels. Each row retains the closest example and four random examples. The figures illustrate differences between methods rather than estimating naming accuracy. All displayed names are unedited top-one predictions.

\begin{figure*}[p]
  \centering
  \includegraphics[width=\textwidth]{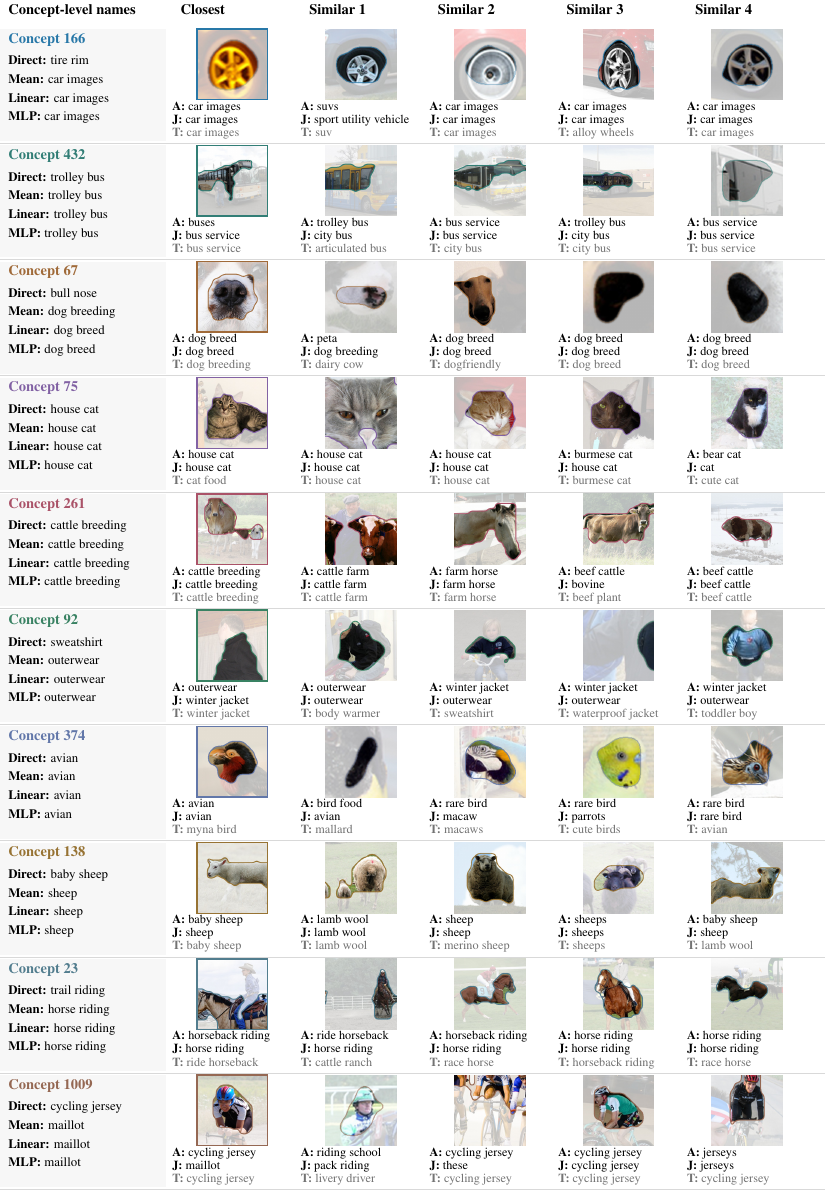}
  \caption{\textbf{PartImageNet-trained CLIP-DINOiser concepts named on VOC (curated examples).}
  Each row shows the closest occurrence and four visually similar saved examples.
  Fixed names appear at left; per-image labels are \textbf{A}, additive, \textbf{J}, joint MLP, and gray \textbf{T}, teacher.}
  \label{fig:naming-clip-voc}
\end{figure*}

\begin{figure*}[p]
  \centering
  \includegraphics[width=\textwidth]{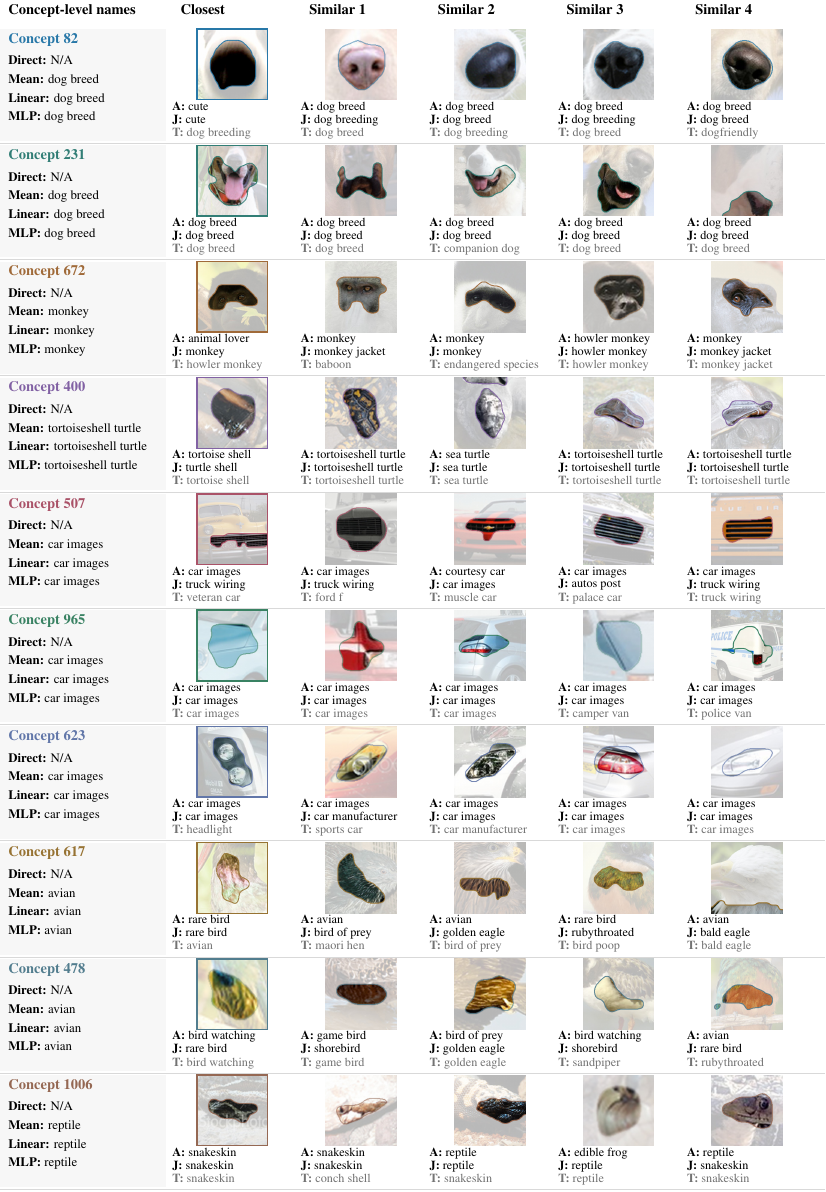}
  \caption{\textbf{ImageNet-trained DINOv2 concepts named on PartImageNet (curated examples).}
  Each row shows the closest occurrence and four visually similar examples selected from the saved gallery. Fixed names appear at left, and the names below each image are \textbf{A}, additive, \textbf{J}, joint MLP, and gray \textbf{T}, teacher.}
  \label{fig:naming-imagenet-pin}
\end{figure*}

\myparagraph{Additional selected examples.}
Figures~\ref{fig:naming-clip-voc} and~\ref{fig:naming-imagenet-pin} apply the same support, foreground-overlap, label-screening rules to two additional model--dataset pairs. Each row retains the closest saved occurrence and four visually similar examples chosen manually from the remaining saved gallery images.

\end{document}